\documentclass[journal]{vgtc}                
\ifpdf
  \pdfoutput=1\relax                   
  \usepackage{graphicx}                
  \DeclareGraphicsExtensions{.pdf,.png,.jpg,.jpeg} 
\else
  \usepackage{graphicx}                
  \DeclareGraphicsExtensions{.eps}     
\fi%

\graphicspath{{figures/}{pictures/}{images/}{./}} 

\usepackage{microtype}                 
\PassOptionsToPackage{warn}{textcomp}  
\usepackage{textcomp}                  
\usepackage{mathptmx}                  
\usepackage{times}                     
\usepackage{cite}                      
\usepackage{tabu}                      
\usepackage{booktabs}                  

\usepackage{amsmath}

\usepackage{color} 
\usepackage{colortbl}
\usepackage{cases}
\definecolor{light-gray}{gray}{0.6}
\definecolor{lavender}{rgb}{0.5,0.5,1.0}

\def\_{\rule{.3em}{.15ex}}      

\newcommand{\hot}[1]{{\color{black}#1}}
\newcommand{\rev}[1]{{\color{black}#1}}

\usepackage{xspace}
\usepackage{placeins}

\newcommand{\ours}{ours \xspace}

\newenvironment{myitemize}{
\begin{itemize}
  \setlength{\itemsep}{1pt}
  \setlength{\parskip}{0pt}
  \setlength{\parsep}{0pt}}{\end{itemize}
}

\usepackage{lipsum}
\usepackage{multirow}

\usepackage{tikz}
\definecolor{SomeColor}{RGB}{0,150,255}

\usepackage{caption}
\usepackage{subcaption}
\usepackage{amssymb}
\usepackage{float}

\onlineid{1640}

\vgtccategory{Research}
\vgtcpapertype{Technique}

\title{
From Scalars to Time Series: Rethinking Implicit Neural Representations for Time-Varying Volumetric Data
}

\author{
Weihan Zhang,
Xuan Zhao,
Yenwen Peng,
Yuqi Chen,
and Jun Tao, \textit{Member, IEEE}
}

\authorfooter{
\item
Weihan Zhang, Xuan Zhao, and Yenwen Peng are with the
School of Computer Science and Engineering, Sun Yat-sen University.
E-mail: \{zhangwh79, zhaox269, pengyw23\}@mail2.sysu.edu.cn.

\item
Yuqi Chen is with the Department of Computer Science,
Hong Kong Baptist University.
E-mail: csyqchen@comp.hkbu.edu.hk.

\item
Jun Tao is with the School of Computer Science and Engineering,
Sun Yat-sen University, Guangdong Province Key Laboratory of
Computational Science, and National Supercomputer Center in Guangzhou.
E-mail: taoj23@mail.sysu.edu.cn.
He is the corresponding author.
}

\abstract{
Implicit neural representations (INRs) for time-varying volumetric data are typically trained using dense sampling over spatiotemporal coordinates, where each observation corresponds to a single point in space and time. This coordinate-wise formulation requires extensive sampling during optimization, leading to high computational cost and inefficient use of temporal structure.
In this work, we revisit this design choice and show that dense spatiotemporal sampling is not necessary for learning time-varying fields. Instead, we represent the data as a collection of spatially indexed time series and train INRs using sequence-level supervision over each spatial location, rather than coordinate-wise scalar samples. This reformulation eliminates the need for dense spatiotemporal sampling and instead learns each spatial location from its full temporal evolution in a structured manner.
We demonstrate that this representation is compatible with a range of existing INR architectures and consistently improves reconstruction quality, while significantly reducing training cost. Furthermore, we show that this formulation can be combined with mixture-of-experts architectures, and that our MoE instantiation further improves reconstruction quality compared to both the base reformulation and existing MoE-based INR methods, providing a stronger capacity allocation under heterogeneous temporal dynamics.
} 

\keywords{Volume visualization, time-varying volumetric data, implicit neural representation}

\begin{document}



\maketitle

\section{Introduction}
Efficient handling of large-scale scientific data is essential for downstream tasks such as visualization and analysis, making data compression a fundamental component in modern workflows. Recently, implicit neural representations (INRs), also referred to as coordinate-based neural networks, have demonstrated strong capability in modeling complex high-dimensional signals with compact parameterizations, and have been increasingly explored for time-varying volumetric data. However, existing INR-based approaches~\cite{sitzmann2020siren,han2022coordnet,lu2021neurcomp} typically rely on a coordinate-wise formulation, where each observation corresponds to a scalar value at a specific spatiotemporal coordinate. Learning under this formulation requires dense sampling across both space and time, leading to substantial computational cost, with training often taking hours or even days. To mitigate this issue, prior work has explored partitioning strategies along spatial~\cite{tang2024ecnr,han2025dcinr} or temporal~\cite{han2023kdinr} dimensions, dividing the domain into smaller regions and training separate models to accelerate optimization by exploiting local continuity.

However, these approaches share a fundamental limitation. They continue to rely on a coordinate-wise formulation, where learning is driven by dense sampling over the joint spatiotemporal domain. While partitioning strategies reduce the effective scale of optimization, they do not remove the need for extensive sampling, and thus only partially alleviate the computational burden. In addition, such methods often introduce increased design complexity and may disrupt continuity across spatial or temporal boundaries, leading to artifacts and reduced robustness. More importantly, by treating each observation as an independent scalar value, these formulations fail to fully exploit the structured temporal coherence inherent in time-varying data. 

To address this limitation, we revisit the conventional formulation of INR-based compression for time-varying data. 
\hot{Rather than viewing a time-varying volume only as a dense four-dimensional scalar field, we reinterpret it as a spatial grid of temporal sequences.
This changes the basic learning unit from an individual spatiotemporal scalar sample to the complete temporal trajectory associated with each spatial location.}
Instead of relying on dense spatiotemporal sampling, we reformulate the problem by organizing the data as a set of spatially indexed temporal sequences. Under this formulation, each spatial location is associated with its full temporal evolution, and the network is trained to learn sequence-level mappings rather than independent scalar values.
This perspective eliminates the need for exhaustive sampling in the joint spatiotemporal domain and enables a more structured representation of temporal dynamics. 

Building upon this reformulation, we develop \hot{a framework} for modeling time-varying volumetric data using coordinate-based neural networks. In this framework, the input to the model is a spatial coordinate, and the output is the corresponding temporal signal. 
This abstraction is independent of specific network designs and can be applied to a wide range of INR architectures. 
By shifting the unit of learning from individual space-time samples to temporally coherent signals, the framework eliminates the need for dense sampling in the joint spatiotemporal domain and better aligns the learning process with the inherent temporal structure of the data.

While \hot{this framework} can benefit a wide range of coordinate-based neural networks, it is particularly well suited for mixture-of-experts (MoE)~\cite{nowlan1990moe,han2025moeinr,zhao2023moec} architectures, as it explicitly exposes heterogeneity in temporal patterns across the spatial domain. Under the proposed representation, each spatial location is associated with a time series, and different regions in the volume often exhibit distinct temporal dynamics. This spatial variation in temporal behavior naturally calls for adaptive modeling strategies that can allocate capacity according to local characteristics.

In this context, MoE models provide a natural mechanism to exploit such heterogeneity. By assigning different experts to spatial locations with distinct temporal patterns, the model can specialize in modeling diverse types of temporal dynamics, rather than relying on a single shared representation. This leads to a more flexible and expressive model that better captures non-uniform patterns across the domain.

Based on this insight, we develop an MoE-based INR model within the proposed framework. To improve expert routing, we incorporate a trainable embedding that captures global temporal characteristics and is fused with spatial features to guide expert assignment. In addition, to control the parameter growth introduced by sequence prediction and multiple experts, we adopt a lightweight decoder design based on low-rank adaptation (LoRA)~\cite{hu2022lora,zhang2025kimi}, enabling efficient specialization with limited overhead.
We evaluate our framework on multiple time-varying volumetric datasets from several perspectives, including reconstruction numerical accuracy, perceptual quality, and geometric consistency. Both quantitative results and visual comparisons show that our framework achieves competitive reconstruction performance against existing learning-based methods and traditional compression techniques. In addition, our framework demonstrates significant compression and decompression efficiency, benefiting from the sequence-level formulation and one-shot prediction strategy.

In summary, our contributions can be described as follows:
\begin{myitemize}
\item We propose \hot{a framework} that reformulates time-varying volumetric data as spatially indexed temporal sequences. This shifts coordinate-based neural representations from point-wise scalar prediction to sequence-level learning, providing a unified perspective for structured and efficient modeling of temporal dynamics.

\item Building on this framework, we exploit spatial heterogeneity in temporal patterns, where different regions exhibit distinct dynamics. This motivates a mixture-of-experts design that assigns experts to locations with different temporal behaviors. A trainable temporal embedding is introduced to guide expert routing.

\item We evaluate the proposed framework on multiple time-varying datasets, showing consistent compression quality across compression ratios. The formulation also yields substantial efficiency gains and generalizes across INR architectures.
\end{myitemize}
\section{Related Work}

\textbf{Lossy volumetric data compression.}
With the increasing demand for data reduction in scientific computing, considerable research efforts have been directed toward compressing large-scale scientific simulation data. Among them, lossy compression methods for volumetric data have been a central research focus for several decades.
Gobbetti~et~al.~\cite{gobbetti2012covra} developed a dictionary-based multi-resolution compression and rendering architecture to support interactive GPU visualization of massive volumetric datasets.
Lindstrom~\cite{lindstrom2014zfp} introduced a block-based transform compression method for floating-point volumetric data that achieves high compression efficiency and fast throughput.
TAMRESH~\cite{suter2013tamresh} and TTHRESH~\cite{ballester2019tthresh} were developed to exploit tensor decomposition frameworks and enable efficient compression of multidimensional data on regular grids.
\rev{Representative error-bounded methods include SZ3~\cite{liang2022sz3} and TopoSZ~\cite{yan2023toposz}, which aim to improve reconstruction fidelity while satisfying specified error constraints.
Hierarchical compression methods~\cite{ainsworth2019multilevel,hoang2020efficient} leverage multilevel representations to support adaptive error control and flexible resolution–precision trade-offs.}

\textbf{Deep learning for volumetric data.}
\rev{Deep learning has been widely adopted in scientific visualization for volumetric data representation and analysis. Early studies primarily relied on convolutional architectures for volumetric super-resolution, adaptive sampling, and reconstruction~\cite{zhou2017volume,weiss2019volumetric,weiss2020learning,wurster2022deep}. Generative adversarial networks (GANs) were subsequently explored for temporal and spatial super-resolution, volume generation, compression, and completion tasks~\cite{han2019tsrtvd,han2020ssrtvd,han2021stnet,liu2019novel,han2022vcnet}. More recently, coordinate-based neural representations have emerged as an effective paradigm for volumetric compression and reconstruction through continuous coordinate-to-value mappings~\cite{lu2021neurcomp,han2022coordnet,sahoo2022neural}.}

\textbf{Implicit neural representation.}
Recent approaches~\cite{sitzmann2020siren,han2022coordnet} adopt implicit neural representations (INRs) to model volumetric data through coordinate-based mappings. For example, SIREN~\cite{sitzmann2020siren} introduces periodic activations to better capture high-frequency signals, while Instant-NGP~\cite{muller2022instant} leverages hash encoding to significantly accelerate training. 
To improve compression, KD-INR~\cite{han2023kdinr} employs knowledge distillation to encode large-scale time-varying data into a single network.
To further reduce optimization cost, meta-learning-based methods such as MetaSDF~\cite{sitzmann2020metasdf} and Meta-INR~\cite{yang2025metainr} reuse learned priors to enable faster convergence across different volumetric datasets.

For volumetric data, many INR-based methods incorporate spatial partitioning strategies, dividing the domain into blocks to better capture local patterns and improve representation efficiency.
Martel~et~al.~\cite{martel2021acorn} employed an octree-based hierarchical partitioning strategy to decompose data into blocks, which were then modeled using implicit neural representations to enhance data representation.
Reiser~et~al.~\cite{reiser2021kilonerf} replaced a single large MLP with thousands of compact neural fields distributed across spatial partitions, achieving substantial speedups through parallelization.
Han~et~al.~\cite{han2025dcinr} proposed DCINR, which adopts a divide-and-conquer strategy by partitioning time-varying volumetric data into non-overlapping blocks, enabling independent modeling and parallel training of each block.
Beyond block-wise partitioning, allowing different blocks to operate at adaptive scales further enhances representation flexibility. Saragadam~et~al.~\cite{saragadam2022miner} proposed MINER, a multi-scale implicit neural representation designed for image and mesh reconstruction.
Tang~et~al.~\cite{tang2024ecnr} proposed ECNR, which partitions volumetric data into small blocks and assigns similar blocks to the same MLP to enable balanced parallelization.

\hot{Beyond explicit spatial partitioning, recent works have explored alternative mechanisms for handling heterogeneous scientific data.
Distributed neural representations~\cite{wu2025distributed} decompose large-scale simulations across multiple neural fields to support scalable training and reactive in-situ visualization.
Another line of research employs parameter-conditioned representations and hypernetworks to model data variability.
For example, HyperINR~\cite{wu2023hyperinr} predicts compact INR weights through a hypernetwork, while HyperFLINT~\cite{Gadirov2025hyperflint} uses parameter-conditioned networks for flow estimation and temporal interpolation in scientific ensemble visualization.
Compared with these approaches, our work focuses on modeling heterogeneous temporal dynamics through sequence-based representations and expert specialization.
}

Mixture-of-experts (MoE) architectures have also been applied to implicit neural representations for 3D scene modeling~\cite{zhenxing2022switchnerf,ben2024neuralexperts}. MoEC~\cite{zhao2023moec} employs a learnable gating network to jointly optimize spatial partitioning and local INRs, enabling adaptive and efficient compression. 
Han~et~al.~\cite{han2025moeinr} proposed a mixture-of-experts–based implicit neural representation that adaptively partitions time-varying volumetric data to achieve improved compression accuracy.

\textbf{Comparison with the existing works.}
Compared to existing INR-based approaches, our framework differs in several key aspects. Unlike grid-based methods that rely on predefined spatial partitioning~\cite{han2025dcinr,tang2024ecnr}, we do not impose explicit spatial divisions, but instead exploit temporal information by directly modeling full temporal sequences. In contrast to temporal decomposition strategies~\cite{han2023kdinr}, which partition data along the time dimension, our formulation captures the complete temporal evolution at each spatial location. Furthermore, while prior MoE-based INRs~\cite{zhao2023moec,han2025moeinr} focus on enhancing representation capacity under coordinate-to-scalar mappings, our work reformulates the task as a mapping from spatial coordinates to temporal sequences, enabling more flexible capacity allocation through routing.

\section{Framework}
This section presents \hot{our framework} for modeling and compressing time-varying volumetric data, together with its MoE-based network instantiation. The framework begins with a sequence-wise reformulation that treats each spatial location as a temporal signal, shifting the learning objective from point-wise scalar prediction to structured sequence modeling. This perspective highlights strong spatial heterogeneity in temporal dynamics, where different regions exhibit distinct evolution patterns. To exploit this structure, we introduce a mixture-of-experts (MoE)-based INR architecture that performs adaptive routing based on spatial features, followed by a lightweight decoder for efficient sequence reconstruction.


\begin{figure*}
    \centering
    \includegraphics[width=1\linewidth]{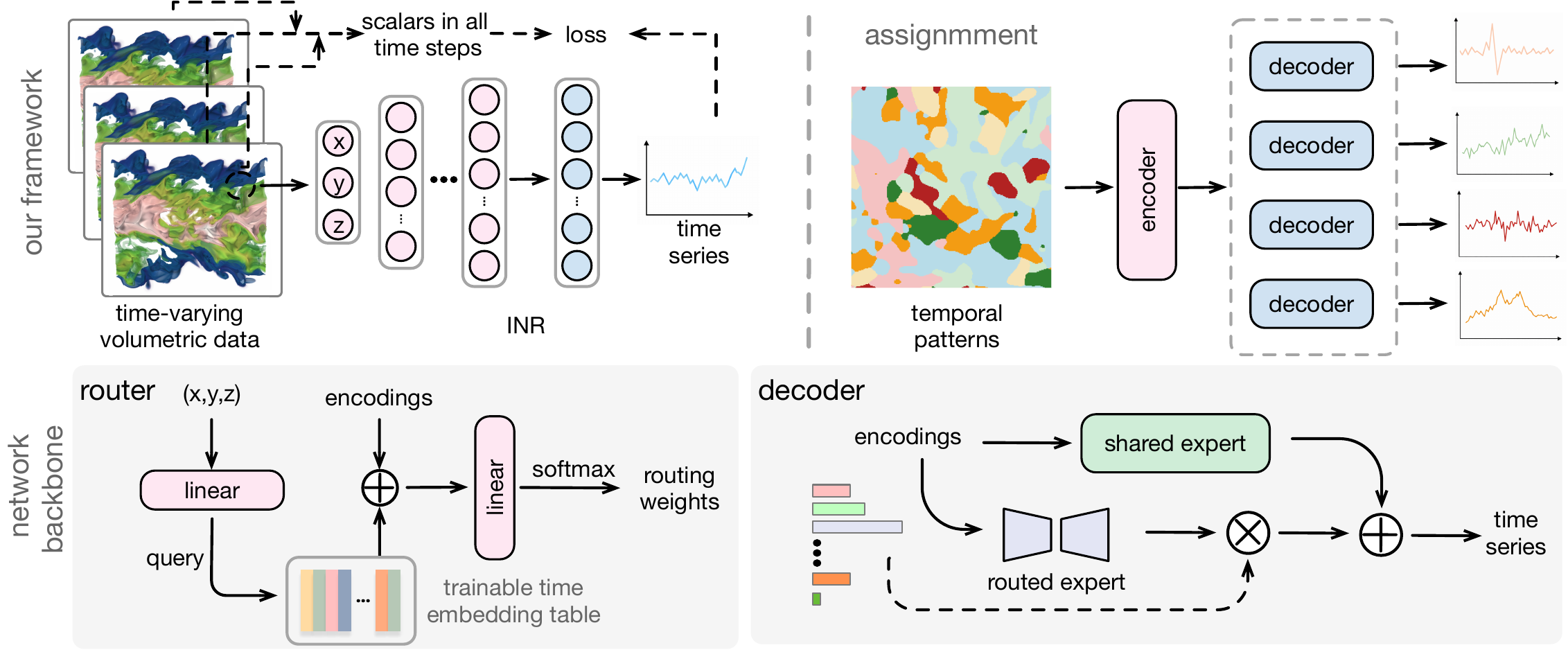}
    \caption{Overview of \hot{our framework}. \rev{Time-varying volumetric data are reformulated as spatially indexed temporal sequences, where each 3D coordinate is mapped to its corresponding scalar values across all time steps. This formulation is network-agnostic and can be instantiated with different INR backbones. Such a spatial assignment explicitly reveals heterogeneous temporal patterns across locations and motivates adaptive modeling. In this work, we instantiate the formulation with a MoE-based INR network, where routing guided by spatial features and time embeddings enables flexible expert assignment for sequence decoding.}}
    \label{fig:SequenceINR}
\end{figure*}

\subsection{Framework Design}

The upper-left of \cref{fig:SequenceINR} illustrates \hot{our framework} for compressing time-varying volumetric data. Given a sequence of volumetric fields with $T$ time steps, the dataset is conventionally represented as $\mathbf{V} = \{ \mathbf{V}^{(t)} \}_{t=1}^{T}$, where each $\mathbf{V}^{(t)}: \Omega \rightarrow \mathbb{R}$ is defined over a spatial domain $\Omega \subset \mathbb{R}^3$. Instead of modeling each time step independently, we reformulate the dataset as a mapping from spatial coordinates to temporal signals. Specifically, we define $\mathbf{F} : \Omega \rightarrow \mathbb{R}^{T}$ such that
$\mathbf{F}(\mathbf{x}) = [\mathbf{V}^{(1)}(\mathbf{x}), \mathbf{V}^{(2)}(\mathbf{x}), \dots, \mathbf{V}^{(T)}(\mathbf{x})]$
for any spatial location $\mathbf{x} \in \Omega$. Under this formulation, each voxel is associated with a temporal sequence, and the compression task reduces to learning the mapping $\mathbf{F}$ using a coordinate-based neural network.

This formulation is broadly applicable to existing coordinate-based INR architectures, as it shifts the learning target from point-wise scalars to sequence-level signals without modifying the underlying network design. In this work, we further exploit this perspective by introducing an MoE design to better capture spatial heterogeneity in temporal dynamics, as shown in the upper-right of \cref{fig:SequenceINR}.

\subsection{MoE-based Network Instantiation}

While the proposed sequence-level formulation enables structured modeling of temporal dynamics, it also introduces increased modeling complexity due to the diverse temporal patterns across different spatial locations. In practice, a single network often struggles to capture such heterogeneous behaviors effectively.
\rev{To address this issue, we instantiate the proposed formulation using a MoE-based INR architecture, which enables different experts to specialize in distinct temporal behaviors across spatial locations.} This design enables more flexible and specialized modeling while maintaining efficiency~\cite{zhao2023moec,han2025moeinr}.
As will be further demonstrated in~\cref{tab:learning_with_sequence}, such a formulation is particularly compatible with MoE-based models, which exhibit smaller performance degradation when extended to sequence-level prediction.

\textbf{Spatial Encoder.}
Since our model takes three-dimensional spatial coordinates as input, the encoder is responsible for mapping these low-dimensional inputs into a higher-dimensional feature space to enhance representational capacity. Directly feeding raw three-dimensional coordinates into a multilayer perceptron is insufficient for capturing complex spatial structures, especially in volumetric data where fine-scale variations and sharp transitions frequently occur. This limitation is largely attributed to the spectral bias of neural networks, which favors smooth, low-frequency components during optimization.
Following Han~et~al.~\cite{han2025dcinr}, we adopt a trainable positional encoding to enhance the representational capacity of spatial coordinates.
\hot{We choose this lightweight design due to its low parameter overhead, while still achieving competitive reconstruction performance in our setting (see Appendix~B).}
Given a spatial coordinate $\mathbf{x} \in \mathbb{R}^3$, we project the input coordinate into a higher-dimensional harmonic space using a learnable linear transformation $\mathbf{z} = \mathbf{w}\mathbf{x} + \mathbf{b}$, where $\mathbf{w} \in \mathbb{R}^{P \times 3}$ and $\mathbf{b} \in \mathbb{R}^{P}$ are trainable parameters, and $P$ denotes the embedding dimension. The positional encoding is then defined as

\begin{equation}
    \phi(\mathbf{x}) = \left[ \sin(2\pi \mathbf{z}), \cos(2\pi \mathbf{z}) \right].
\end{equation}

This learnable frequency mapping allows the network to adaptively capture spatial variations of different scales, providing a richer representation for subsequent modeling.

The positional encoding is further processed by multiple fully connected layers with sinusoidal activation functions to produce high-dimensional feature embeddings~\cite{sitzmann2020siren}. 
To improve optimization stability and facilitate feature propagation across layers, we incorporate residual connections between consecutive blocks. This residual harmonic encoding structure enhances representational capacity while maintaining stable training behavior.

\textbf{Router.}
Since temporal sequence patterns vary significantly across different spatial locations, a single shared decoder is insufficient to model all structural variations effectively. Therefore, we adopt the MoE architecture to enhance modeling flexibility within our framework.
Specifically, a Router is introduced to automatically assign spatial coordinates to different expert decoders. Each expert decoder is specialized in capturing one or more representative temporal patterns, enabling the overall model to adapt to diverse spatiotemporal structures while maintaining parameter efficiency.

Using raw spatial coordinates alone for expert routing is insufficient to distinguish complex structural variations. Han~et~al.~\cite{han2025moeinr} \rev{use standard Fourier mapping to encode spatiotemporal coordinates} before routing decisions are made, \hot{but this ignores the temporal patterns of sequences.}
\hot{
To this end, we introduce a trainable time embedding table \(\mathbf{E} \in \mathbb{R}^{T \times M}\), where \(T\) is the number of time steps and \(M\) is the hidden feature dimension. The table is optimized jointly with the INR network and provides a global temporal context for expert routing, as shown in the lower left corner at~\cref{fig:SequenceINR}. In the router, \(\mathbf{E}\) is projected to the routing feature space and summarized by mean and variance pooling along the temporal dimension to form a temporal descriptor. This temporal descriptor is fused with spatial routing features, so that the expert selection becomes aware of both spatial and temporal dynamics.}

The router outputs a set of routing scores, which are normalized by a softmax function to obtain a probability distribution over all expert decoders. The routing decision is based on the concatenated feature representation, combining the spatial encoding and temporal-aware features extracted by the router.
In practice, we adopt a hard routing strategy, ensuring sparse expert utilization.

\textbf{Decoder.}
Each expert in the decoder takes the high-dimensional feature representation generated by the spatial encoder as input and predicts the corresponding temporal sequence for the queried spatial coordinate. According to the routing decision of the router, only a single expert is activated under the hard routing mechanism to map the encoded features into the temporal output space.
Although all experts share the same network architecture, they are parameterized independently to specialize in different temporal patterns. Each expert is composed of fully connected layers with sinusoidal activation functions, followed by a final linear layer that outputs the predicted time-series values. However, since the output dimension of each expert corresponds to the full temporal sequence of length $T$, maintaining multiple independently parameterized experts would significantly increase the overall parameter count.

To address this issue, we incorporate a Low-Rank Adaptation (LoRA) mechanism~\cite{hu2022lora} into the expert decoders. Instead of learning completely independent weight matrices for each expert, we introduce a shared base decoder and model expert-specific variations through lightweight low-rank updates~\cite{han2025moeinr,zhang2025kimi}, as shown in~\cref{fig:SequenceINR}. This design substantially reduces the parameter footprint while preserving the specialization capability of individual experts. Let the shared decoder weight be denoted as
$\mathbf{W}_0 \in \mathbb{R}^{D \times T}$, where $D$ is the feature dimension and $T$ is the length of the temporal sequence. For the $k$-th expert, the effective weight matrix is defined as:
\begin{equation}
    \mathbf{W}_k = \mathbf{W}_0 + \mathbf{A}_k \mathbf{B}_k,
\end{equation}

\noindent where $\mathbf{A}_k \in \mathbb{R}^{D \times r}$, $\mathbf{B}_k \in \mathbb{R}^{r \times T}$, and $r$ is the rank of the low-rank adaptation. This formulation enables all experts to share the dominant transformation $\mathbf{W}_0$, while modeling expert-specific variations through lightweight low-rank updates, thereby increasing the compression ratio. Moreover, since the encoded features already provide a high-dimensional and expressive representation, the decoder does not require a deep architecture to reconstruct the temporal sequences.

\subsection{Optimization}
\label{optimization}
Proper initialization is critical for stabilizing INR optimization~\cite{sitzmann2020siren}, especially in MoE-based architectures where the router determines expert specialization. An improperly initialized policy network may lead to unstable expert assignments and slow convergence during early training. To address this issue, we introduce a warm-up strategy to guide the initial training of the routing mechanism.

Specifically, we first cluster the temporal sequences associated with all spatial coordinates. Each spatial location is assigned a cluster label according to the similarity of its corresponding time series. These cluster labels serve as pseudo ground-truth assignments for the router during the warm-up phase. The router is then trained using a cross-entropy loss to predict the assigned cluster for each coordinate. After this initialization stage, the entire network is jointly optimized without explicit clustering supervision.

After the warm-up phase, we optimize the framework in an end-to-end manner. During this stage, the routing assignments are no longer supervised by clustering labels but are determined by the learned router under the hard routing mechanism.
Given a spatial coordinate $\mathbf{x}$, the router selects a single expert $k$ according to
$k = \arg\max_i p_i(\mathbf{x})$, where $p_i(\mathbf{x})$ denotes the routing probability of the $i$-th expert. Only the selected expert decoder is activated to predict the temporal sequence $\mathbf{s}_{k}(\mathbf{x}) \in \mathbb{R}^{T}$.
The reconstruction loss is defined as the squared error between the predicted and ground-truth temporal sequences:
\begin{equation}
    \text{Loss}(\mathbf{x}) =
\left\|
\mathbf{s}_{k}(\mathbf{x}) -
\mathbf{s}(\mathbf{x})
\right\|_2^2,
\end{equation}

\noindent where $\mathbf{y}(\mathbf{x})$ is the ground-truth temporal sequence associated with coordinate $\mathbf{x}$. This hard routing strategy encourages expert specialization while maintaining computational efficiency.

\begin{table}[t]
\centering
\caption{The dimensions and size of each data set.}
\label{tab:dataset_dimensions}
\renewcommand{\arraystretch}{1.2}
\scriptsize
\resizebox{\columnwidth}{!}{
\begin{tabular}{l l c c}
\toprule
\textbf{Data Set} & \textbf{Variable} & \textbf{Dimension ($x \times y \times z \times t$)} & \textbf{Data Size (GB)} \\
\midrule
argon bubble   & intensity           & $640 \times 256 \times 256 \times 150$ & 23.44 \\
combustion     & CHI, YOH, MF        & $480 \times 720 \times 120 \times 100$ & 15.45 \\
ionization     & H2, H+, PD          & $600 \times 248 \times 248 \times 100$ & 13.74 \\
vortex         & vorticity           & $256 \times 256 \times 256 \times 90$  & 5.63  \\
Tangaroa         & vorticity magnitude           & $300 \times 180 \times 120 \times 150$  & 3.62  \\
\bottomrule
\end{tabular}
}
\end{table}
\begin{table*}[h]
\centering
\caption{Quantitative comparison of different learning-based INR models on various data set. 
CR denotes compression ratio, CT and DT indicate compression (in hours) and decompression time (in minutes). 
Best results are highlighted in \textbf{bold}.}
\vspace{2mm}
\resizebox{\linewidth}{!}{
\begin{tabular}{lccccc lcccccc}
\toprule
\rowcolor{gray!10}
\textbf{Data set} & \textbf{CR} & \textbf{Method} & \textbf{PSNR (dB)$\uparrow$} & \textbf{CT$\downarrow$} & \textbf{DT$\downarrow$} &
\textbf{Data set} & \textbf{CR} & \textbf{Method} & \textbf{PSNR (dB)$\uparrow$} & \textbf{CT$\downarrow$} & \textbf{DT$\downarrow$} \\
\midrule

\multirow{6}{*}{\centering\textbf{combustion (CHI)}} 
& \multirow{6}{*}{\centering 2,585}
& SIREN & 42.28 & 31.48 & 19.38 &
\multirow{6}{*}{\centering\textbf{ionization (PD)}} 
& \multirow{6}{*}{\centering 2,550}
& SIREN & 50.38 & 24.26 & 14.26 \\
& & NeurComp & 43.16 & 31.69 & 26.98 &
& & NeurComp & 50.81 & 24.16 & 21.33 \\
& & CoordNet & \textbf{44.42} & 29.47 & 20.85 &
& & CoordNet & 54.08 & 24.75 & 15.61 \\
& & Switch-NeRF & 41.19 & 29.45 & 34.53 &
& & Switch-NeRF & 52.59 & 29.02 & 26.26 \\
& & Neural Experts & 41.43 & 45.86 & 25.21 &
& & Neural Experts & 52.32 & 34.94 & 19.26 \\
& & MoE-INR & 43.68 & 40.82 & 35.74 &
& & MoE-INR & 53.31 & 35.52 & 25.91 \\
& & \textbf{\ours} & 43.85 & \textbf{0.64} & \textbf{0.93} &
& & \textbf{\ours} & \textbf{57.89} & \textbf{0.59} & \textbf{0.80} \\
\midrule

\multirow{6}{*}{\centering\textbf{argon bubble}} 
& \multirow{6}{*}{\centering 3,140}
& SIREN & 40.39 & 44.90 & 29.58 &
\multirow{6}{*}{\centering\textbf{vortex}} 
& \multirow{6}{*}{\centering 1,120}
& SIREN & 41.39 & 12.19 & 6.21 \\
& & NeurComp & 41.70 & 46.21 & 34.40 &
& & NeurComp & 44.31 & 14.56 & 6.99 \\
& & CoordNet & 42.83 & 45.18 & 30.98 &
& & CoordNet & 45.80 & 18.63 & 11.03 \\
& & Switch-NeRF & 41.05 & 50.66 & 32.88 &
& & Switch-NeRF & 39.53 & 11.15 & 9.85 \\
& & Neural Experts & 39.89 & 61.88 & 25.25 &
& & Neural Experts & 36.53 & 13.32 & 6.73 \\
& & MoE-INR & 42.25 & 68.97 & 27.76 &
& & MoE-INR & \textbf{52.38} & 14.70 & 6.86 \\
& & \textbf{\ours} & \textbf{43.15} & \textbf{0.78} & \textbf{0.72} & 
& & \textbf{\ours} & 48.18 & \textbf{0.54} & \textbf{0.27} \\
\bottomrule
\end{tabular}
}
\label{tab:learning_results}
\end{table*}

\begin{figure*}[!t]
  \centering
  \newcommand{\w}{0.19\textwidth}
  \newcommand{\wsmall}{0.095\textwidth}  
  \begin{subfigure}[t]{\w}\centering
    \includegraphics[width=\linewidth]{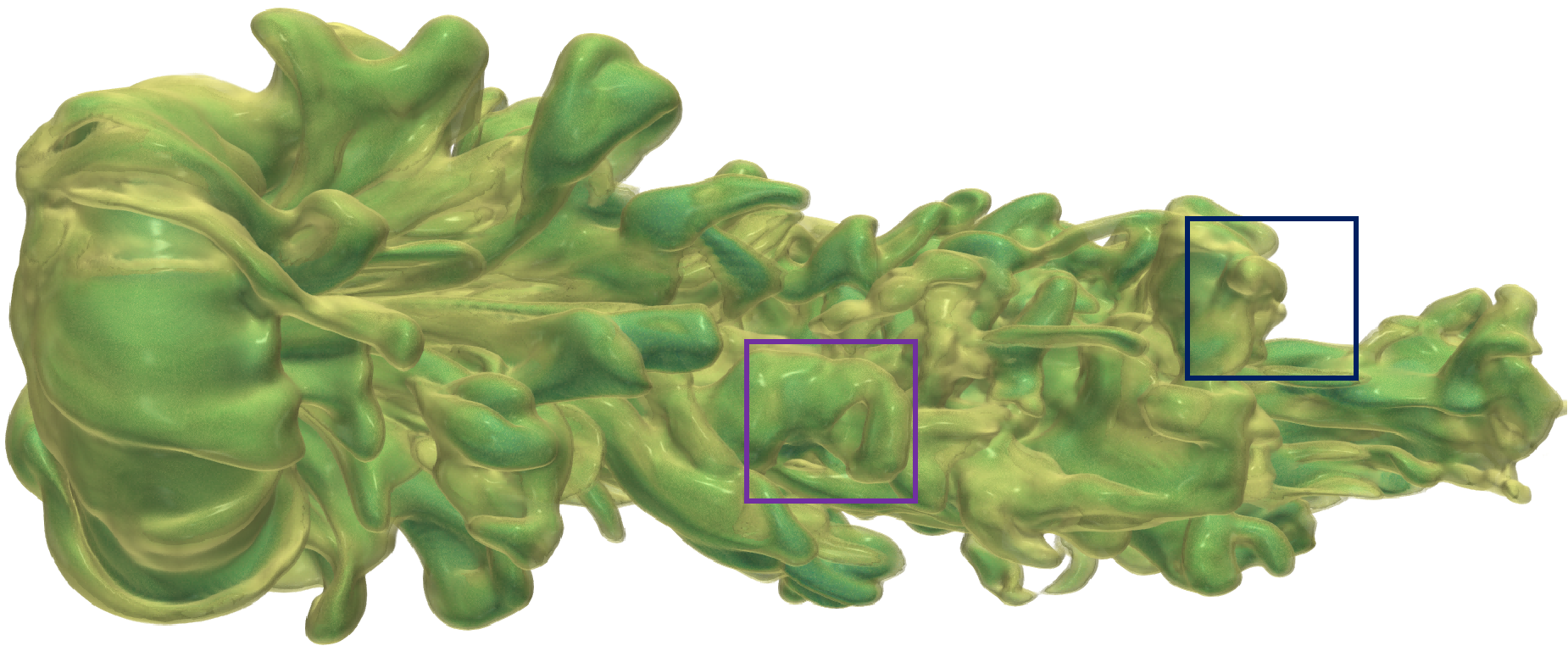}

  \end{subfigure}
  \begin{subfigure}[t]{\w}\centering
    \includegraphics[width=\linewidth]{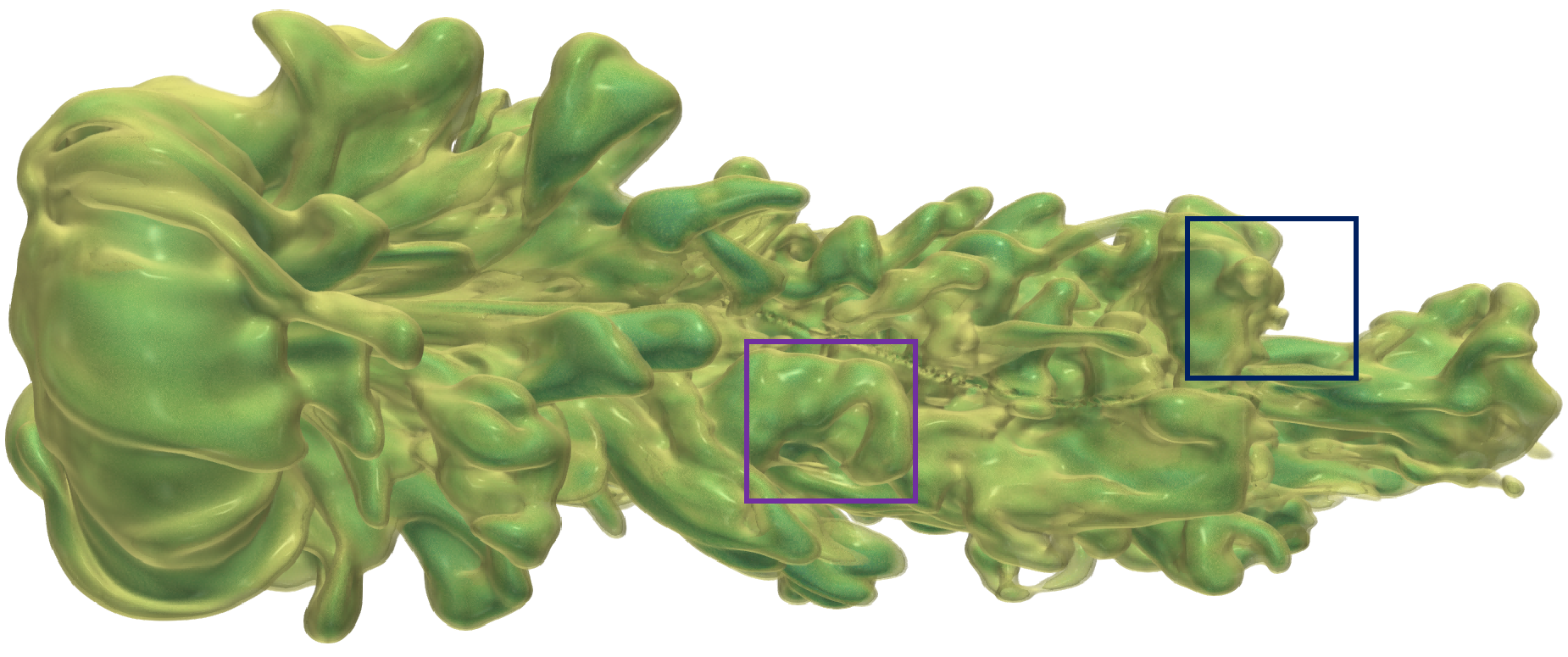}

  \end{subfigure}
  \begin{subfigure}[t]{\w}\centering
    \includegraphics[width=\linewidth]{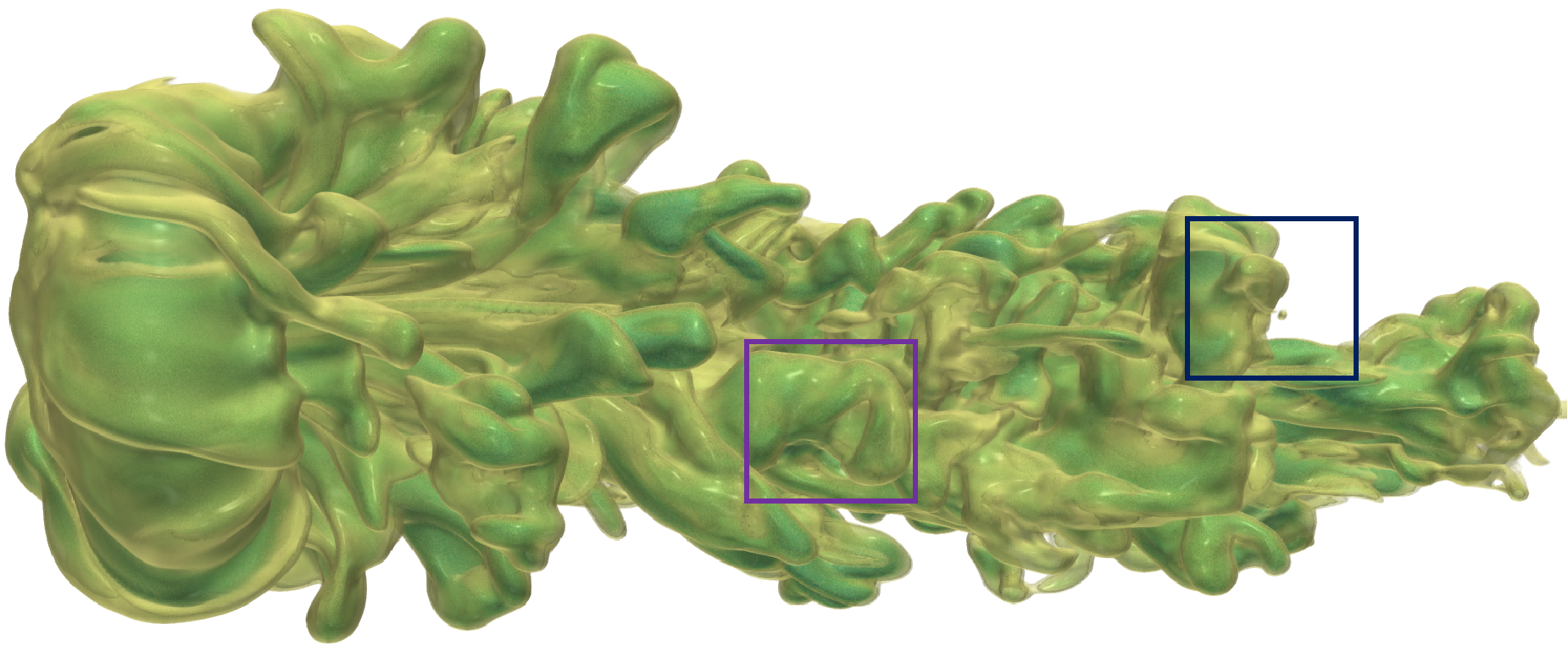}

  \end{subfigure}
  \begin{subfigure}[t]{\w}\centering
    \includegraphics[width=\linewidth]{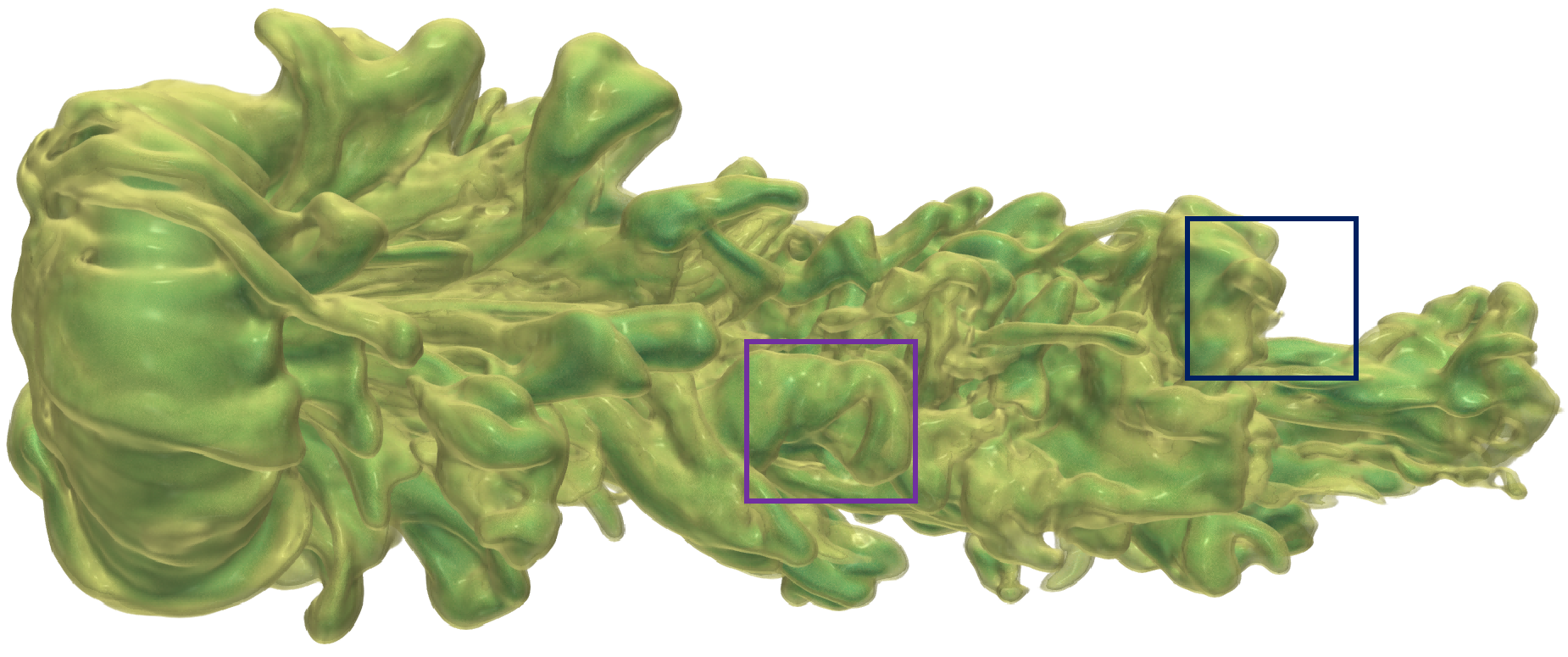}
    
  \end{subfigure}
  \begin{subfigure}[t]{\w}\centering
    \includegraphics[width=\linewidth]{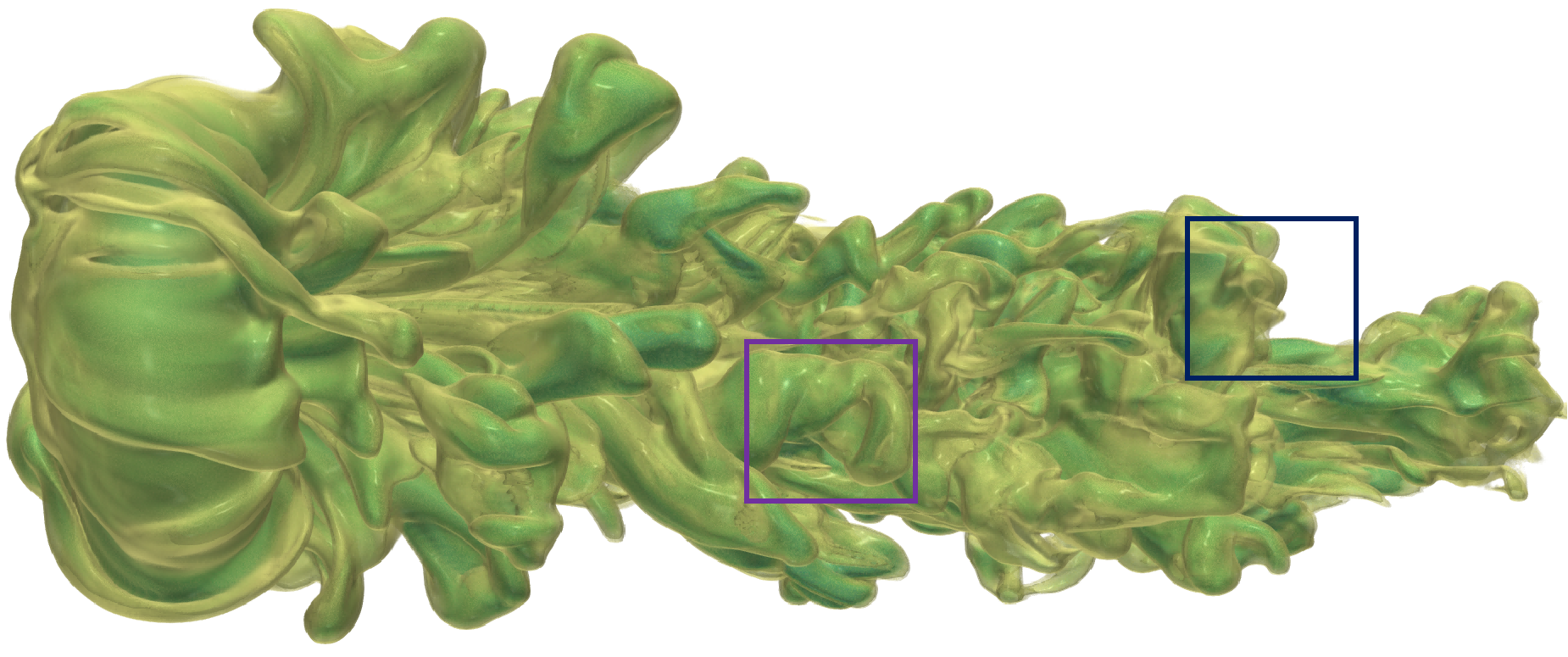}
  \end{subfigure}

\vspace{4pt}

  \begin{subfigure}[t]{\w}\centering
    \includegraphics[width=\linewidth]{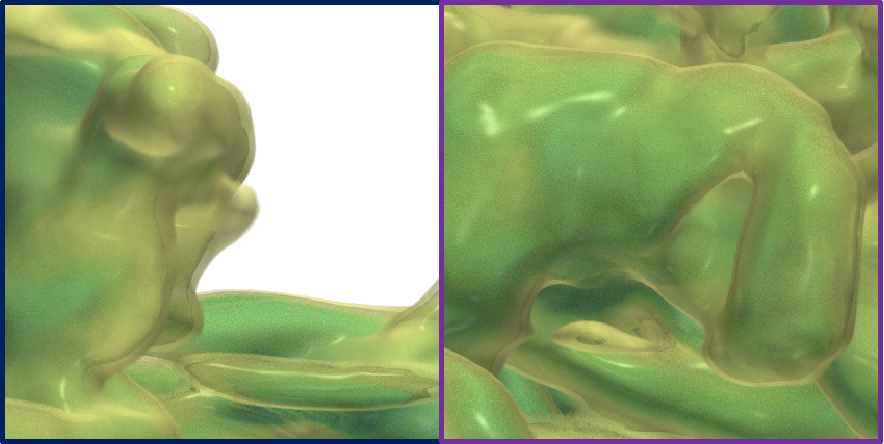}
    \caption{NeurComp}

  \end{subfigure}
  \begin{subfigure}[t]{\w}\centering
    \includegraphics[width=\linewidth]{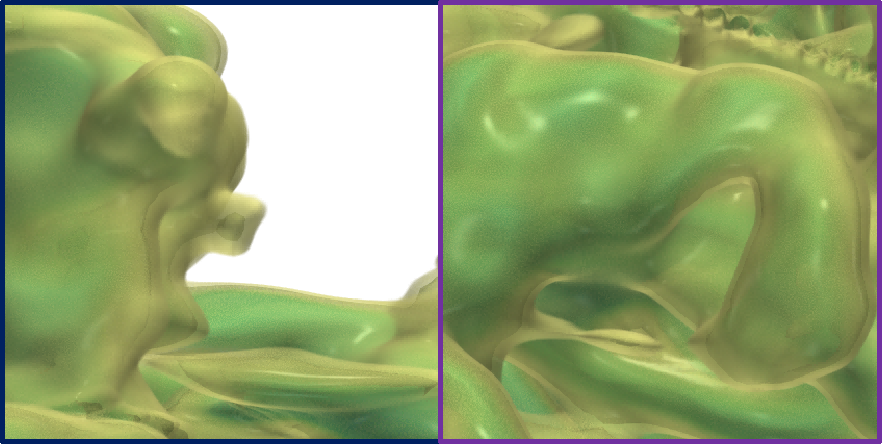}
    \caption{SIREN}

  \end{subfigure}
  \begin{subfigure}[t]{\w}\centering
    \includegraphics[width=\linewidth]{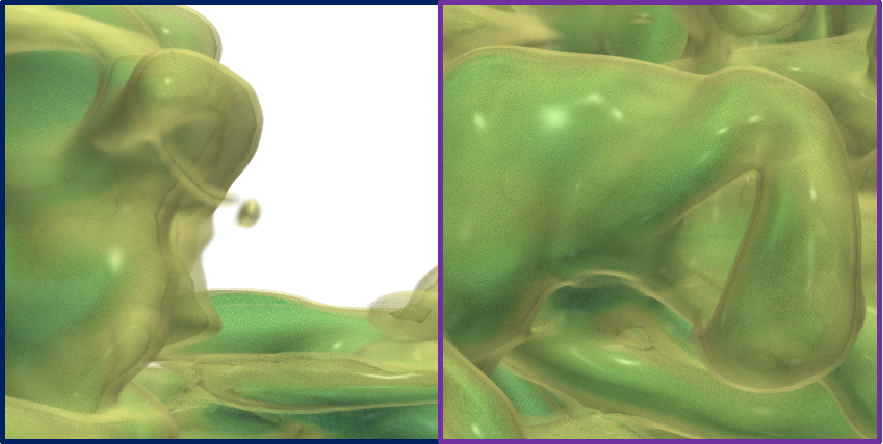}
    \caption{CoordNet}

  \end{subfigure}
  \begin{subfigure}[t]{\w}\centering
    \includegraphics[width=\linewidth]{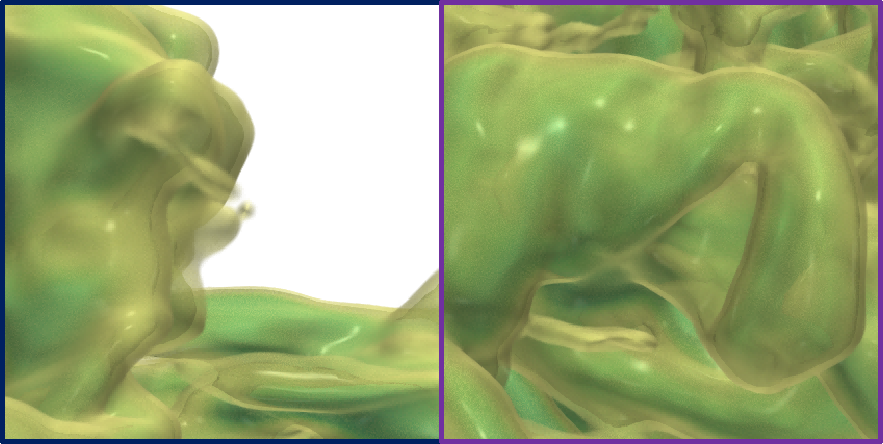}
    \caption{\ours}

  \end{subfigure}
  \begin{subfigure}[t]{\w}\centering
    \includegraphics[width=\linewidth]{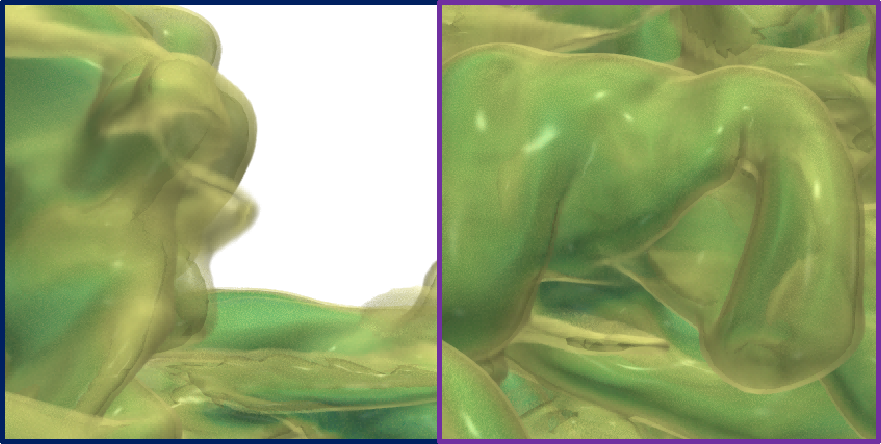}
    \caption{GT}
 
  \end{subfigure}

  \caption{Comparison of volume rendering results on learning-based INR methods using the argon bubble data set, and the CR is 3,140.}
  \label{fig:bubble_learning_render}
\end{figure*}

\begin{table}[!t]
\centering
\caption{Quantitative comparison of visual reconstruction performance across different INR models. 
CR denotes compression ratio. HD measures surface distance. LPIPS and DreamSim measure perceptual similarity. Lower values are better.
The chosen isovalue is 0.2, -0.2, -0.45 and 0, respectively.}
\vspace{2mm}
\resizebox{\linewidth}{!}{
\begin{tabular}{lcccccc}
\toprule
\rowcolor{gray!10}
\textbf{Data set} & \textbf{CR} & \textbf{Method} & \textbf{HD$\downarrow$} & \textbf{LPIPS$\downarrow$} & \textbf{DreamSim$\downarrow$} \\
\midrule

\multirow{6}{*}{\centering\textbf{combustion (CHI)}} 
& \multirow{6}{*}{\centering 2,585}
& SIREN & 100.05 & 0.065 & 0.089 \\
& & NeurComp & \textbf{40.47} & 0.058 & 0.059 \\
& & CoordNet & 100.46 & 0.033 & 0.061 \\
& & Switch-NeRF & 88.94 & 0.098 & 0.088 \\
& & Neural Experts & 94.31 & 0.064 & 0.098 \\
& & MoE-INR & 41.38 & \textbf{0.032} & 0.039 \\
& & \textbf{\ours} & 87.26 & 0.045 & \textbf{0.037} \\
\midrule

\multirow{6}{*}{\centering\textbf{ionization (PD)}} 
& \multirow{6}{*}{\centering 2,550}
& SIREN & 40.79 & 0.087 & 0.136 \\
& & NeurComp & 40.31 & 0.021 & 0.056 \\
& & CoordNet & 40.68 & 0.046 & 0.029 \\
& & Switch-NeRF & 86.49 & 0.068 & 0.021 \\
& & Neural Experts & 34.61 & 0.093 & 0.035 \\
& & MoE-INR & 37.31 & 0.073 & 0.026 \\
& & \textbf{\ours} & \textbf{34.37} & \textbf{0.010} & \textbf{0.013} \\
\midrule

\multirow{6}{*}{\centering\textbf{argon bubble}} 
& \multirow{6}{*}{\centering 3,140}
& SIREN & 36.23 & 0.096 & 0.071 \\
& & NeurComp & 17.32 & 0.087 & 0.066 \\
& & CoordNet & 15.19 & 0.067 & 0.088 \\
& & Switch-NeRF & 18.71 & 0.078 & 0.089 \\
& & Neural Experts & 31.33 & 0.093 & 0.109 \\
& & MoE-INR & \textbf{11.53} & \textbf{0.021} & \textbf{0.045} \\
& & \textbf{\ours} & 13.45 & 0.076 & 0.076 \\
\midrule

\multirow{6}{*}{\centering\textbf{vortex}} 
& \multirow{6}{*}{\centering 1,120}
& SIREN & 39.60 & 0.223 & 0.213 \\
& & NeurComp & 29.06 & 0.032 & 0.061 \\
& & CoordNet & 20.46 & 0.048 & \textbf{0.025} \\
& & Switch-NeRF & 56.32 & 0.151 & 0.125 \\
& & Neural Experts & 73.27 & 0.295 & 0.312 \\
& & MoE-INR & 22.31 & \textbf{0.012} & 0.034 \\
& & \textbf{\ours} & \textbf{17.56} & 0.030 & 0.079 \\

\bottomrule
\end{tabular}
}
\label{tab:surface_metrics}
\end{table}

\section{Results}

In this section, we describe the configurations of evaluation metrics and baselines, and present both quantitative and qualitative results to assess the performance from different perspectives.
\cref{tab:dataset_dimensions} summarizes the datasets used in our evaluation.

\subsection{Configurations}

\textbf{Evaluation metrics.}
The difference between the reconstruction volumes and the ground truth (GT) volumes is assessed from multiple perspectives, including reconstruction quality and perceptual similarity.
We use voxel-level metrics to evaluate numerical fidelity with respect to the ground truth. These include the Peak Signal-to-Noise Ratio (PSNR), which measures pixel-wise intensity differences. We also include the Learned Perceptual Image Patch Similarity (LPIPS)~\cite{zhang2018lpips}, which correlates well with human perception by computing distances between deep features extracted from a pretrained AlexNet~\cite{krizhevsky2012alexnet}.
Beyond low-level fidelity, we evaluate the perceptual and structural realism of the reconstruction using high-level metrics. DreamSim~\cite{fu2023dreamsim} calculates the cosine distance between features extracted by an ensemble of vision-language models, including CLIP~\cite{radford2021clip}, and has been shown to better align with human perceptual judgments. At the geometric level, we compute the Hausdorff Distance (HD) between the isosurfaces extracted from the reconstructed and ground-truth volumes. HD reflects the maximum surface deviation.

\textbf{Baselines.}
We compare our framework with both open-source learning-based INR models and traditional lossy compression techniques. The evaluation includes two major categories: neural implicit representations and conventional error-bounded compressors.
For neural baselines, we consider representative INR architectures, including CoordNet~\cite{han2022coordnet}, NeurComp~\cite{lu2021neurcomp}, and SIREN~\cite{sitzmann2020siren}, which employ coordinate-based MLPs for volumetric representation. We also include mixture-of-experts INRs, such as MoE-INR~\cite{han2025moeinr}, Switch-NeRF~\cite{zhenxing2022switchnerf} and Neural Experts~\cite{ben2024neuralexperts}. Although originally designed for scene rendering, image, or mesh representation, we adapt these MoE-based INRs to the time-varying volumetric compression setting to enable a comparison of MoE modeling strategies.
For traditional lossy compressors, we select three typical solutions: ZFP~\cite{lindstrom2014zfp}, SZ3~\cite{liang2022sz3}, and TTHRESH~\cite{ballester2019tthresh}. These methods represent widely adopted error-bounded compression approaches for scientific volumetric data and serve as strong non-learning baselines in our evaluation.
All learning-based methods are trained under the same optimizer and learning rate schedule for comparison. 
\hot{To evaluate the performance of our method against conventional INRs, all baseline methods are uniformly configured under the scalar prediction setting. For a fair comparison, all methods use the same voxel coordinate sampling strategy and sampling ratio during training. 
To isolate the effect of the proposed sequence-based formulation, we further evaluate the network backbones from learning-based baselines under the sequence prediction setting. The corresponding results and discussion are provided in~\cref{subsec:analysis}.}
For rendering evaluation, identical view parameters, transfer functions, and lighting settings are applied across all methods.

\begin{figure*}[!t]
  \centering
  \newcommand{\w}{0.19\textwidth}
  \newcommand{\wsmall}{0.095\textwidth}  
  \begin{subfigure}[t]{\w}\centering
    \includegraphics[width=\linewidth]{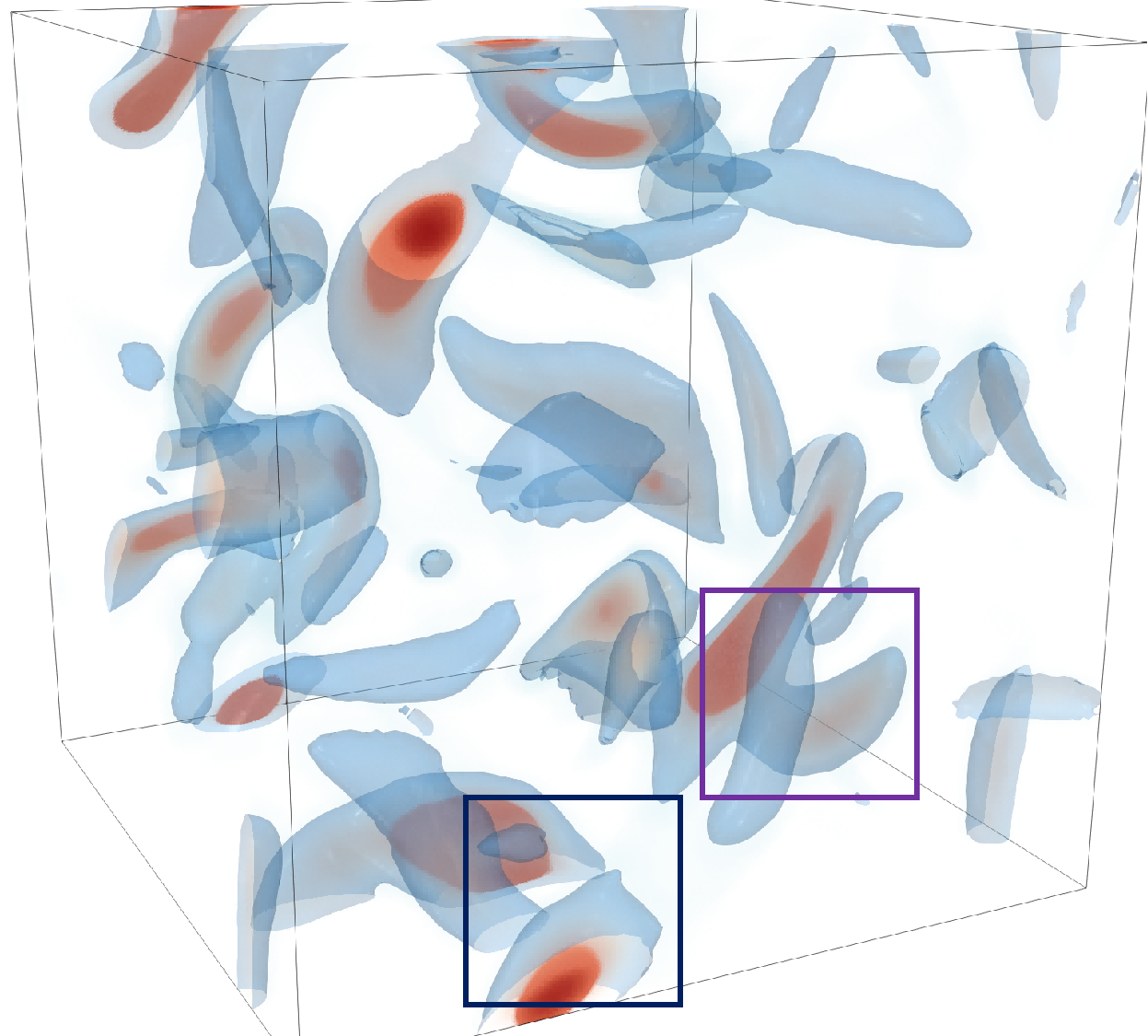}

  \end{subfigure}
  \begin{subfigure}[t]{\w}\centering
    \includegraphics[width=\linewidth]{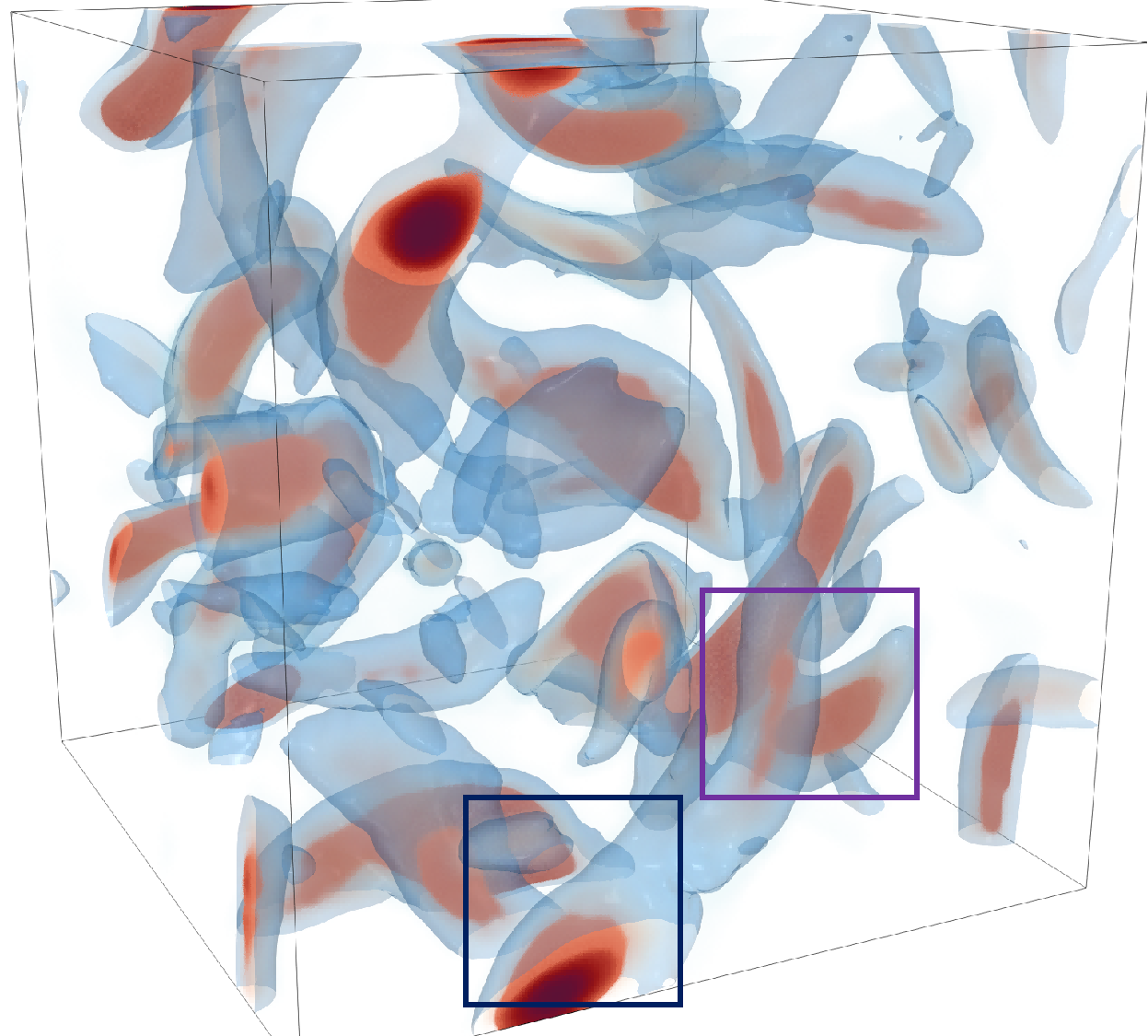}

  \end{subfigure}
  \begin{subfigure}[t]{\w}\centering
    \includegraphics[width=\linewidth]{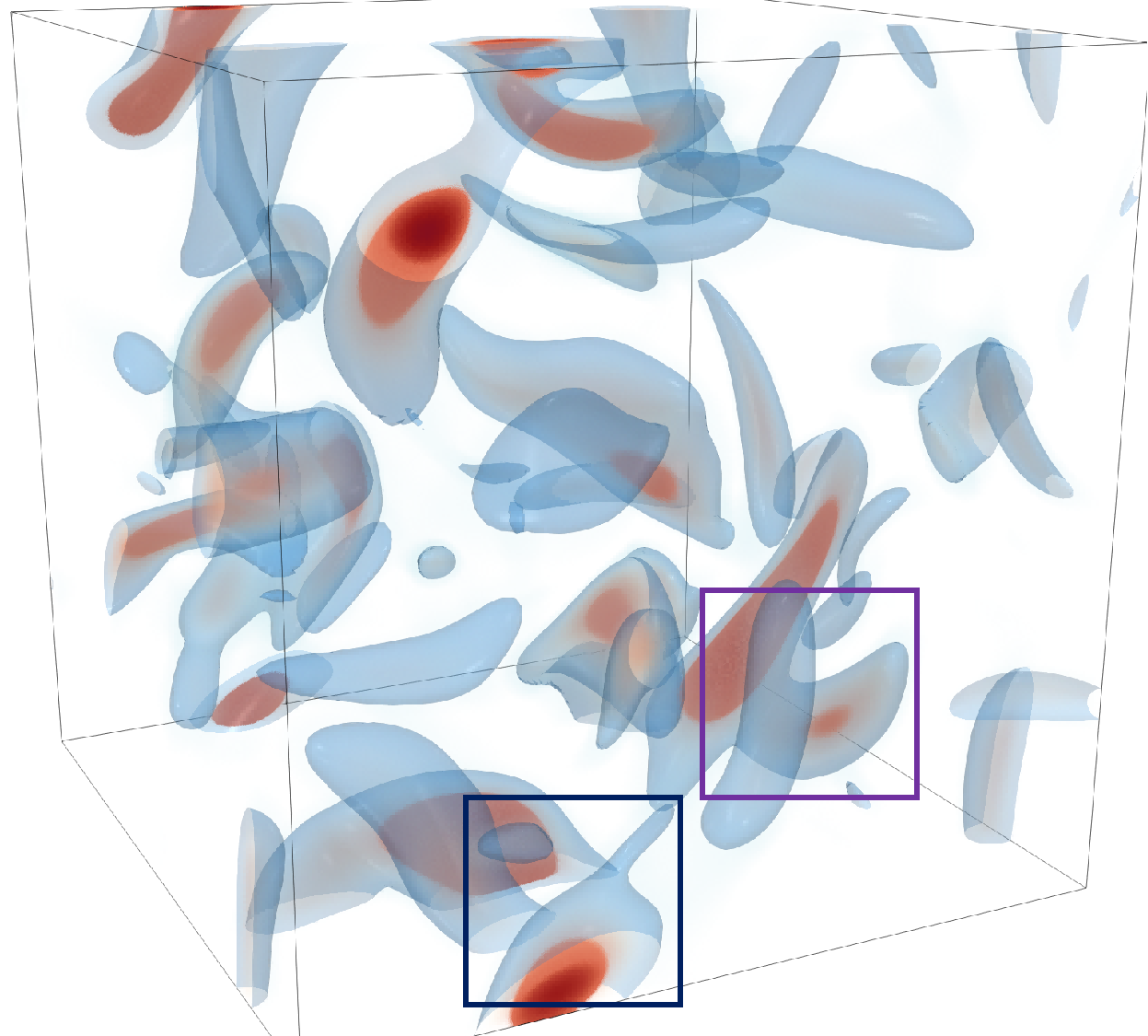}

  \end{subfigure}
  \begin{subfigure}[t]{\w}\centering
    \includegraphics[width=\linewidth]{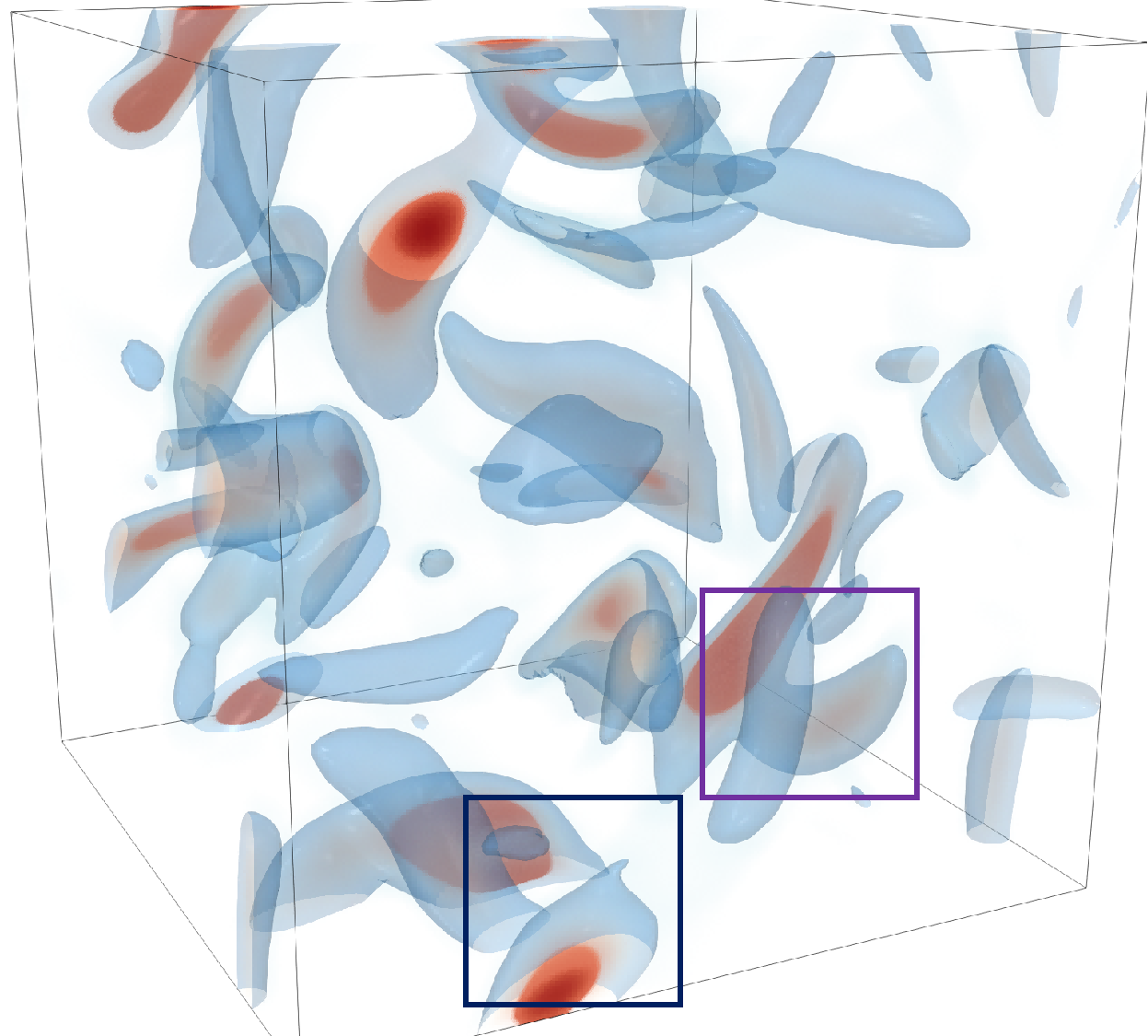}

  \end{subfigure}
  \begin{subfigure}[t]{\w}\centering
    \includegraphics[width=\linewidth]{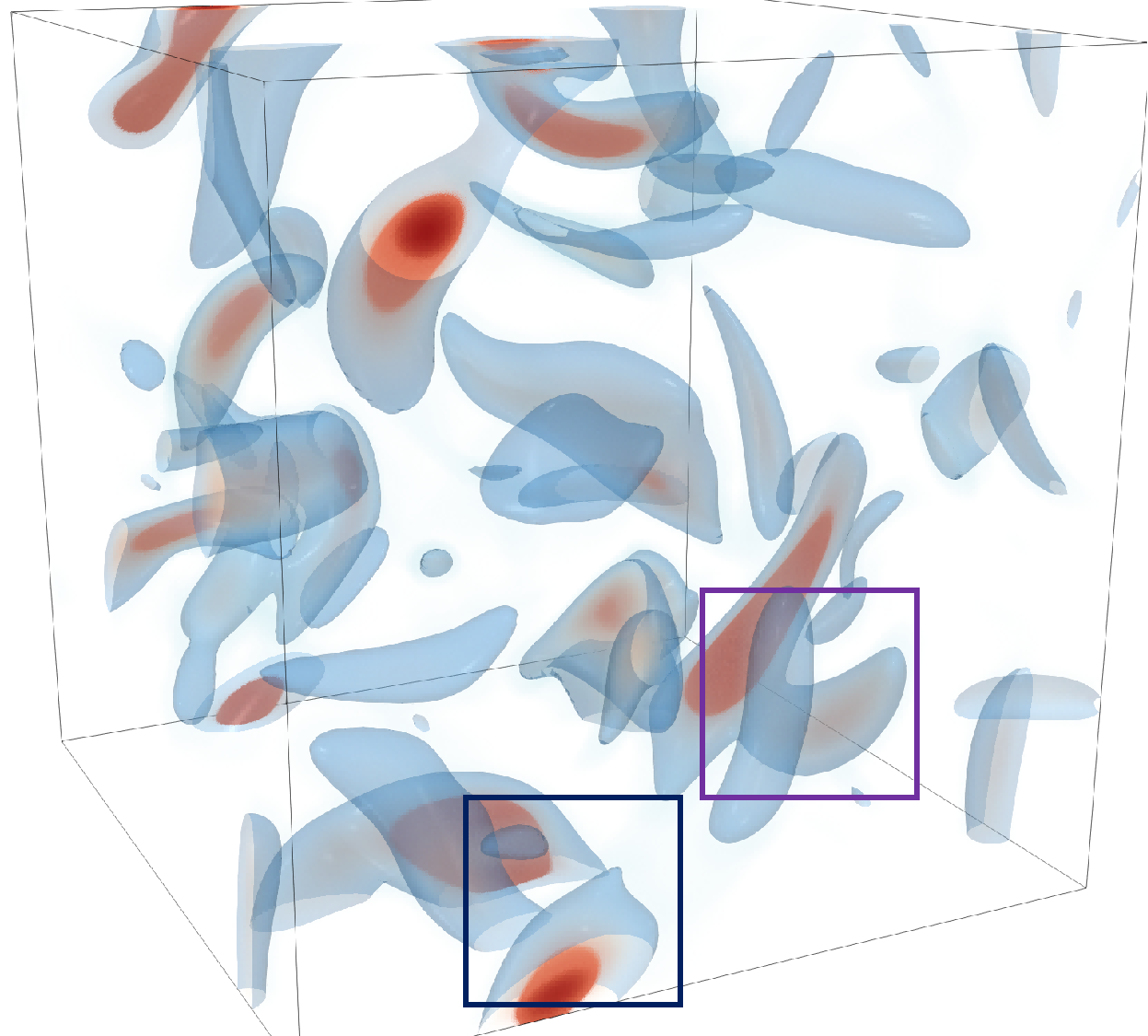}
  \end{subfigure}

\vspace{4pt}

  \begin{subfigure}[t]{\w}\centering
    \includegraphics[width=\linewidth]{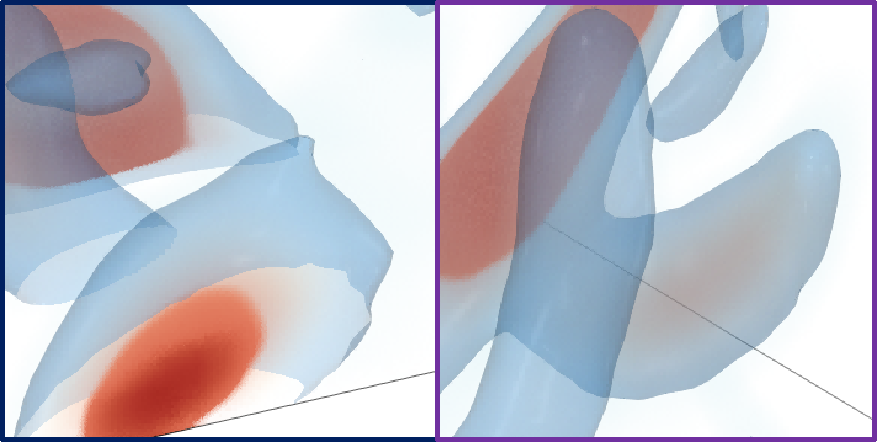}
    \caption{NeurComp}
  \end{subfigure}
  \begin{subfigure}[t]{\w}\centering
    \includegraphics[width=\linewidth]{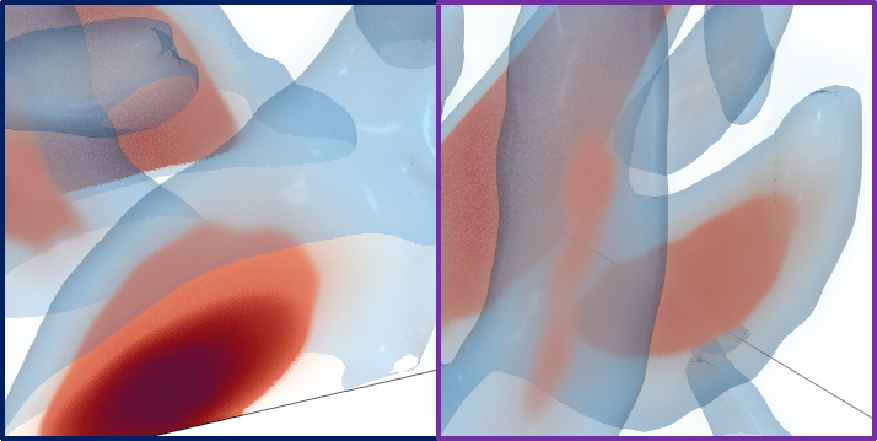}
    \caption{SIREN}
  \end{subfigure}
  \begin{subfigure}[t]{\w}\centering
    \includegraphics[width=\linewidth]{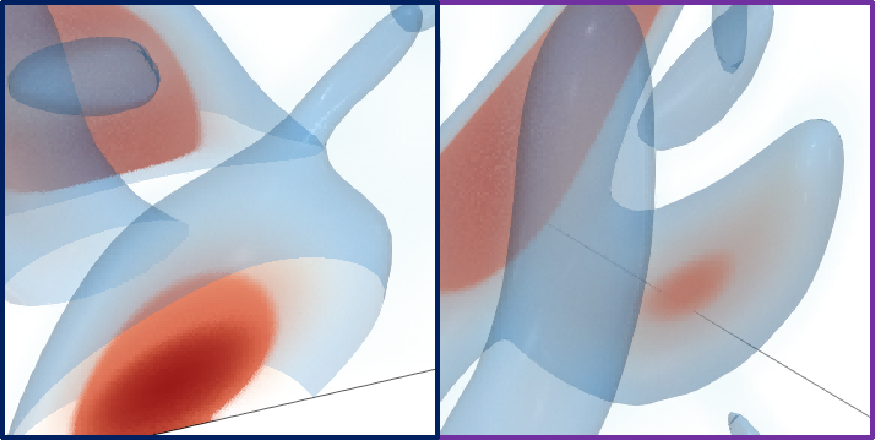}
    \caption{CoordNet}
  \end{subfigure}
  \begin{subfigure}[t]{\w}\centering
    \includegraphics[width=\linewidth]{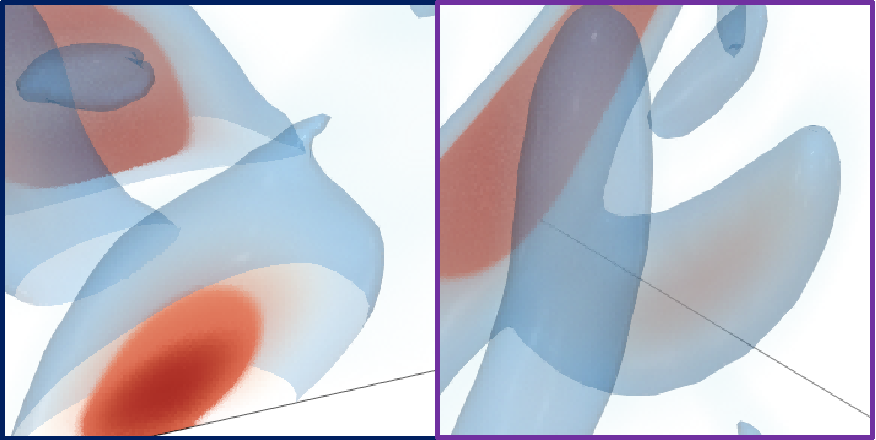}
    \caption{\ours}
  \end{subfigure}
  \begin{subfigure}[t]{\w}\centering
    \includegraphics[width=\linewidth]{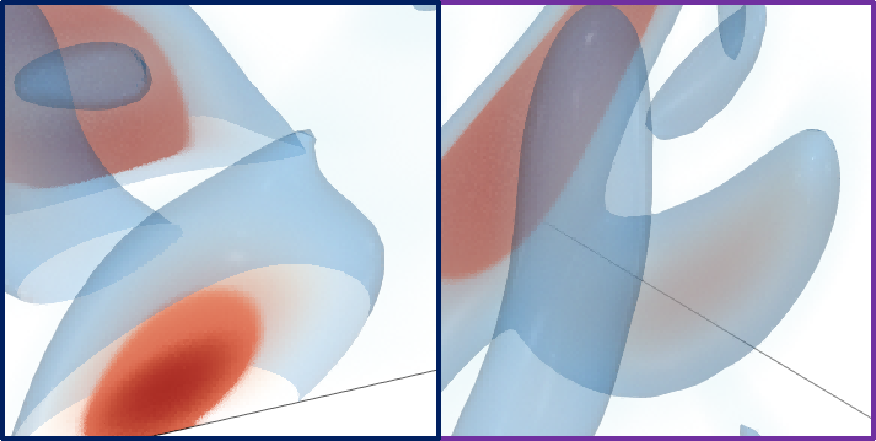}
    \caption{GT}
  \end{subfigure}

  \caption{Comparison of volume rendering results on learning-based INR methods using the vortex data set, and the CR is 1,120.}
  \label{fig:vortex_learning_render}
\end{figure*}

\subsection{Comparison Against Learning-based Methods}

We compare our framework with representative learning-based INR models across multiple time-varying volumetric datasets. The quantitative reconstruction results are summarized in~\cref{tab:learning_results} and \cref{tab:surface_metrics}, covering reconstruction performance, compression ratio, compression time and decompression time.

\textbf{Quantitative comparison.}
As shown in~\cref{tab:learning_results}, our framework is significantly faster across all datasets over learning-based INR baselines, with \textbf{compression speedups ranging from 42.6$\times$ to 59.9$\times$ and decompression speedups from 25.7$\times$ to 38.0$\times$}, corresponding to average improvements of approximately 49.2$\times$ and 31.5$\times$, respectively. This efficiency gain mainly comes from our formulation. 
\hot{Under the same compression ratio settings, our framework achieves the highest PSNR on the ionization (PD) and argon bubble datasets, and the second best PSNR on the other two datasets. 
While the gap to CoordNet on combustion (CHI) is marginal, MoE-INR achieves a larger improvement on vortex, indicating that coordinate-wise MoE models can be advantageous for datasets with complex temporal dynamics. This behavior is related to temporal coherence: full-sequence prediction is more challenging when temporal trajectories are high-frequency and weakly coherent. Detailed per-time-step results and temporal-complexity analysis are provided in Appendix~C.
}
On the ionization (PD) dataset, our framework consistently achieves superior performance across multiple evaluation metrics. This can be attributed to the active reaction regions occupying relatively small areas, while the majority of the domain exhibits more homogeneous and stable temporal patterns. 

\hot{
As listed in~\cref{tab:surface_metrics}, in terms of iso-surface distance (HD), our method achieves the best results on the ionization (PD) and vortex datasets, and remains competitive on the other dataset, i.e, second on argon bubble and third on combustion (CHI). These results indicate that our method provides competitive geometric reconstruction quality. For the two perceptual metrics, our method remains competitive across datasets, achieving the best performance in three out of eight cases and maintaining comparable results in the remaining cases. Although scalar-based representations may provide advantages for certain perceptual measurements, the proposed sequence-level formulation achieves a favorable balance between reconstruction quality and efficiency, with substantial reductions in compression and decompression time.
}

\begin{figure*}[t]
  \centering
  \newcommand{\w}{0.19\textwidth}
  \newcommand{\wsmall}{0.095\textwidth}  
  \begin{subfigure}[t]{\w}\centering
    \includegraphics[width=\linewidth]{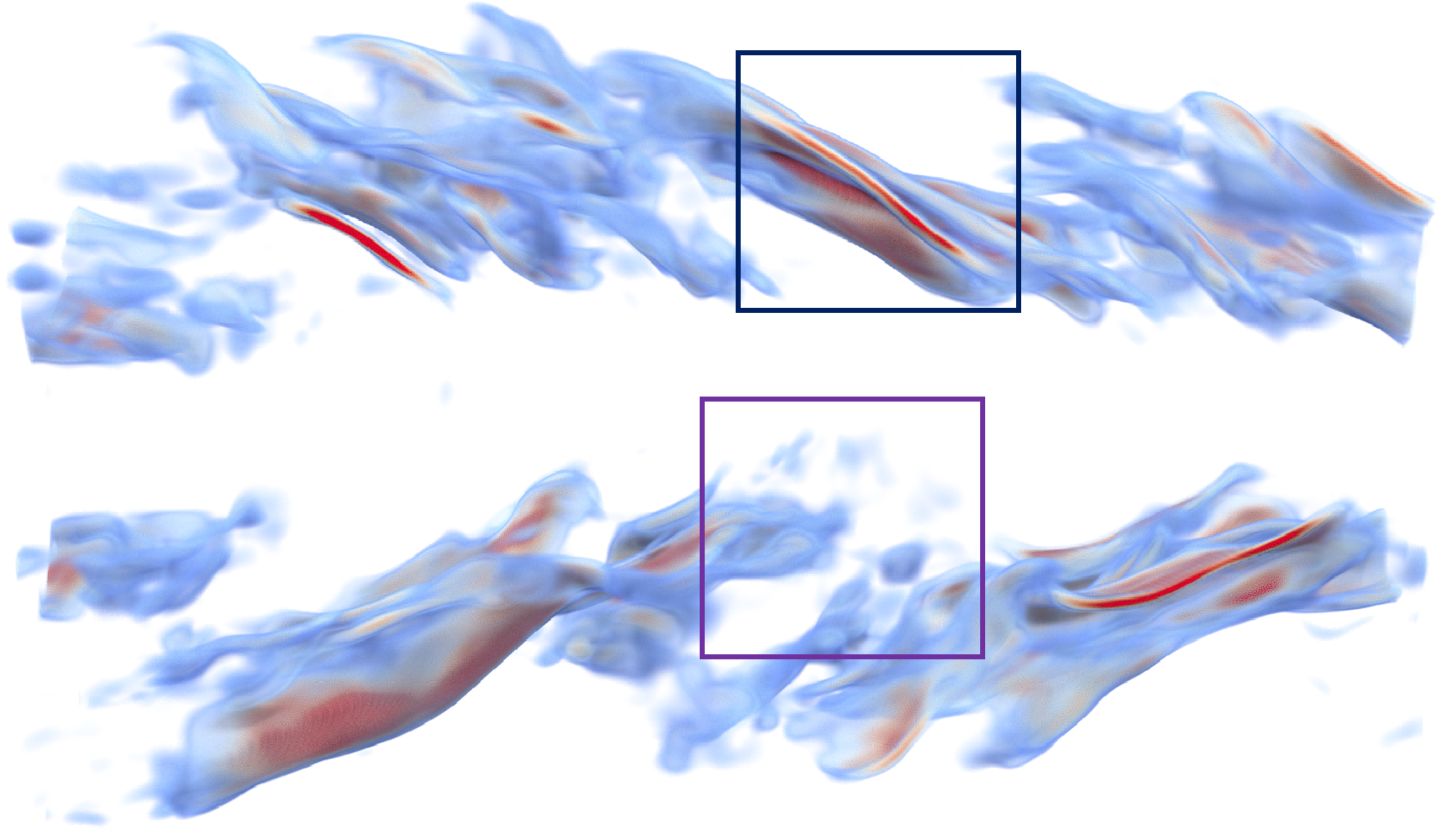}

  \end{subfigure}
  \begin{subfigure}[t]{\w}\centering
    \includegraphics[width=\linewidth]{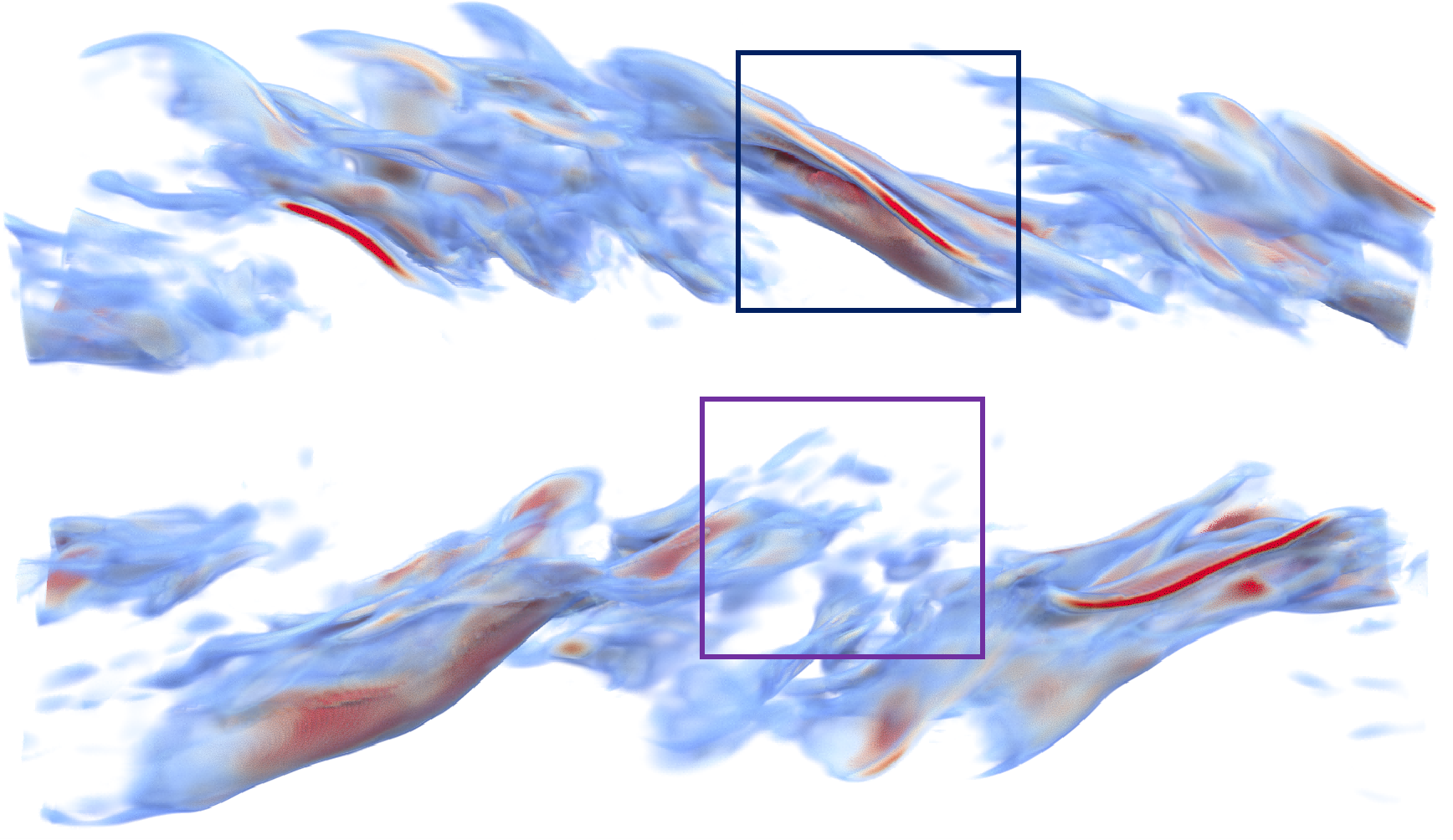}

  \end{subfigure}
  \begin{subfigure}[t]{\w}\centering
    \includegraphics[width=\linewidth]{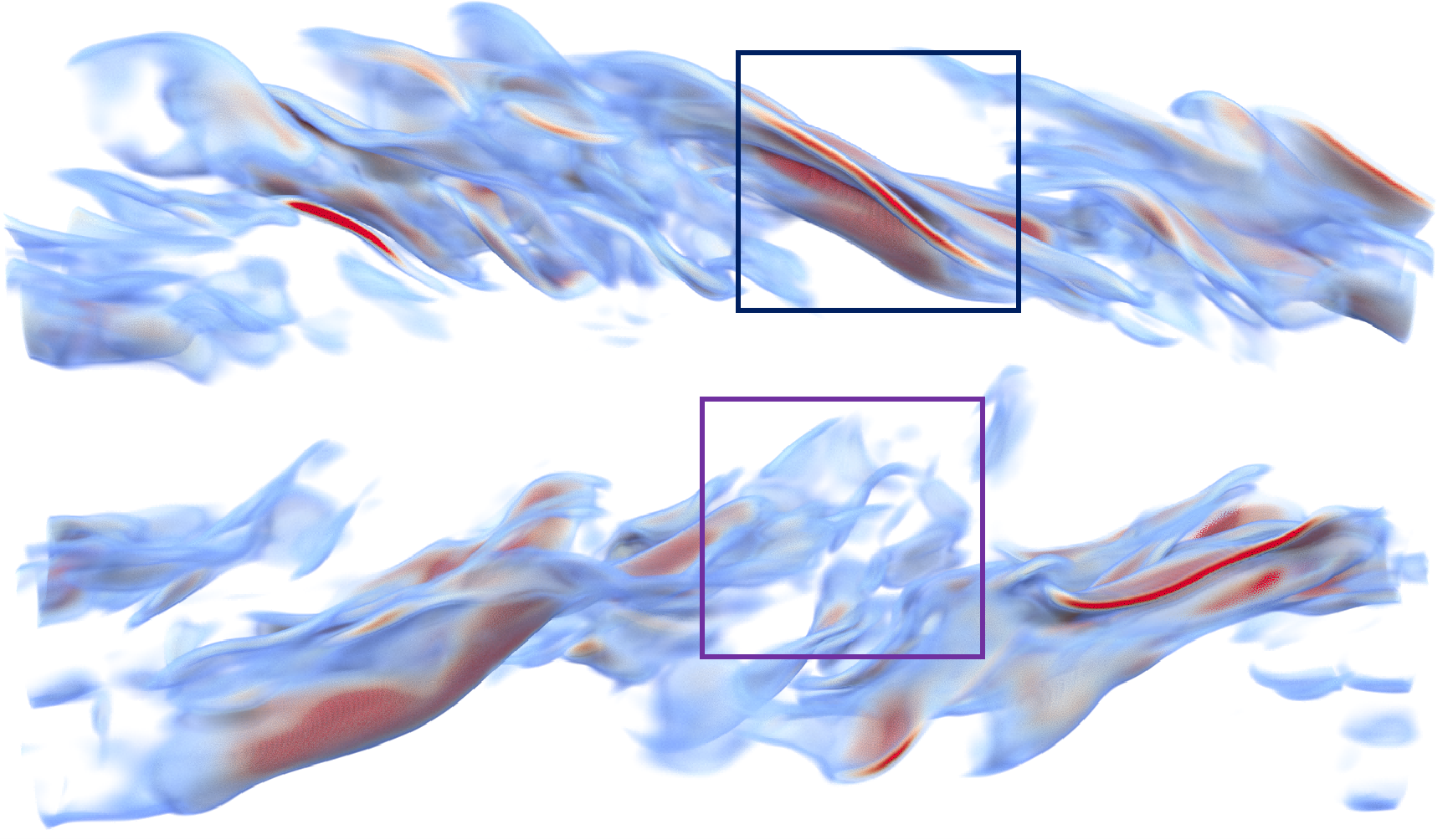}

  \end{subfigure}
  \begin{subfigure}[t]{\w}\centering
    \includegraphics[width=\linewidth]{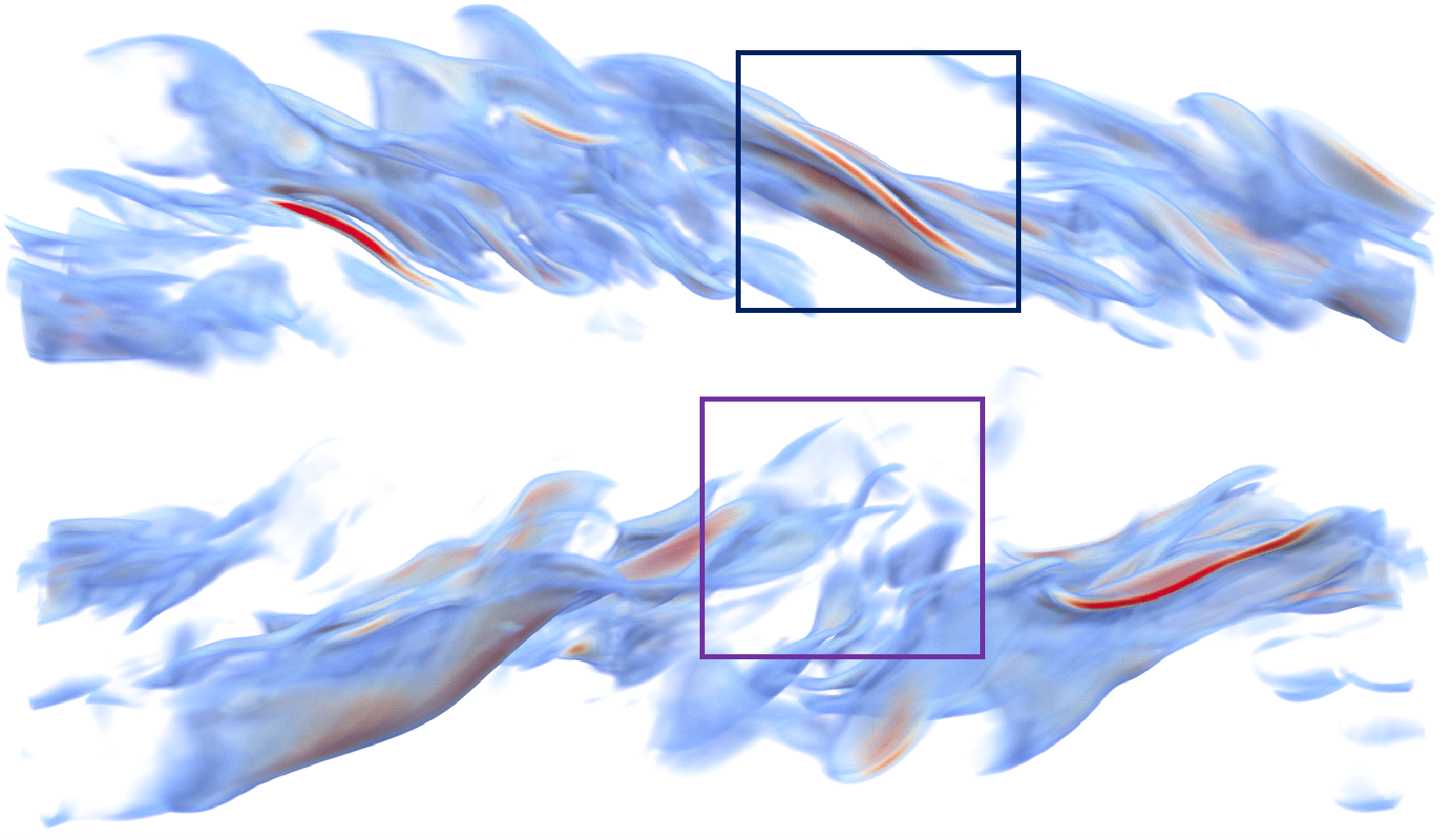}

  \end{subfigure}
  \begin{subfigure}[t]{\w}\centering
    \includegraphics[width=\linewidth]{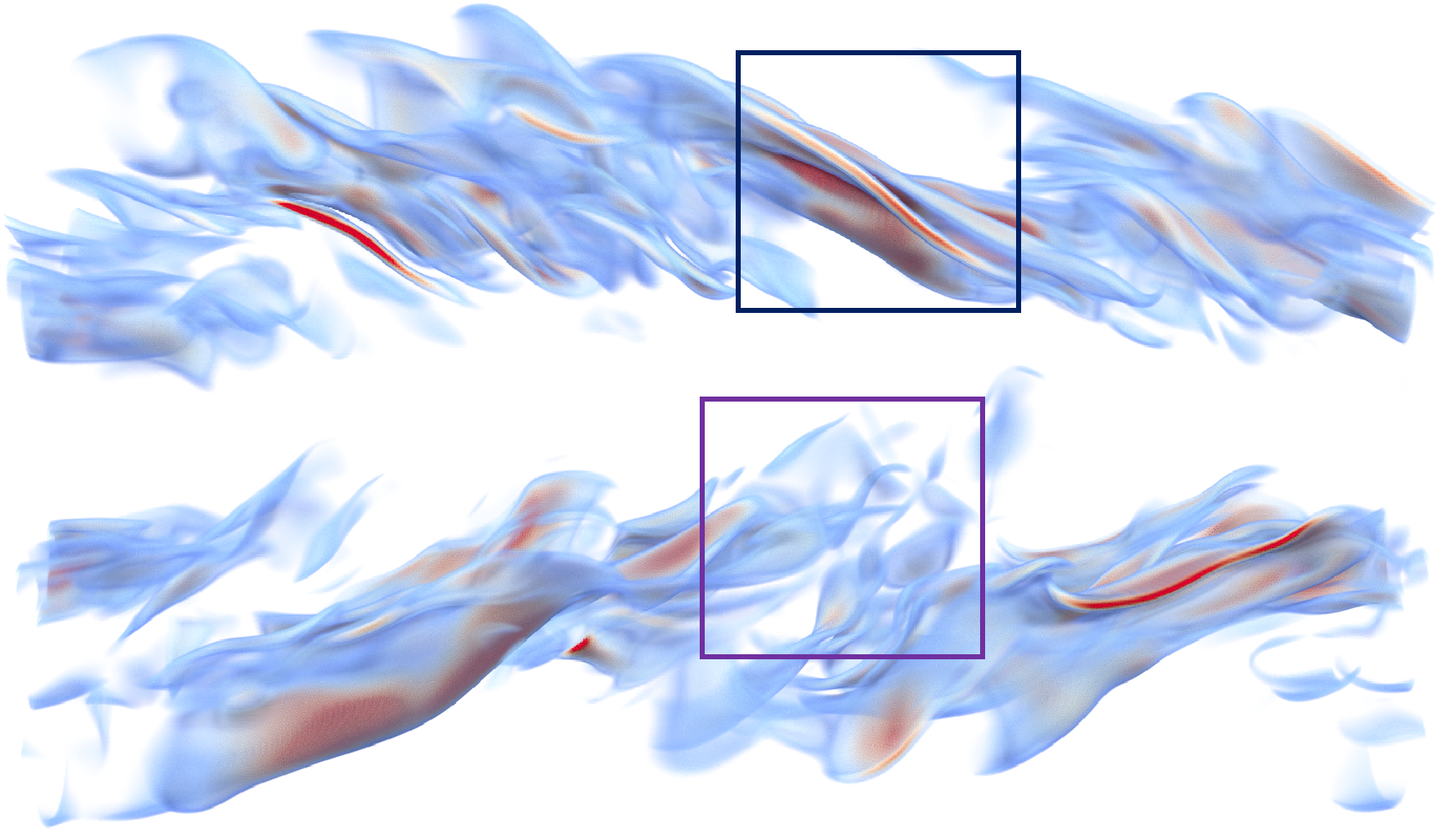}

  \end{subfigure}

\vspace{4pt}

  \begin{subfigure}[t]{\w}\centering
    \includegraphics[width=\linewidth]{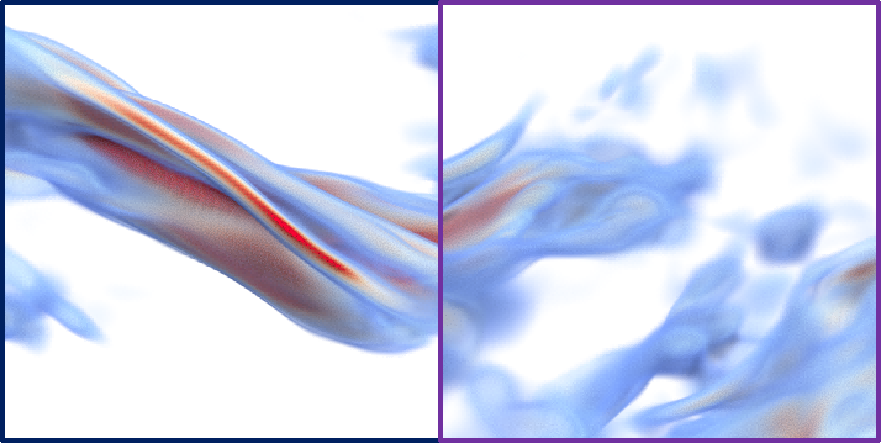}
    \caption{Neural Experts}
  \end{subfigure}
  \begin{subfigure}[t]{\w}\centering
    \includegraphics[width=\linewidth]{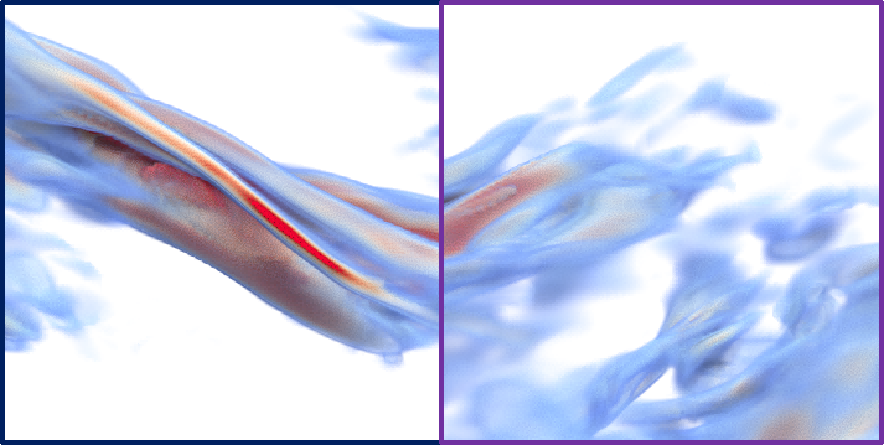}
    \caption{Switch-NeRF}
  \end{subfigure}
  \begin{subfigure}[t]{\w}\centering
    \includegraphics[width=\linewidth]{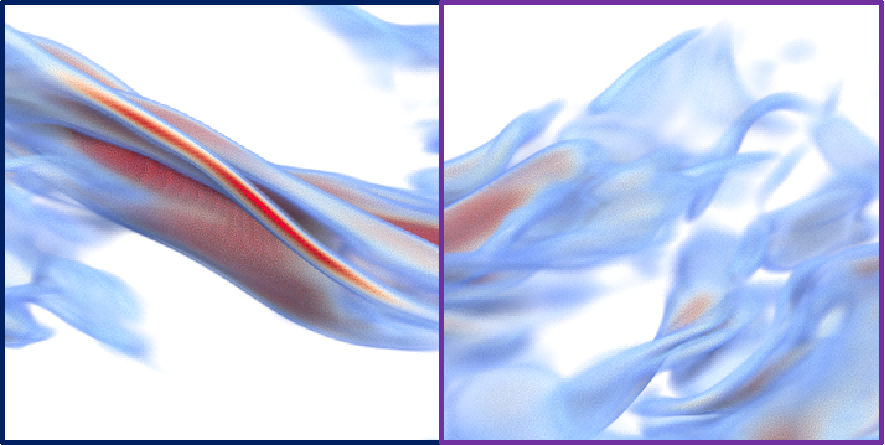}
    \caption{MoE-INR}
  \end{subfigure}
  \begin{subfigure}[t]{\w}\centering
    \includegraphics[width=\linewidth]{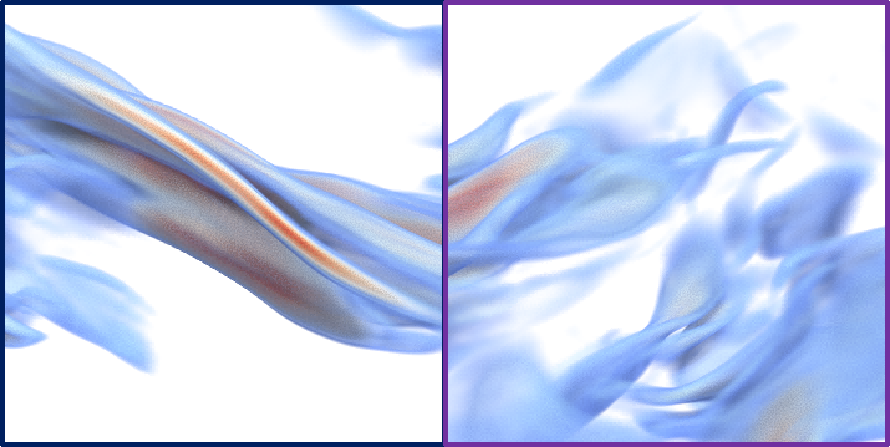}
    \caption{\ours}
  \end{subfigure}
  \begin{subfigure}[t]{\w}\centering
    \includegraphics[width=\linewidth]{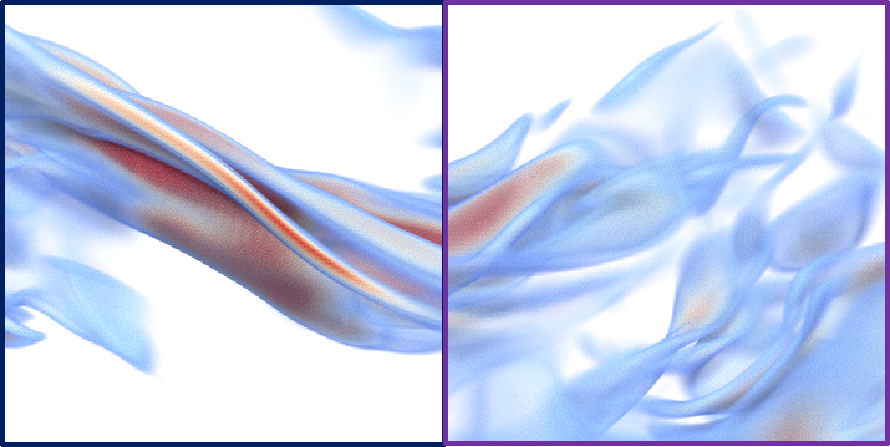}
    \caption{GT}
  \end{subfigure}

  \caption{Comparison of volume rendering results with MoE-based compression INR methods on the combustion (CHI) data set, and the CR is 2,585.}
  \label{fig:chi_moe_images}
\end{figure*}

\begin{figure*}[ht]
  \centering
  \newcommand{\w}{0.19\textwidth}
  \newcommand{\wsmall}{0.095\textwidth}  
  \begin{subfigure}[t]{\w}\centering
    \includegraphics[width=\linewidth]{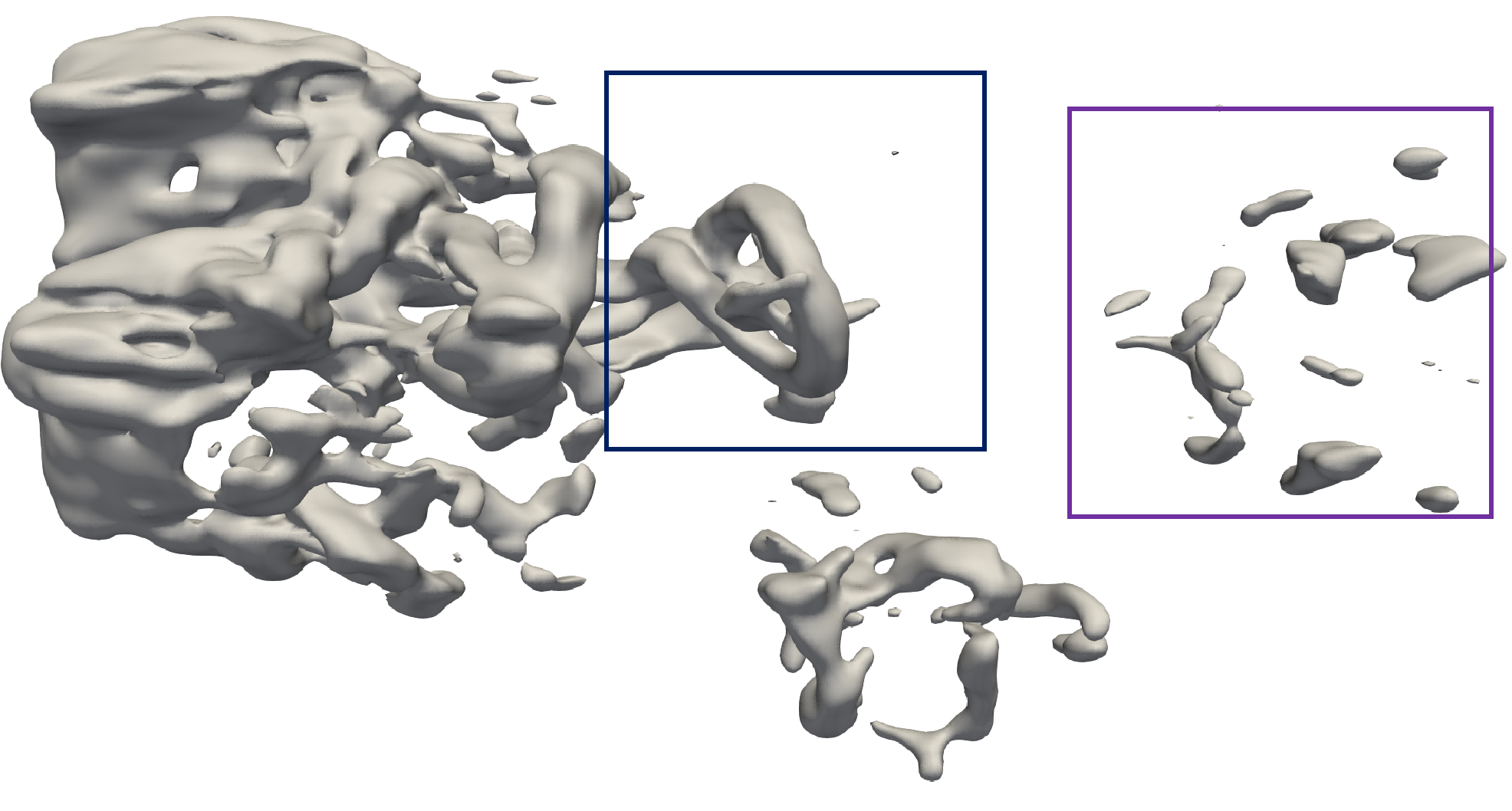}

  \end{subfigure}
  \begin{subfigure}[t]{\w}\centering
    \includegraphics[width=\linewidth]{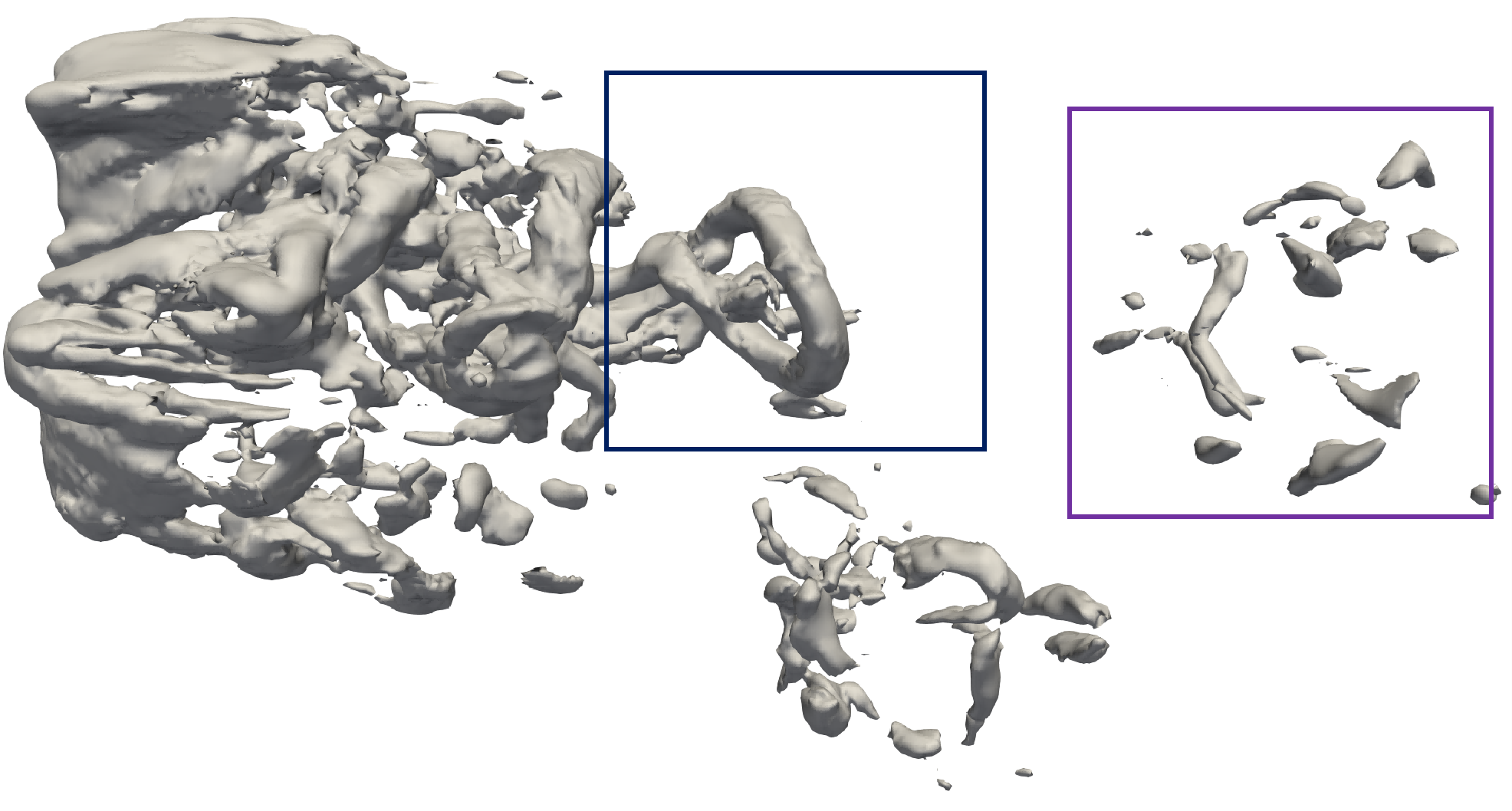}

  \end{subfigure}
  \begin{subfigure}[t]{\w}\centering
    \includegraphics[width=\linewidth]{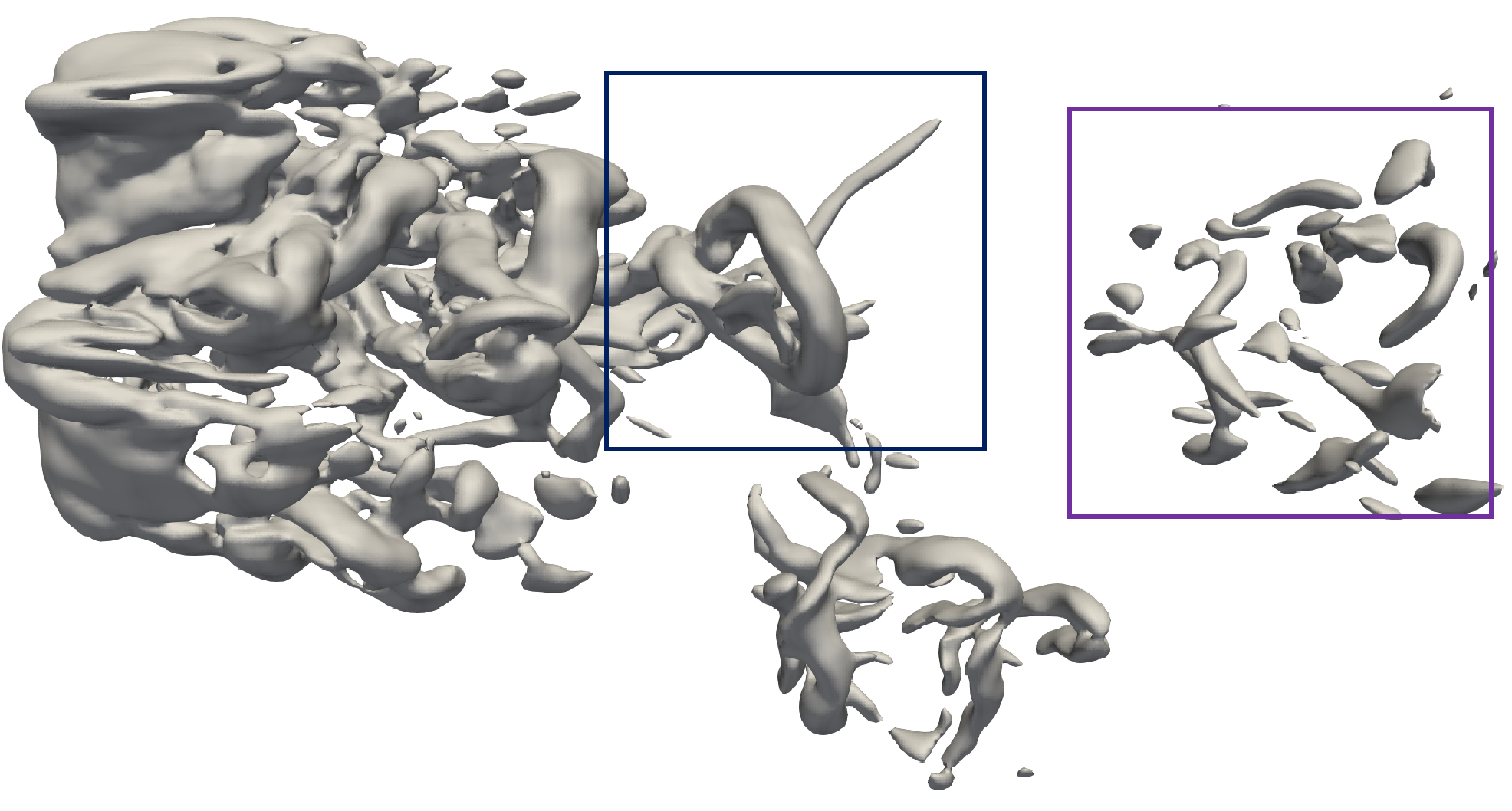}

  \end{subfigure}
  \begin{subfigure}[t]{\w}\centering
    \includegraphics[width=\linewidth]{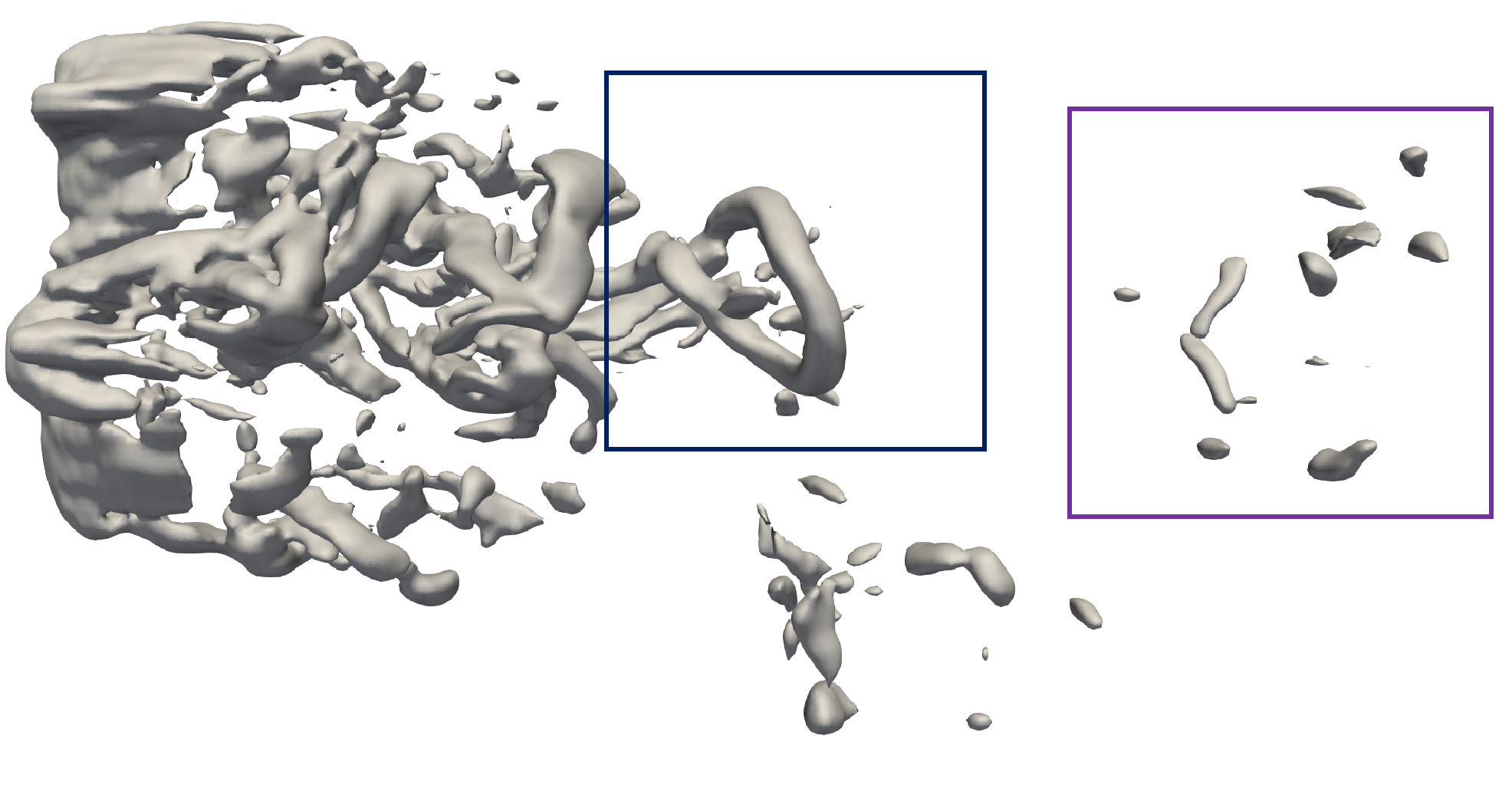}

  \end{subfigure}
  \begin{subfigure}[t]{\w}\centering
    \includegraphics[width=\linewidth]{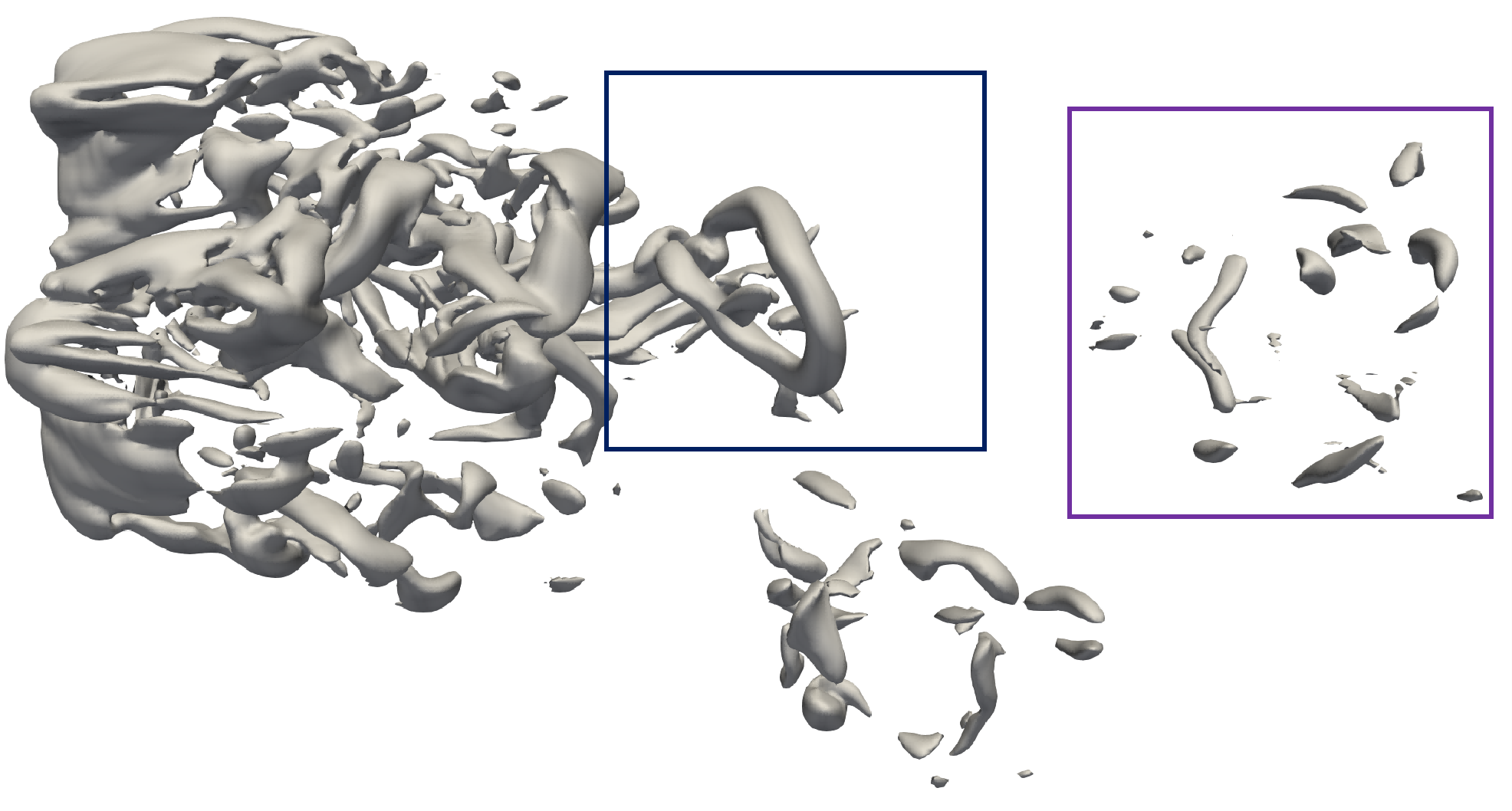}

  \end{subfigure}

\vspace{4pt}

  \begin{subfigure}[t]{\w}\centering
    \includegraphics[width=\linewidth]{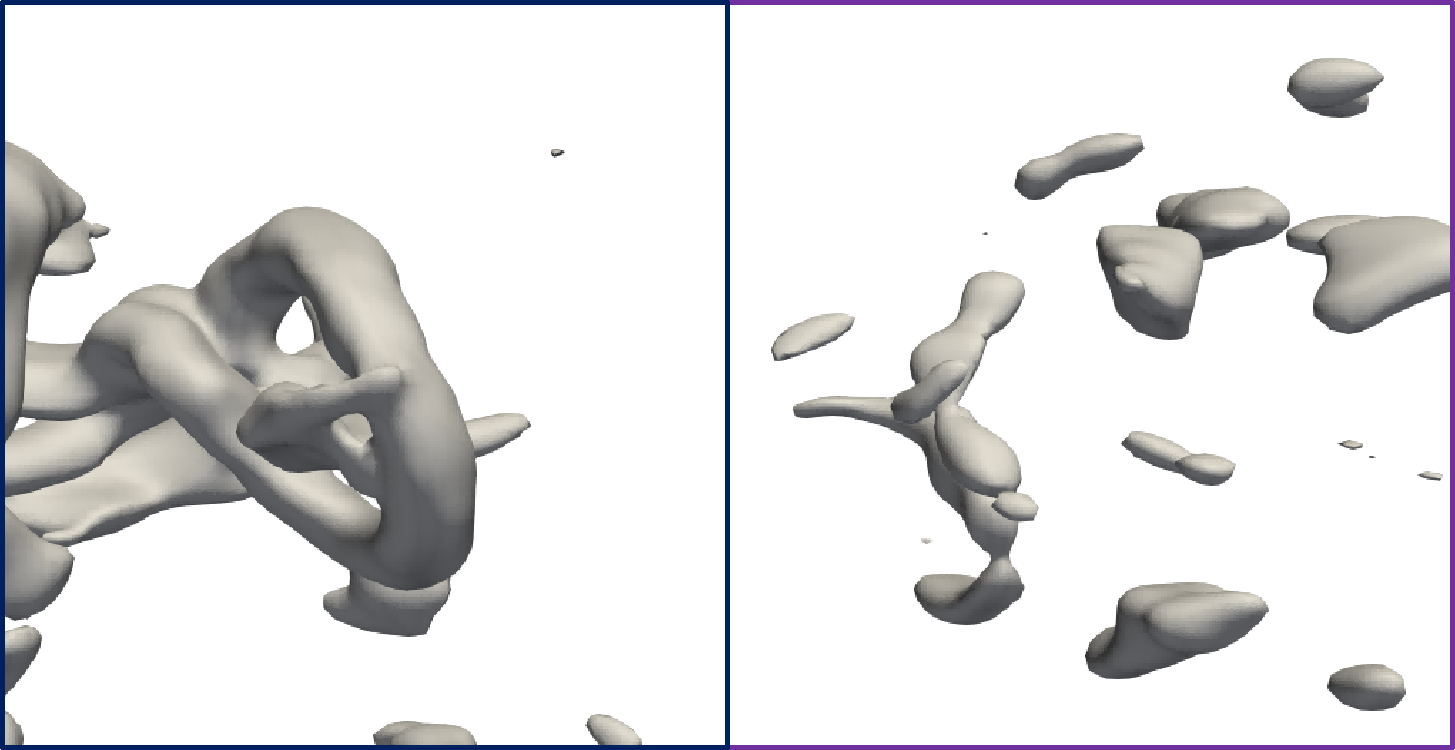}
    \caption{Neural Experts}
  \end{subfigure}
  \begin{subfigure}[t]{\w}\centering
    \includegraphics[width=\linewidth]{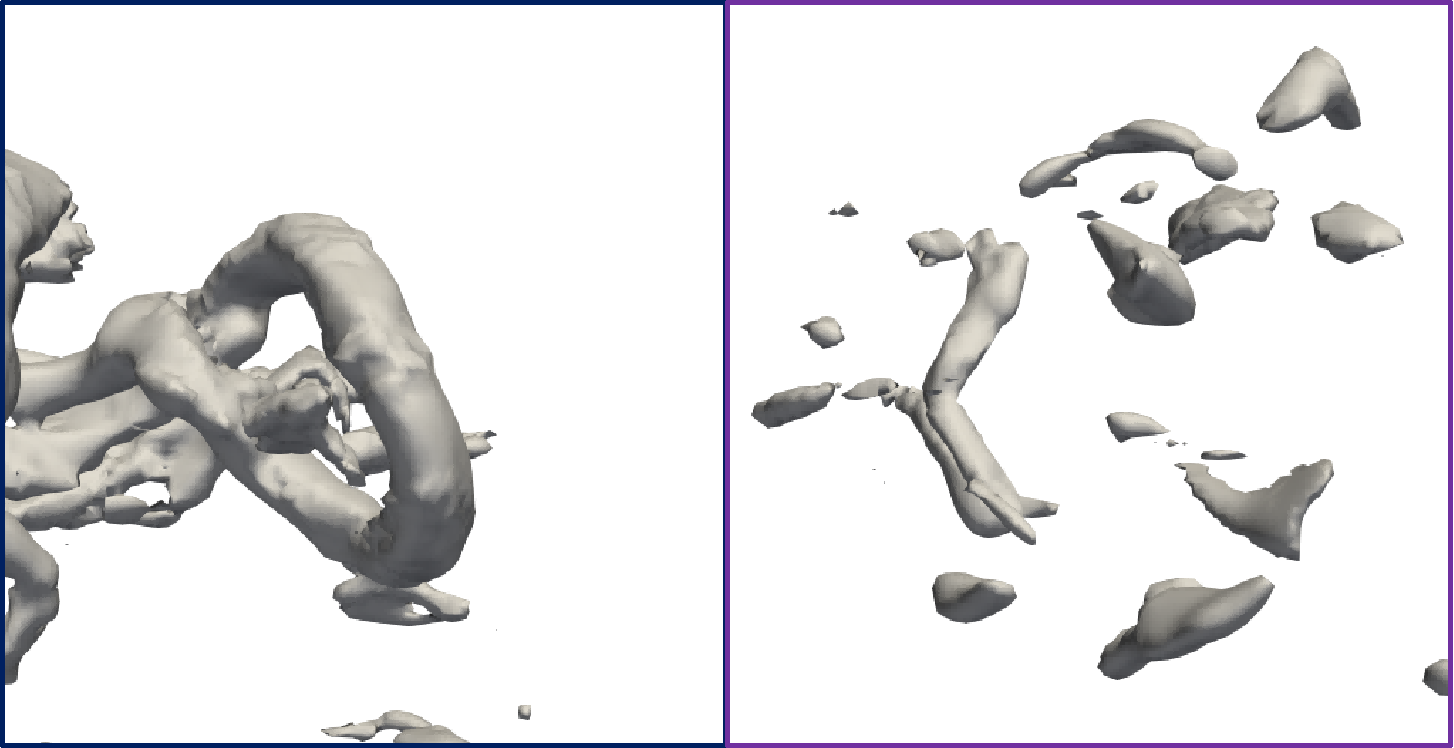}
    \caption{Switch-NeRF}
  \end{subfigure}
  \begin{subfigure}[t]{\w}\centering
    \includegraphics[width=\linewidth]{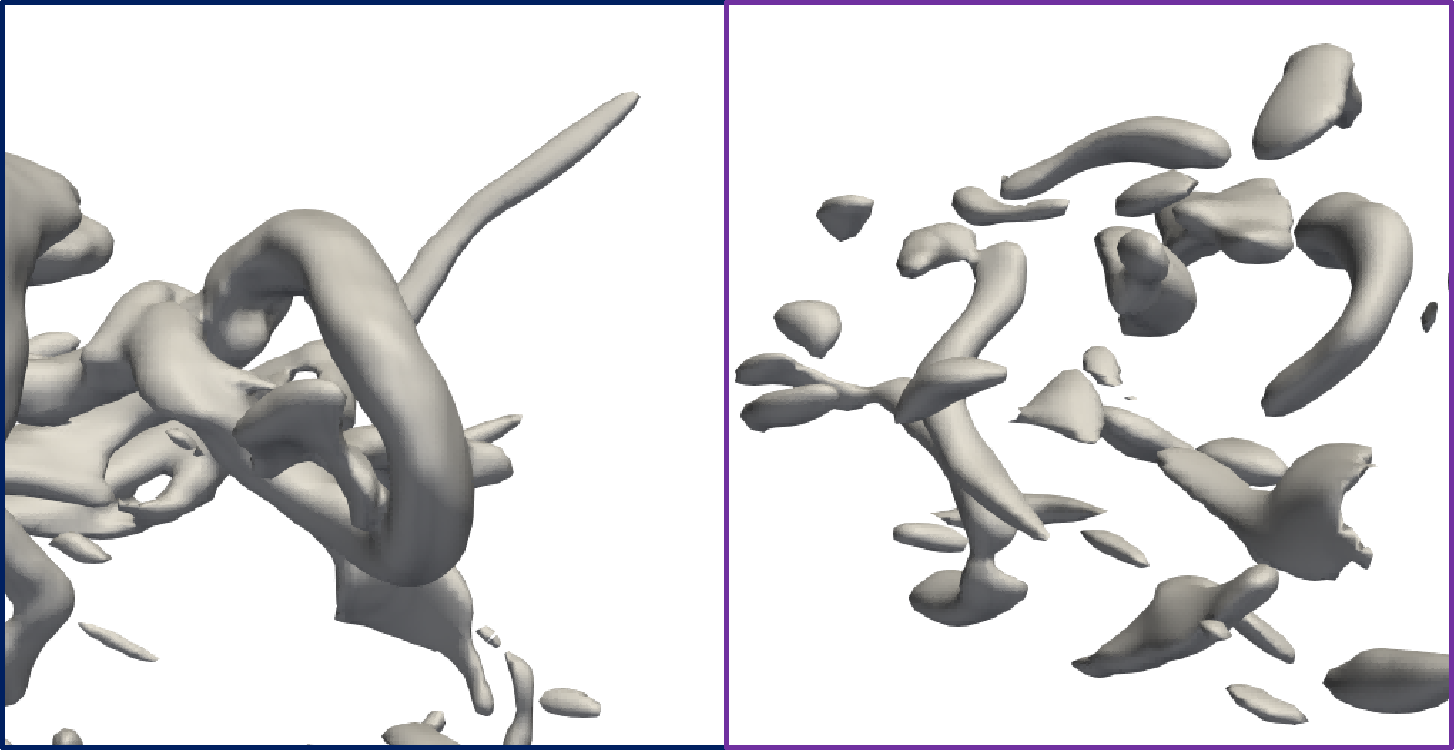}
    \caption{MoE-INR}
  \end{subfigure}
  \begin{subfigure}[t]{\w}\centering
    \includegraphics[width=\linewidth]{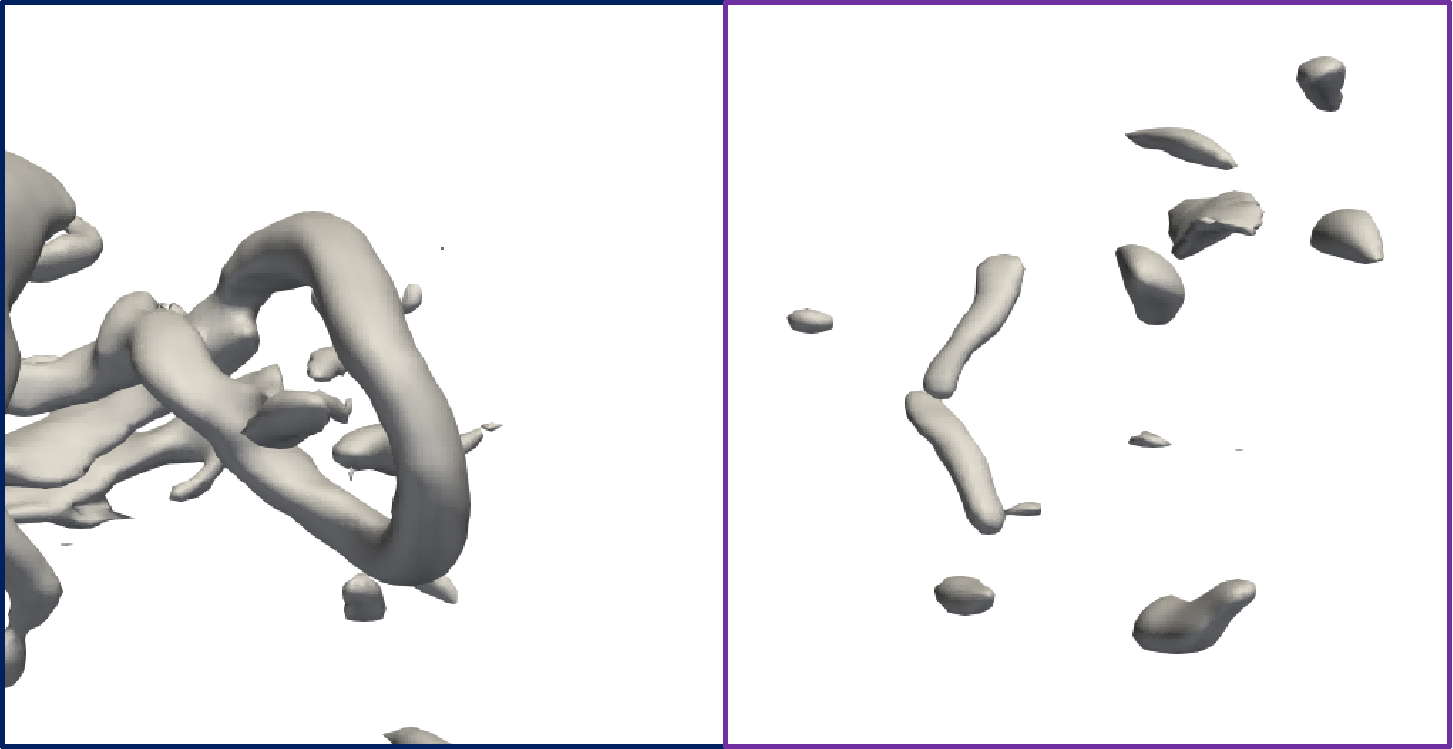}
    \caption{\ours}
  \end{subfigure}
  \begin{subfigure}[t]{\w}\centering
    \includegraphics[width=\linewidth]{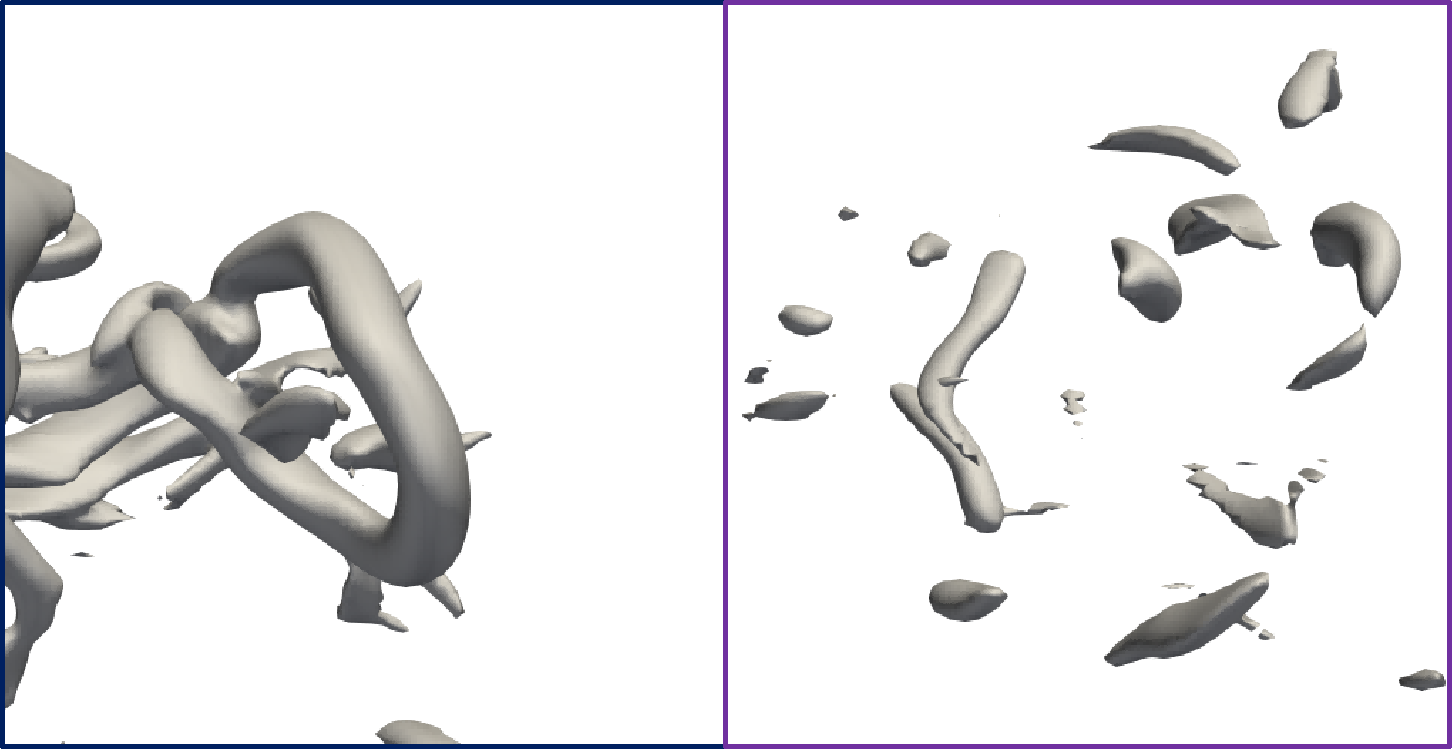}
    \caption{GT}
  \end{subfigure}

  \caption{Comparison of isosurface rendering results with MoE-based compression INR methods on the Tangaroa data set. The CR is 2,314, and the chosen isovalue is -0.7.}
  \label{fig:tangaroa_moe_images}
\end{figure*}

\textbf{Visual comparison.}
On the argon bubble dataset in~\cref{fig:bubble_learning_render}, conventional INRs such as NeurComp and SIREN exhibit slight over-smoothing in regions with thin structures, leading to blurred boundaries and loss of fine geometric details in the zoomed-in areas. 
\hot{In contrast, our framework preserves sharper structural transitions and more continuous bubble surfaces in the highlighted regions. However, this local advantage does not lead to uniformly better perceptual metrics, indicating that local structural fidelity and image-level perceptual similarity emphasize different reconstruction properties.}
As shown in~\cref{fig:vortex_learning_render}, on the vortex dataset, the differences become more pronounced in regions with complex swirling structures. Baseline methods like SIREN tend to produce either fragmented patterns or locally distorted shapes, particularly in areas with strong intensity gradients. Benefiting from expert specialization and temporal-aware routing, our framework maintains more coherent iso-surface shapes and smoother structural evolution, closely matching the ground truth. 
\hot{This observation is consistent with the best HD achieved by our framework in~\cref{tab:surface_metrics}. However, its LPIPS and DreamSim scores are not uniformly the best, suggesting that geometric consistency and rendered-image perceptual similarity capture different reconstruction properties.}
\hot{In the visual comparison with MoE-based INR models on combustion (CHI) in~\cref{fig:chi_moe_images}, Neural Experts exhibits discontinuities near the volume boundaries, while Switch-NeRF additionally smooths some fine-scale structures. Similar blurring and mild noise are visible for Switch-NeRF on Tangaroa in~\cref{fig:tangaroa_moe_images}.}
It is likely related to its MoE routing behavior: when expert assignments vary frequently across neighboring samples, the final prediction effectively becomes a blend of multiple experts. 

\begin{figure*}[ht]
  \centering
  \newcommand{\w}{0.19\textwidth}
  \newcommand{\wsmall}{0.095\textwidth}  
  \begin{subfigure}[t]{\w}\centering
    \includegraphics[width=\linewidth]{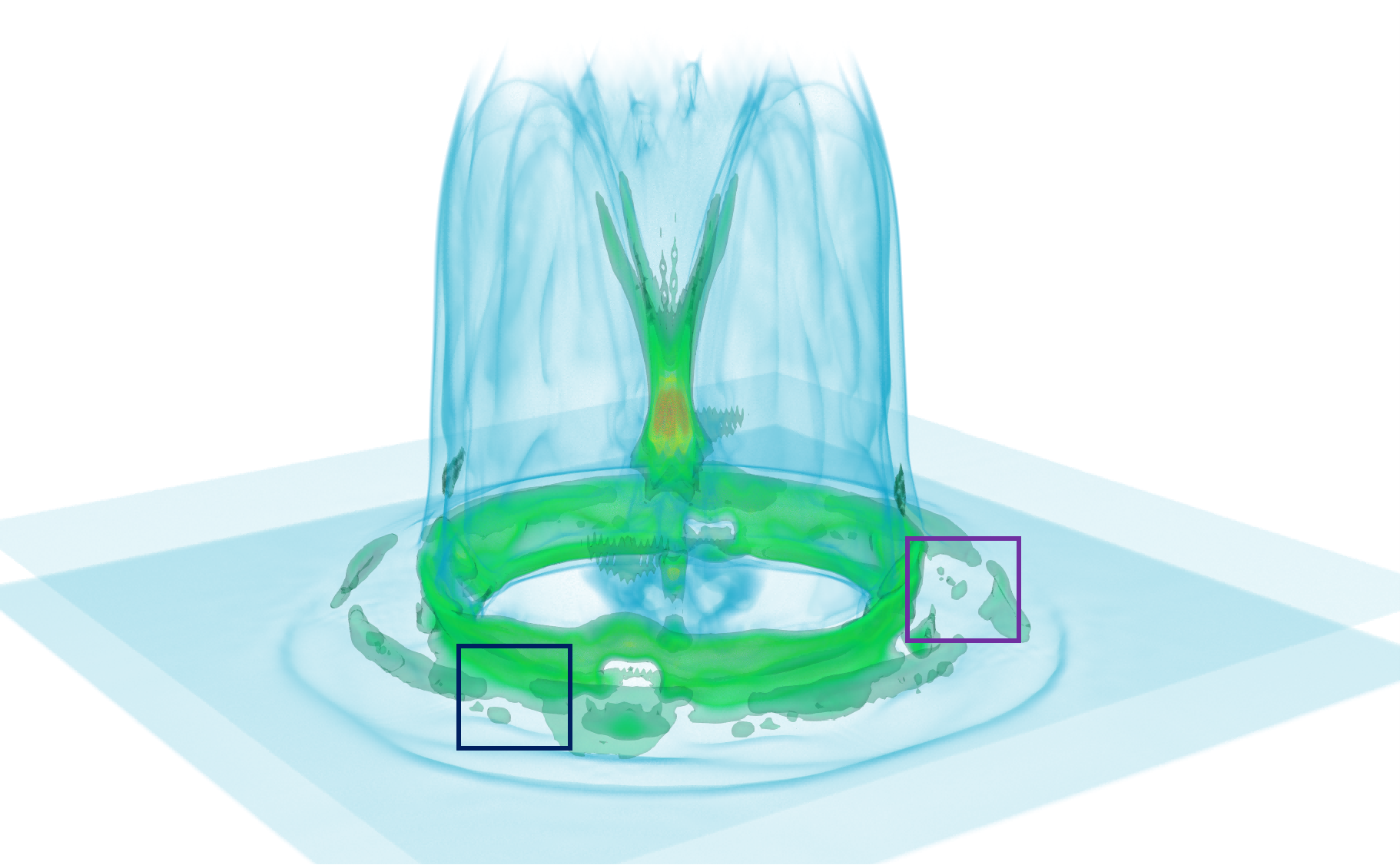}

  \end{subfigure}
  \begin{subfigure}[t]{\w}\centering
    \includegraphics[width=\linewidth]{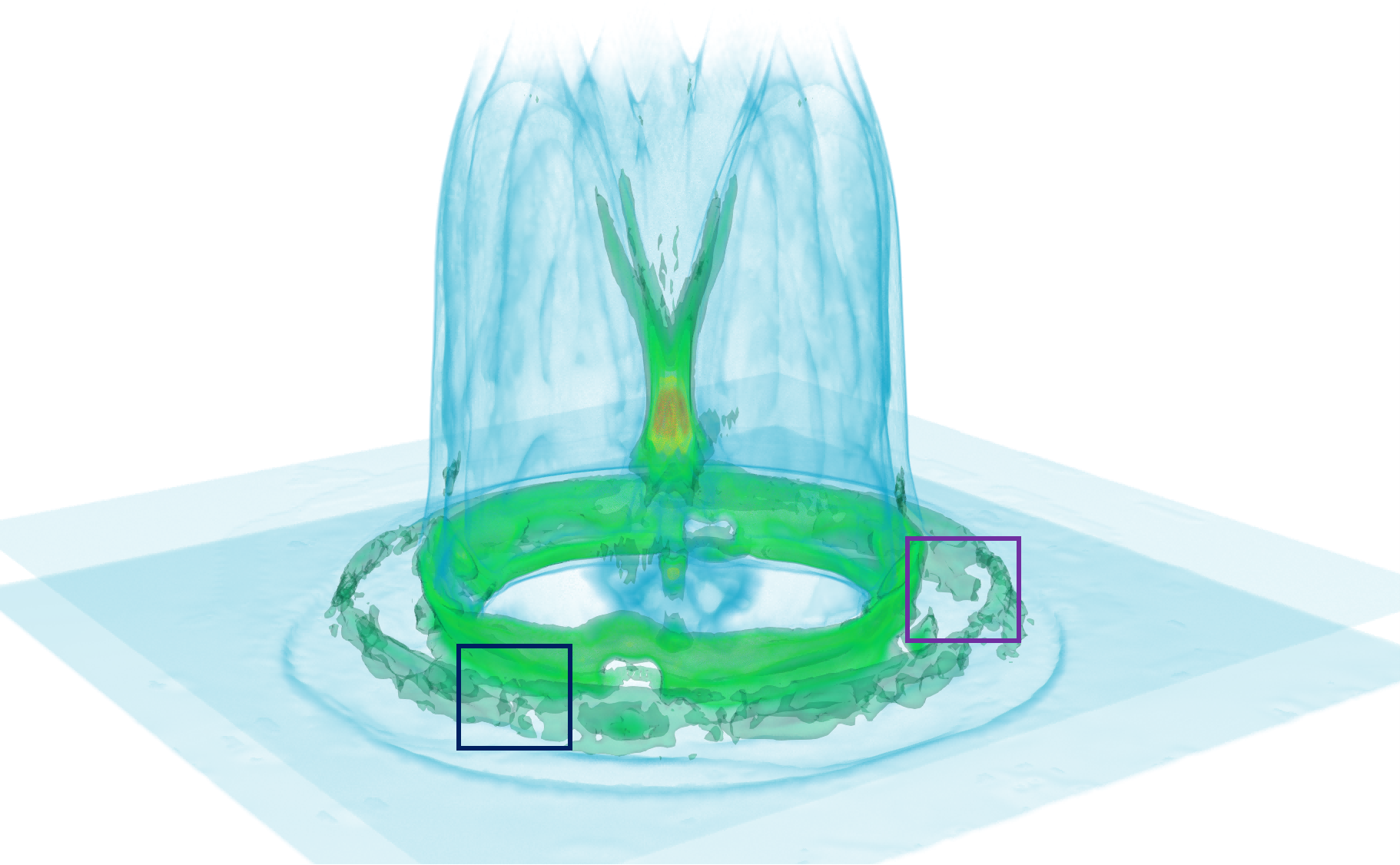}

  \end{subfigure}
  \begin{subfigure}[t]{\w}\centering
    \includegraphics[width=\linewidth]{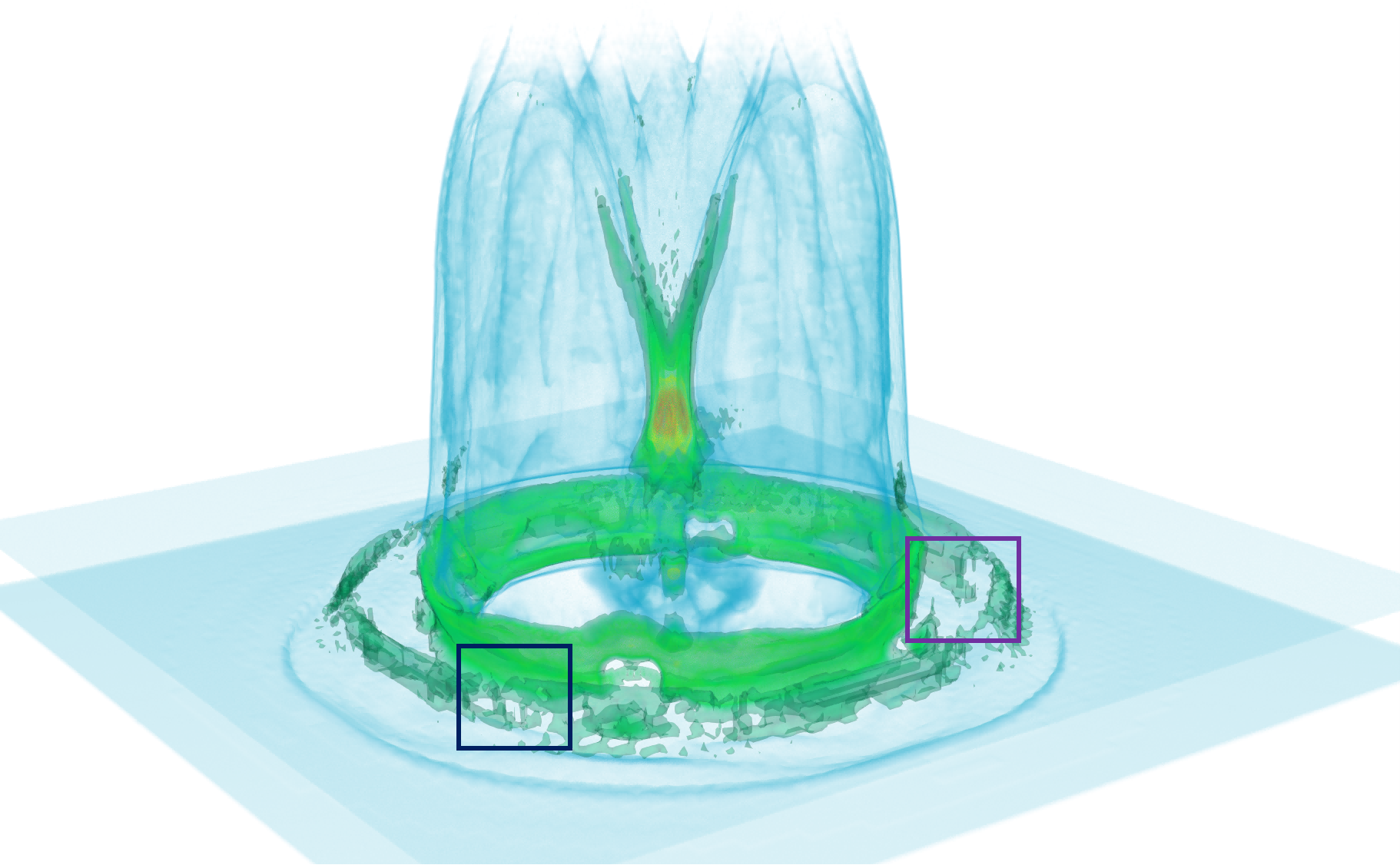}

  \end{subfigure}
  \begin{subfigure}[t]{\w}\centering
    \includegraphics[width=\linewidth]{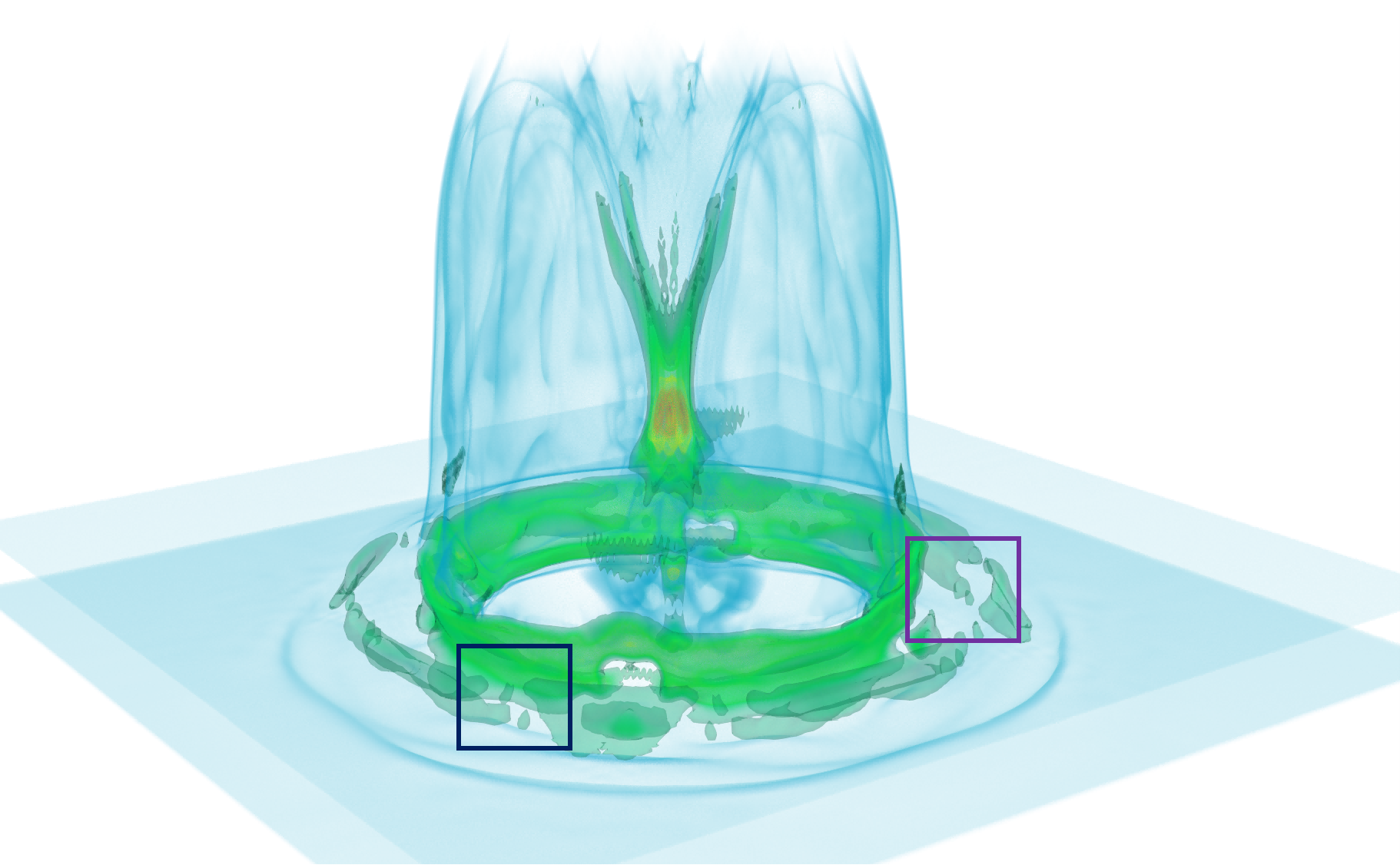}

  \end{subfigure}
  \begin{subfigure}[t]{\w}\centering
    \includegraphics[width=\linewidth]{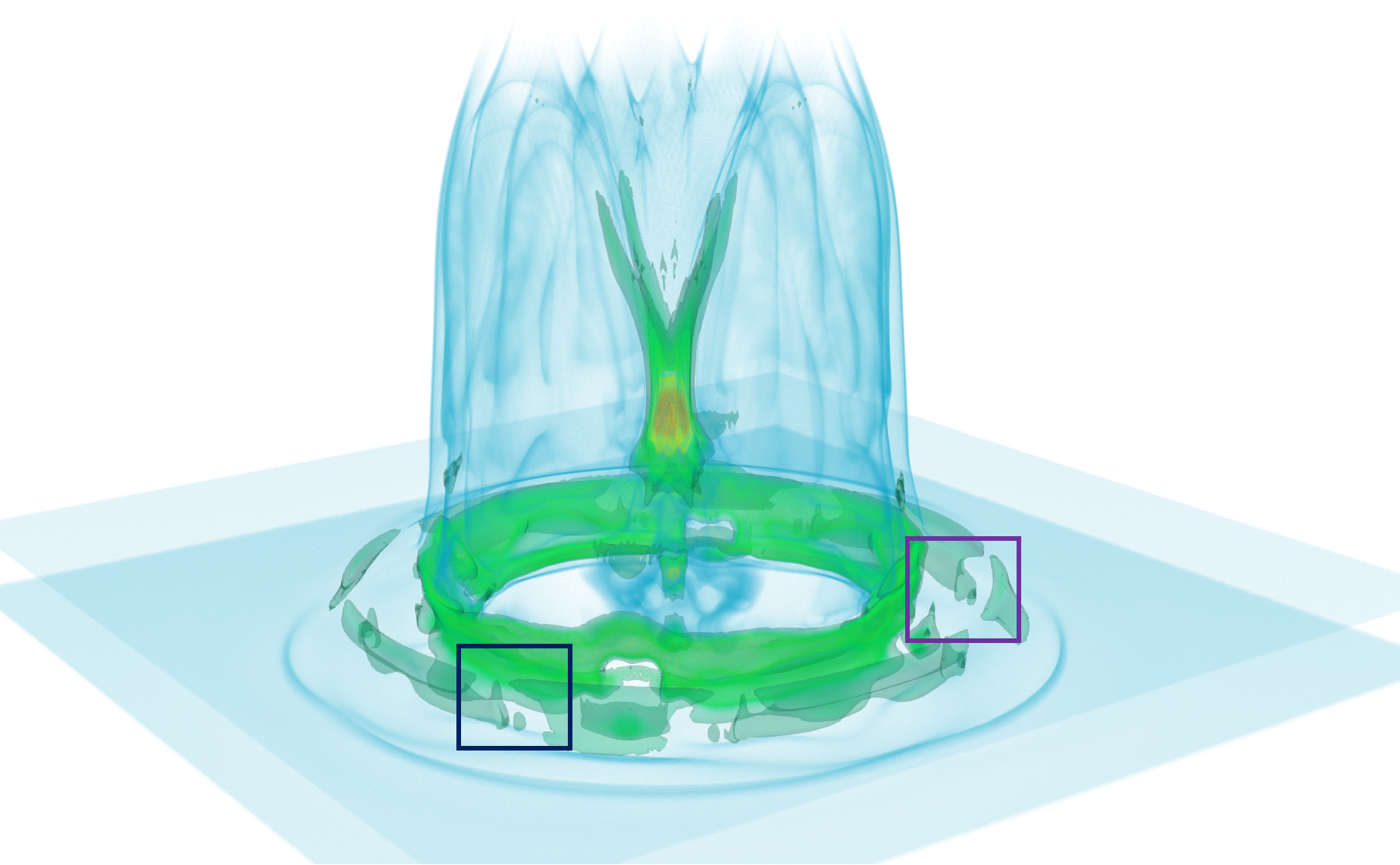}

  \end{subfigure}

\vspace{4pt}

  \begin{subfigure}[t]{\w}\centering
    \includegraphics[width=\linewidth]{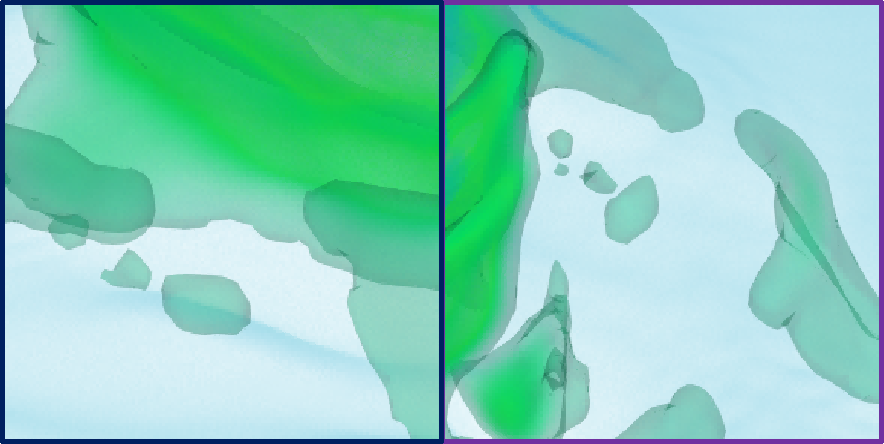}
    \caption{TTHRESH}
  \end{subfigure}
  \begin{subfigure}[t]{\w}\centering
    \includegraphics[width=\linewidth]{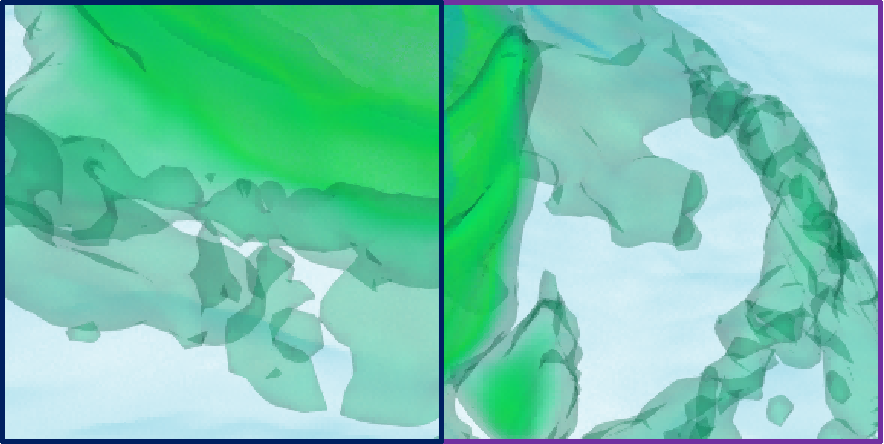}
    \caption{SZ3}
  \end{subfigure}
  \begin{subfigure}[t]{\w}\centering
    \includegraphics[width=\linewidth]{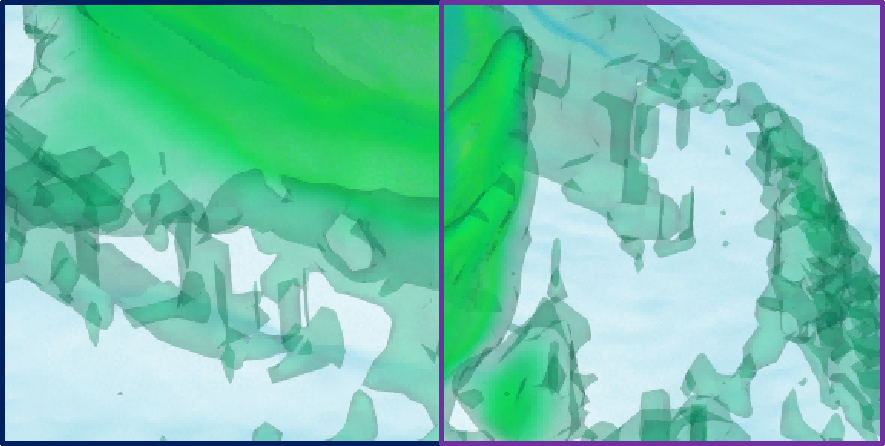}
    \caption{ZFP}
  \end{subfigure}
  \begin{subfigure}[t]{\w}\centering
    \includegraphics[width=\linewidth]{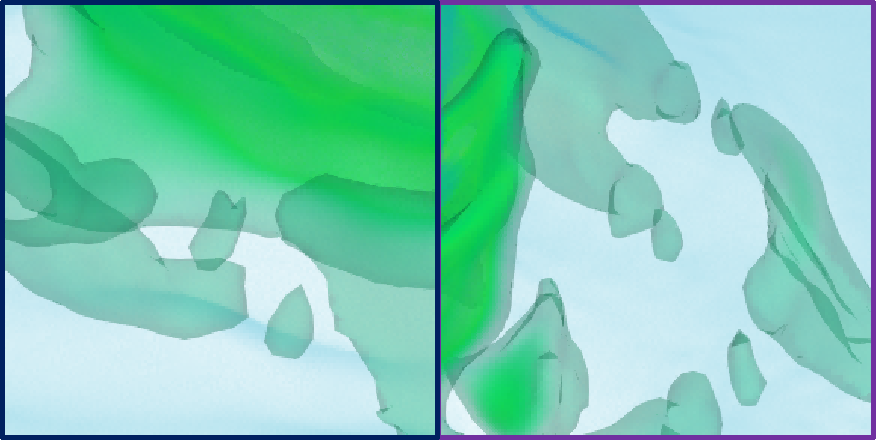}
    \caption{\ours}
  \end{subfigure}
  \begin{subfigure}[t]{\w}\centering
    \includegraphics[width=\linewidth]{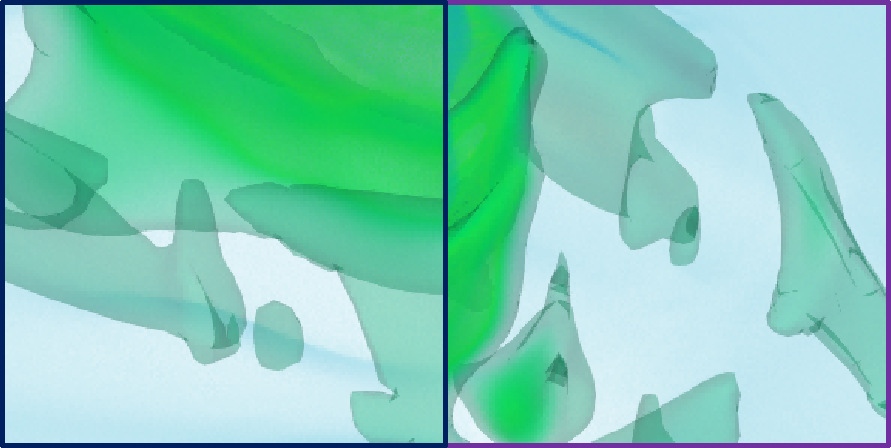}
    \caption{GT}
  \end{subfigure}

  \caption{\hot{Comparison of volume rendering results on the ionization (PD) dataset with traditional lossy compression methods under the same reconstruction quality, and the PSNR is 57.89 dB.}}
  \label{fig:PD_lossy_render}
\end{figure*}

\begin{figure*}[ht]
  \centering
  \newcommand{\w}{0.19\textwidth}
  \newcommand{\wsmall}{0.095\textwidth}  
  \begin{subfigure}[t]{\w}\centering
    \includegraphics[width=\linewidth]{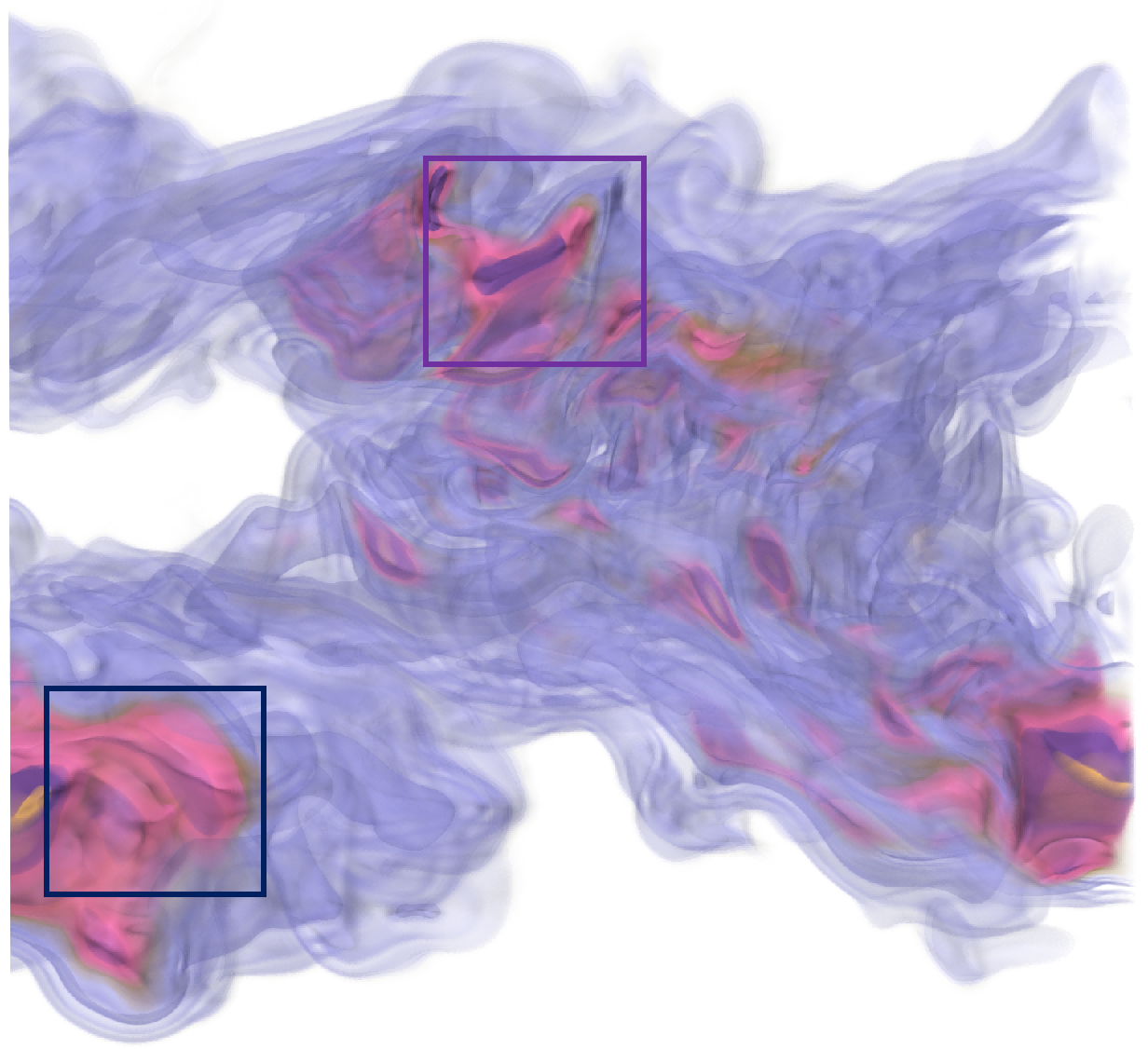}

  \end{subfigure}
  \begin{subfigure}[t]{\w}\centering
    \includegraphics[width=\linewidth]{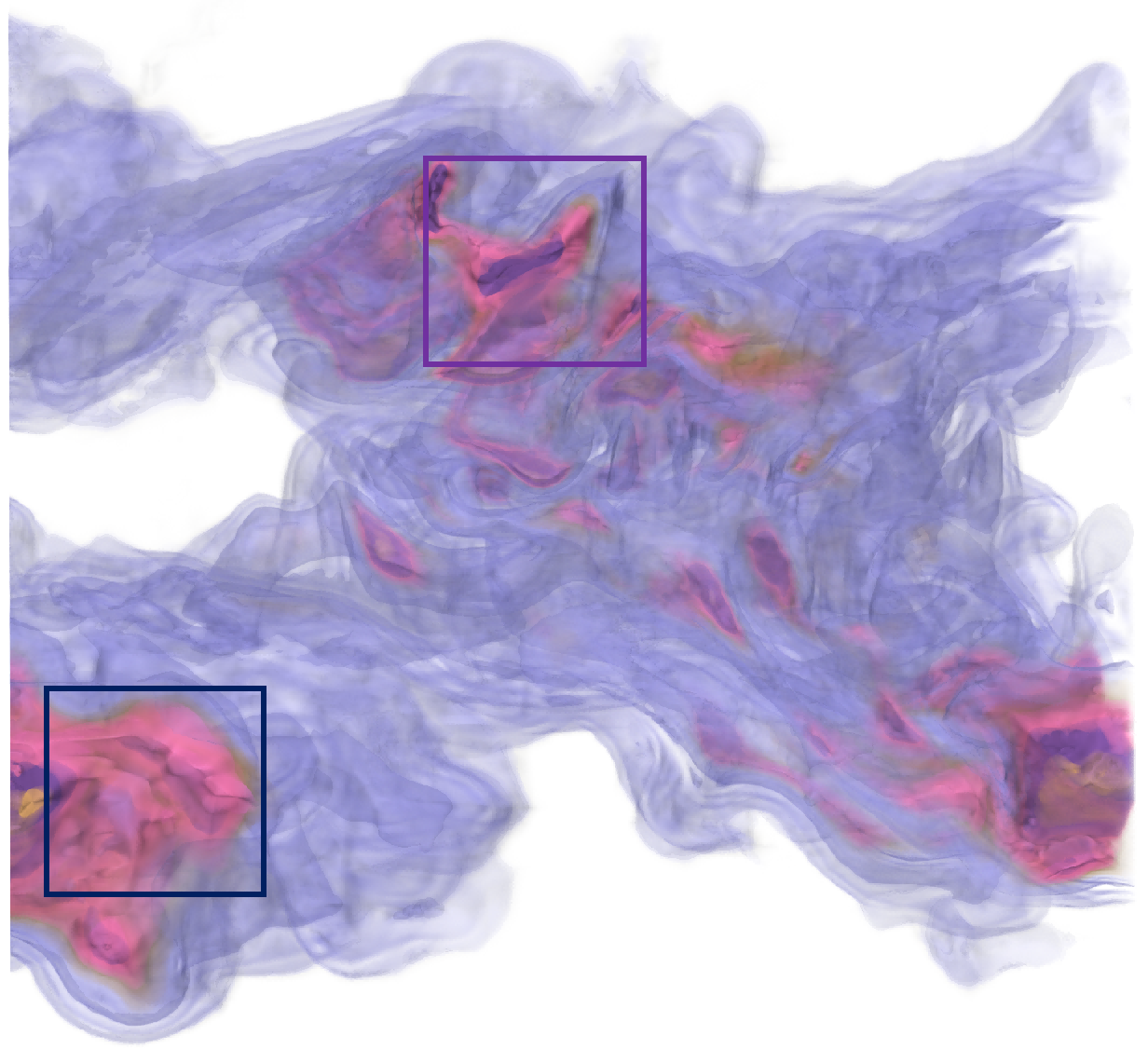}

  \end{subfigure}
  \begin{subfigure}[t]{\w}\centering
    \includegraphics[width=\linewidth]{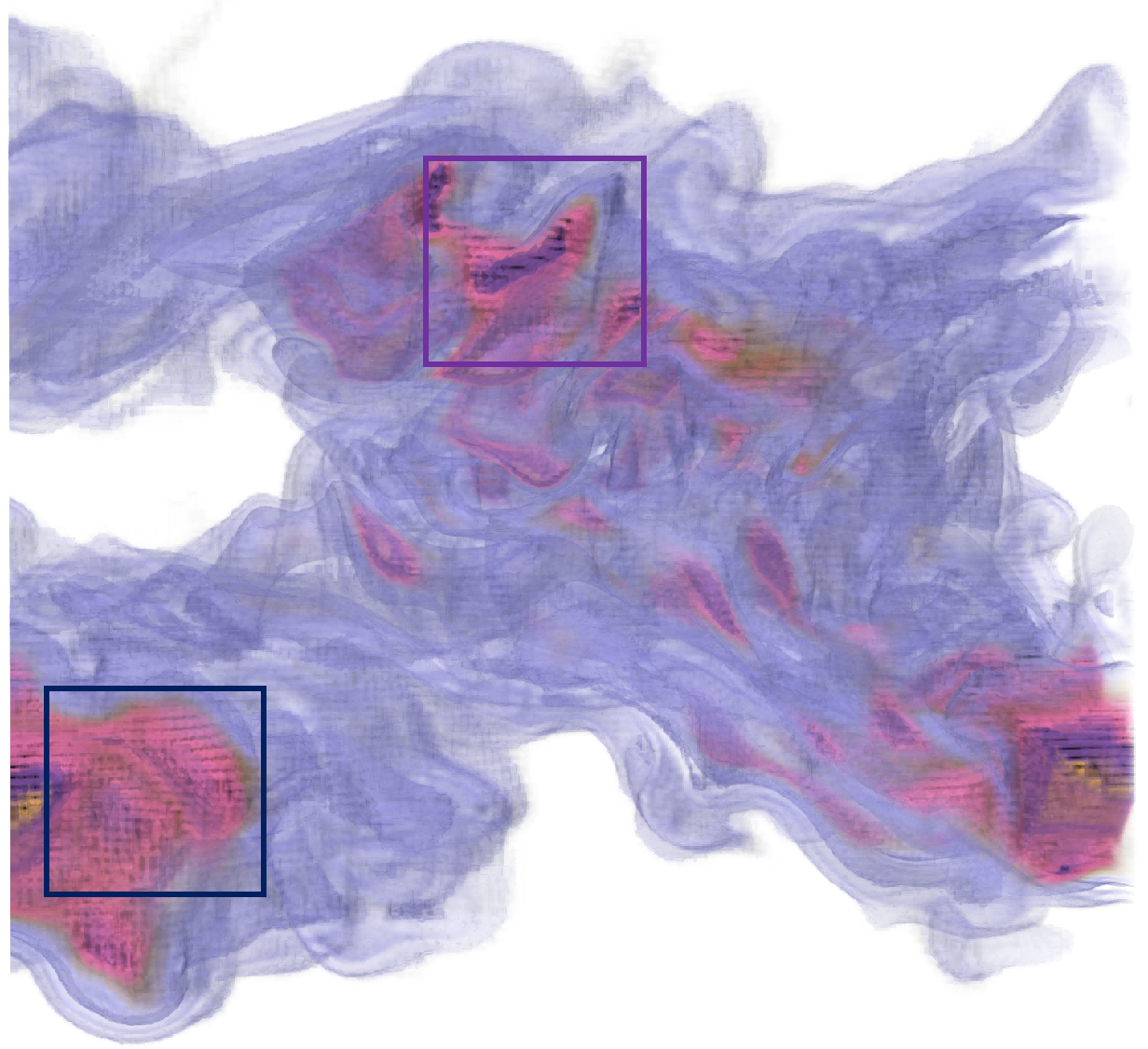}

  \end{subfigure}
  \begin{subfigure}[t]{\w}\centering
    \includegraphics[width=\linewidth]{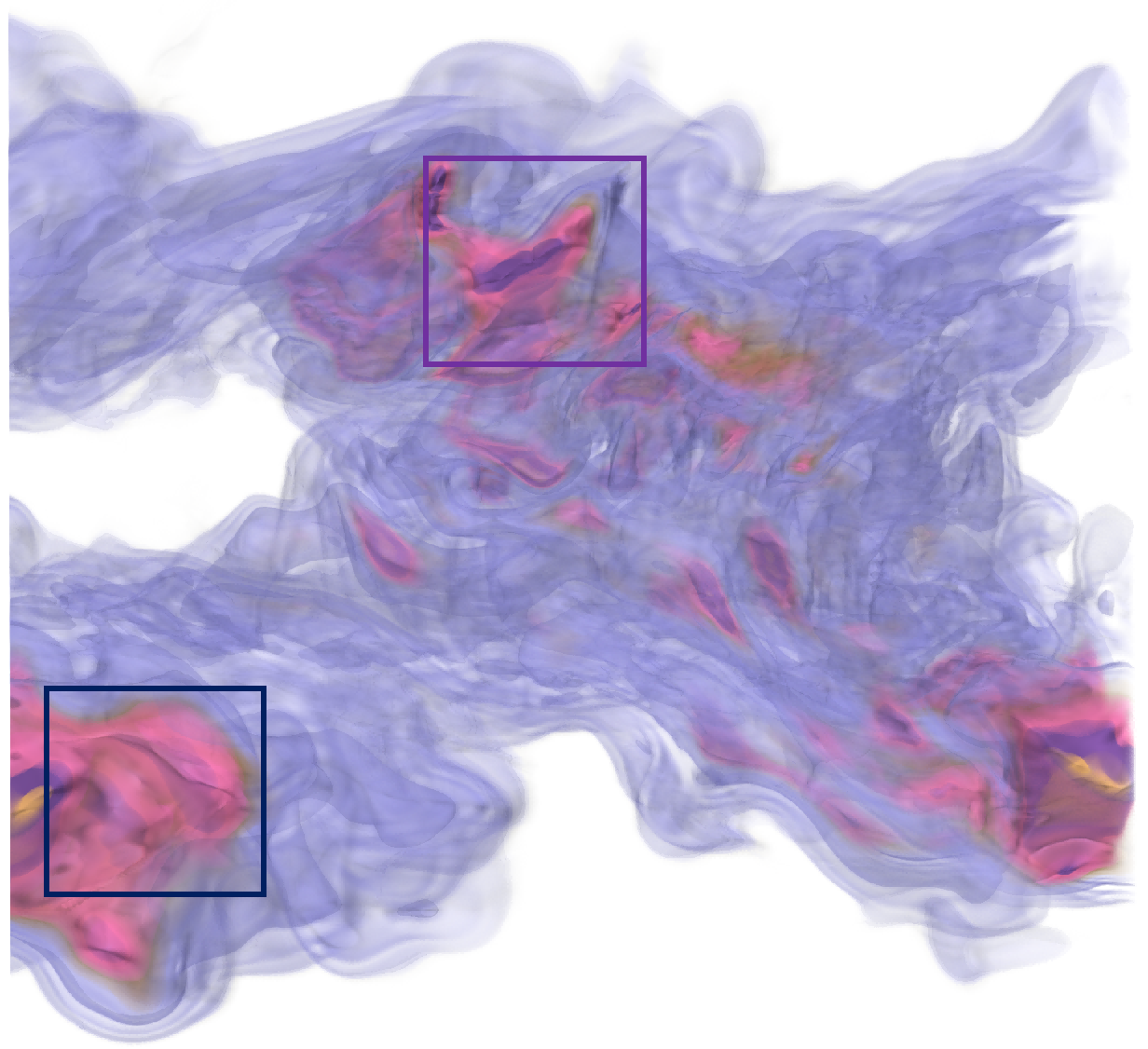}

  \end{subfigure}
  \begin{subfigure}[t]{\w}\centering
    \includegraphics[width=\linewidth]{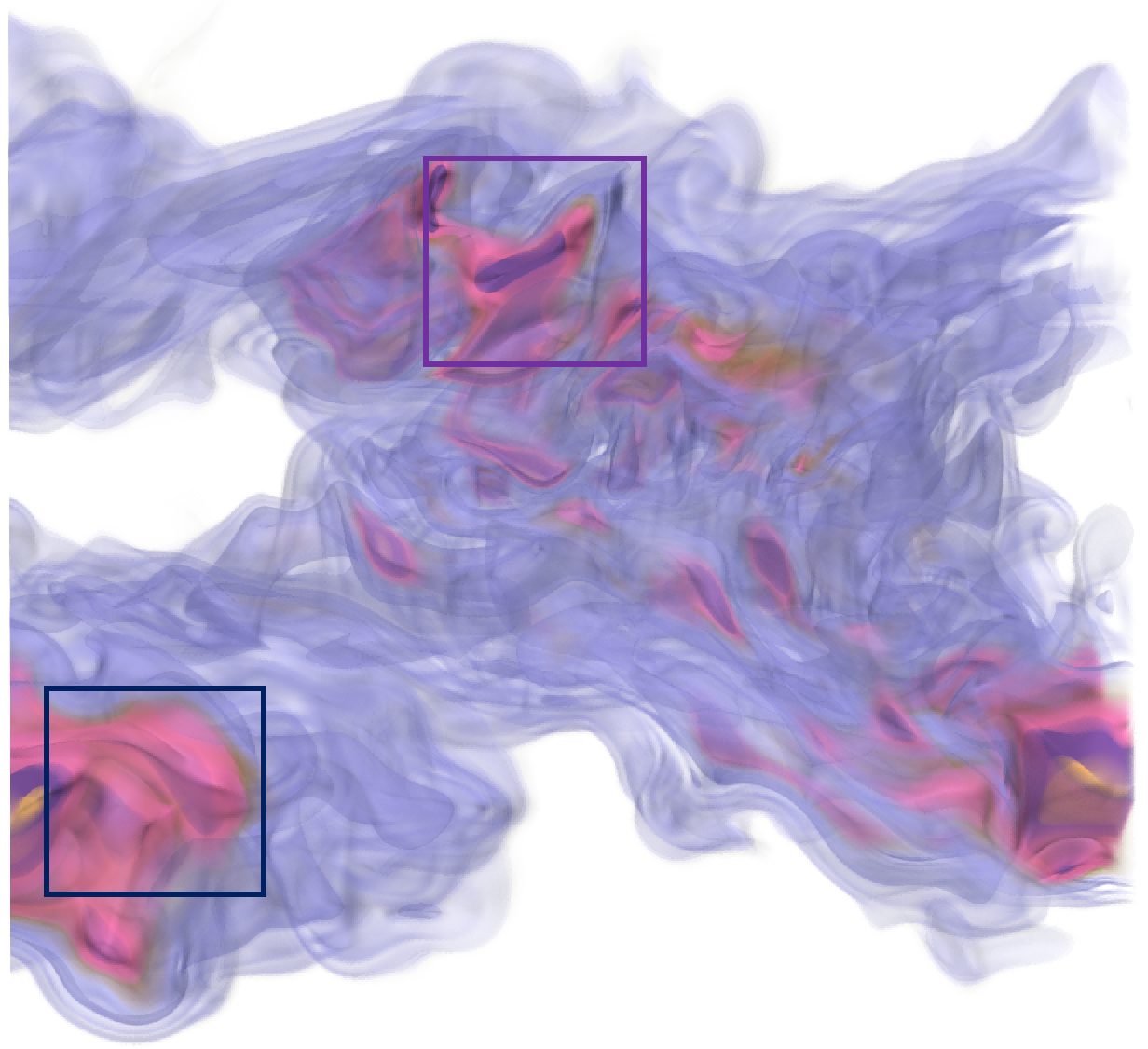}

  \end{subfigure}

\vspace{4pt}

  \begin{subfigure}[t]{\w}\centering
    \includegraphics[width=\linewidth]{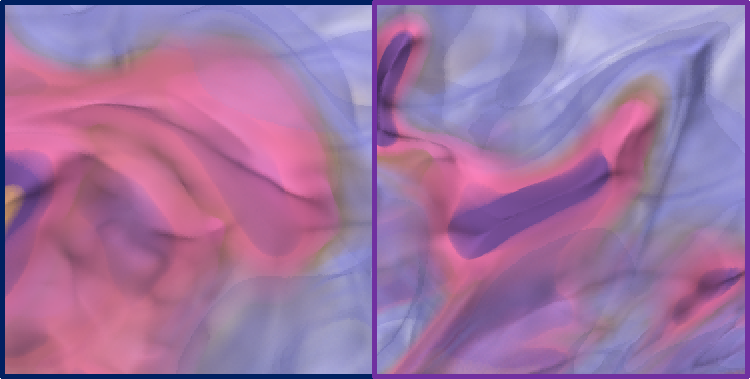}
    \caption{TTHRESH}
  \end{subfigure}
  \begin{subfigure}[t]{\w}\centering
    \includegraphics[width=\linewidth]{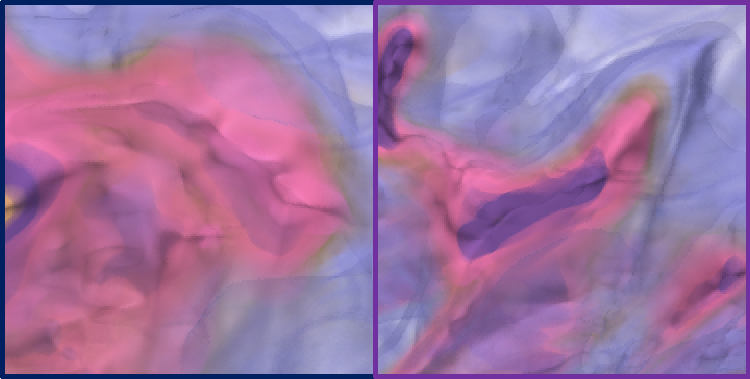}
    \caption{SZ3}
  \end{subfigure}
  \begin{subfigure}[t]{\w}\centering
    \includegraphics[width=\linewidth]{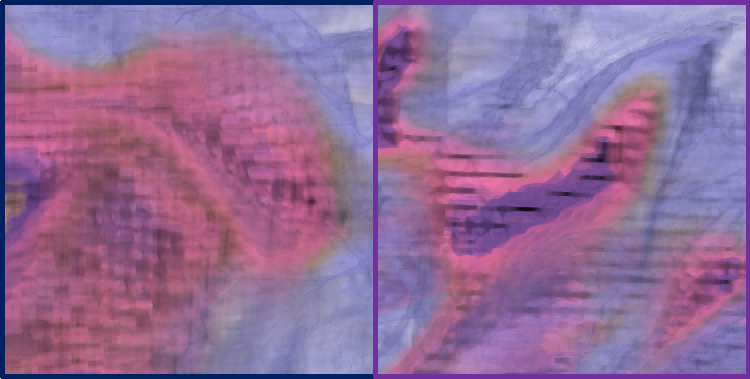}
    \caption{ZFP}
  \end{subfigure}
  \begin{subfigure}[t]{\w}\centering
    \includegraphics[width=\linewidth]{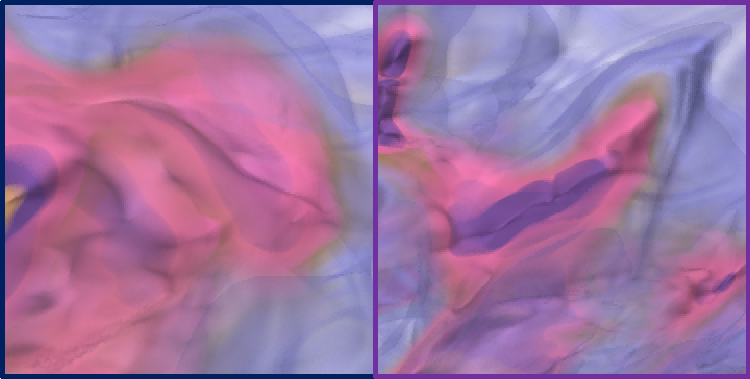}
    \caption{\ours}
  \end{subfigure}
  \begin{subfigure}[t]{\w}\centering
    \includegraphics[width=\linewidth]{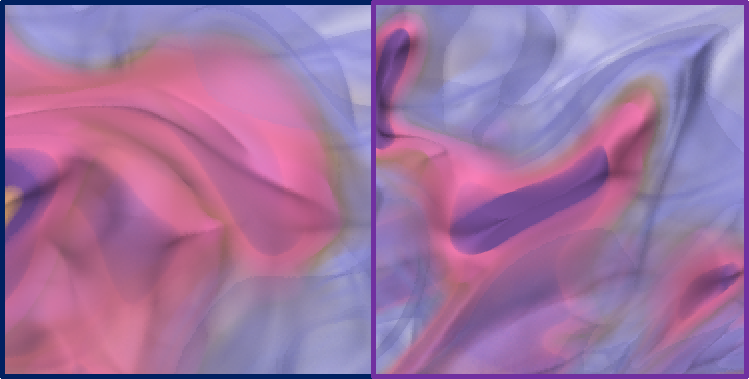}
    \caption{GT}
  \end{subfigure}

  \caption{\hot{Comparison of volume rendering results on the combustion (YOH) dataset with traditional lossy compression methods under the same reconstruction quality, and the PSNR is 42.25 dB.}}
  \label{fig:yoh_lossy_render}
\end{figure*}

\subsection{Comparison Against Lossy Compression Methods}
We further compare our framework with representative traditional lossy compression methods, including ZFP, SZ3, and TTHRESH. To ensure fair comparison, we fix the target PSNR and adjust error bounds accordingly. As shown in~\cref{tab:lossy_performance}, our framework consistently achieves higher \textbf{compression ratios ranging from 9.6$\times$ to 102.2$\times$} than traditional compressors at comparable reconstruction quality, demonstrating stronger representation compactness. 

\hot{For geometric consistency (HD), our framework achieves the best performance on 2/3 datasets, while traditional compressors perform better on ionization (H2). Although SZ3 achieves the best HD on H2, it requires 102.2$\times$ more storage than our approach. TTHRESH provides consistent performance, ranking second across all three datasets. In terms of efficiency, traditional compressors remain faster due to their lightweight encoding procedures. However, they achieve lower compression ratios, highlighting the trade-off between encoding speed and representation compactness.}

Visual comparisons further highlight these differences. 
\hot{On the ionization (PD) dataset (~\cref{fig:PD_lossy_render}), ZFP and SZ3 produce smoother reconstructions but blur some thin structures in the highlighted regions, while TTHRESH introduces mild local distortions. Our framework retains sharper boundaries and more visible internal details in these regions, although this local visual advantage is not uniformly reflected by the global HD metric. On the combustion (YOH) dataset (~\cref{fig:yoh_lossy_render}), which contains stronger high-frequency turbulence, ZFP exhibits visible block artifacts, while SZ3 smooths some small-scale structures. TTHRESH preserves several high-frequency features more effectively but introduces local distortions in other regions. Among the traditional methods, TTHRESH preserves some high-frequency details particularly well and locally outperforms our framework in the highlighted regions, possibly because its transform-domain representation effectively captures sparse high-frequency structures. In contrast, our framework provides competitive overall geometric consistency, as reflected by the HD results in~\cref{tab:lossy_performance}.}

\begin{table}[t]
\centering
\caption{Quantitative comparison of reconstruction quality across different lossy compressors. CR indicates compression ratio, HD denotes surface distance, and total CT (in hours) is measured under the same PSNR with isovalue -0.8, 0.8, and 0.2.}
\vspace{2mm}
\resizebox{\linewidth}{!}{
\begin{tabular}{lcccccc}
\toprule
\rowcolor{gray!10}
\textbf{Data set} & \textbf{PSNR (dB)} & \textbf{Method} & \textbf{CR$\uparrow$} & \textbf{HD$\downarrow$} & \textbf{CT$\downarrow$} \\
\midrule

\multirow{4}{*}{\centering\textbf{combustion (YOH)}} 
& \multirow{4}{*}{\centering 42.25}
& ZFP & 148 & 72.14 & 0.06 \\
& & SZ3 & 84 & 57.85 & \textbf{0.02} \\
& & TTHRESH & 235 & 51.46 & 0.55 \\
& & \textbf{\ours} & \textbf{2,256} & \textbf{26.06} & 0.97 \\
\midrule

\multirow{4}{*}{\centering\textbf{ionization (H2)}} 
& \multirow{4}{*}{\centering 52.38}
& ZFP & 62 & 72.93 & 0.02 \\
& & SZ3 & 55 & \textbf{40.77} & \textbf{0.02} \\
& & TTHRESH & 278 & 54.61 & 0.40 \\
& & \textbf{\ours} & \textbf{5,619} & 72.58 & 0.78 \\
\midrule

\multirow{4}{*}{\centering\textbf{vortex}} 
& \multirow{4}{*}{\centering 44.12}
& ZFP & 43 & 63.96 & \textbf{0.01} \\
& & SZ3 & 118 & 63.92 & 0.02 \\
& & TTHRESH & 146 & 39.74 & 0.03 \\
& & \textbf{\ours} & \textbf{2,192} & \textbf{21.83} & 0.30 \\

\bottomrule
\end{tabular}
}
\label{tab:lossy_performance}
\end{table}

\begin{figure}[!t]
    \centering
    \includegraphics[width=1\linewidth]{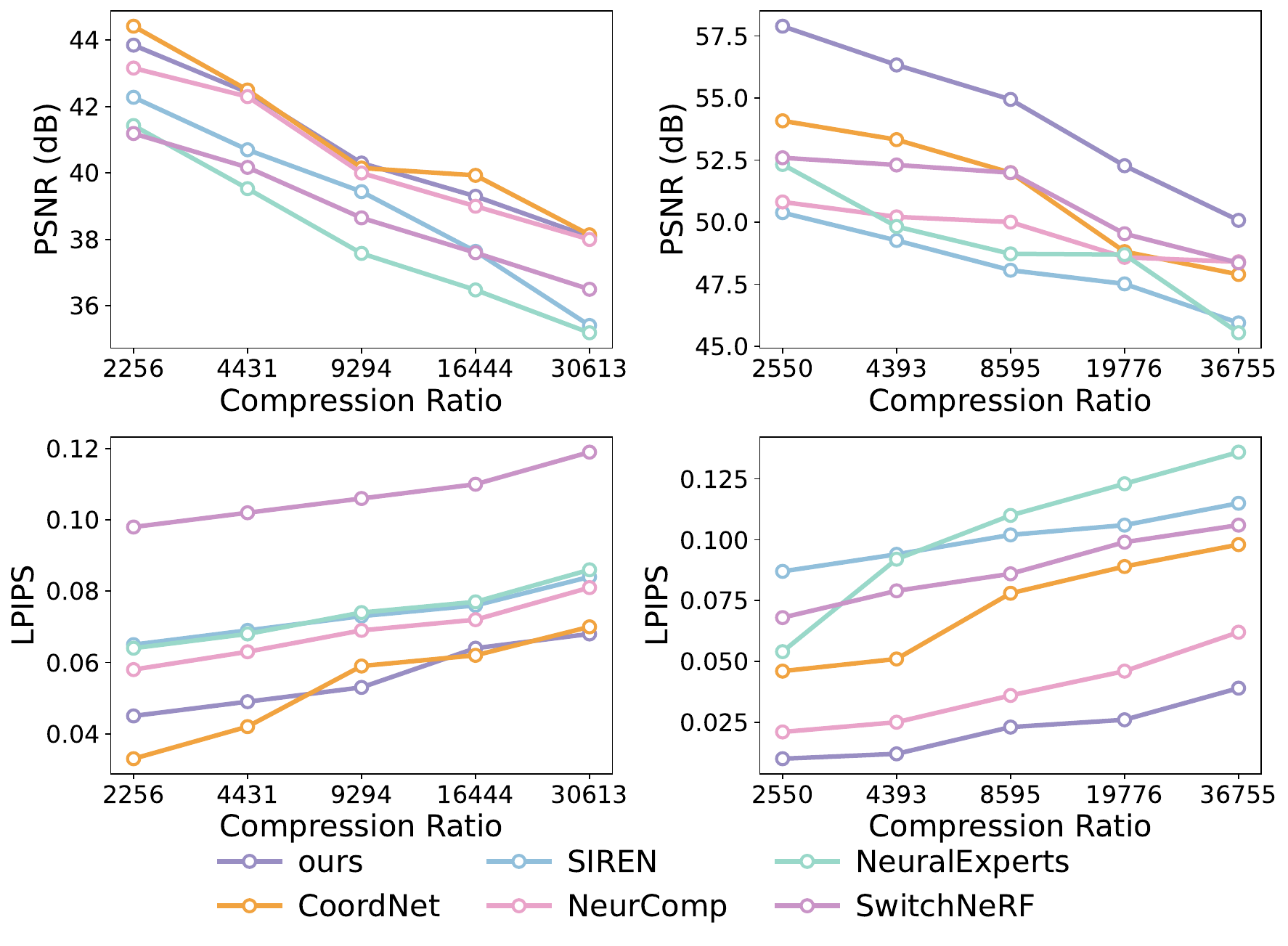}
    \caption{Comparison of PSNR (top) and LPIPS (bottom) under varying CRs among different learning-based INR compressors.
    (Left) combustion (CHI); 
    (Right) ionization (PD).}
    \label{fig:CR_scalability}
\end{figure}

\subsection{Performance Under Varying Compression Ratios}
\cref{fig:CR_scalability} illustrates the reconstruction performance under varying compression ratios for learning-based INR methods. 
On the combustion (CHI) dataset, the performance gap among methods becomes more pronounced at higher compression ratios. While several models achieve comparable PSNR at moderate compression levels, their degradation trends differ as the compression ratio increases. In particular, LPIPS shows more noticeable divergence than PSNR, suggesting that voxel-wise accuracy does not always correlate with perceptual quality under aggressive compression.
As for the ionization (PD) dataset, the degradation patterns appear more consistent across methods. The curves exhibit relatively smooth and monotonic trends, indicating that this dataset may contain more spatially coherent or temporally stable structures.
\section{Ablation Study and Analysis}

In this section, we study and analyze the modules used in our framework through a series of ablation experiments. In particular, we investigate the effectiveness of each component and further analyze how temporal sequences are routed across experts.

\subsection{Ablation Study}


\rev{\textbf{LoRA.} To investigate the effect of LoRA, we compare our framework with and without LoRA in the decoder. As shown in~\cref{tab:lora}, LoRA consistently improves the compression ratio by reducing the number of trainable parameters. As a low-rank reparameterization of the original model, it preserves most of the reconstruction quality with a more compact parameterization. The impact on PSNR is dataset dependent, with a slight decrease on combustion (MF) and an improvement on vortex, indicating that the effectiveness of LoRA is influenced by the underlying data characteristics. A more detailed analysis of different LoRA ranks and their effect on expert diversity is provided in Appendix~B.}

\begin{table}[t]
\centering
\caption{Comparison of PSNR (dB) and compression ratio (CR) with and without LoRA on two data sets.}
\small
\begin{tabular}{l l c c}
\toprule
\textbf{Data set} & \textbf{Method} & \textbf{PSNR (dB) $\uparrow$} & \textbf{CR $\uparrow$} \\
\midrule
\multirow{2}{*}{combustion (MF)} 
    & w/o LoRA & 42.35 & 2,013 \\
    & w/ LoRA & 40.87 & 2,256 \\
\midrule
\multirow{2}{*}{vortex} 
    & w/o LoRA & 46.17 & 876 \\
    & w/ LoRA & 48.18 & 1,120 \\
\bottomrule
\label{tab:lora}
\end{tabular}
\vspace{-0.5cm}\end{table}
\begin{table}[t]
\centering
\caption{Comparison of PSNR (dB) and LPIPS with and without time embedding on two data set.}
\small
\begin{tabular}{l l c c}
\toprule
\textbf{Data set} & \textbf{Method} & \textbf{PSNR (dB) $\uparrow$} & \textbf{LPIPS $\downarrow$} \\
\midrule
\multirow{2}{*}{ionization (H+)} 
    & w/o time embedding & 48.63 & 0.067 \\
    & w/ time embedding & 51.54 & 0.056 \\
\midrule
\multirow{2}{*}{combustion (VORT)} 
    & w/o time embedding & 36.75 & 0.139 \\
    & w/ time embedding & 39.26 & 0.081 \\
\bottomrule
\end{tabular}
\label{tab:time_embedding}
\vspace{-0.2cm}\end{table}

\textbf{Time embedding.}
To examine the impact of time embedding in router network on the decompression performance of routing experts, we evaluate the reconstruction quality using PSNR and assess the perceptual similarity through LPIPS for visual results.
As shown in~\cref{tab:time_embedding}, incorporating time embedding improves both reconstruction accuracy and perceptual quality across datasets. For the ionization (H+) dataset, PSNR increases from 48.63 to 51.54, and LPIPS drops from 0.067 to 0.056. A similar trend is observed in the combustion (VORT) dataset, where PSNR improves by 2.5 dB and LPIPS is reduced. 
These results suggest that time embedding provides a performance gain, particularly in improving temporal consistency during decompression. Predicting full temporal sequences introduces higher modeling complexity. By incorporating trainable time embeddings, the network is better guided to capture temporal patterns across time steps, leading to improved reconstruction stability and temporal consistency.

\begin{table}[h]
\centering
\caption{Quantitative comparison of different INR methods adapted to a \textbf{sequential modeling setting (-seq)}. All methods are extended from their original formulations to process temporal sequences. Higher PSNR, lower CT (in hours) and DT (in minutes) indicate better performance.
The compression ratio is set to 4,400 for the ionization (H2) dataset and 2,300 for the vortex data set. 
}
\label{tab:learning_with_sequence}
\footnotesize
\begin{tabular}{lccc|ccc}
\toprule
 & \multicolumn{3}{c|}{\textbf{ionization (H2)}} & \multicolumn{3}{c}{\textbf{vortex}} \\
\cmidrule(lr){2-4} \cmidrule(lr){5-7}

\textbf{Method} 
& \textbf{PSNR$\uparrow$} 
& \textbf{CT$\downarrow$} & \textbf{DT$\downarrow$}
& \textbf{PSNR$\uparrow$} 
& \textbf{CT$\downarrow$} & \textbf{DT$\downarrow$} \\
\midrule

\rowcolor{red!5}
\multicolumn{7}{l}{\textit{Conventional Networks}} \\
\midrule
SIREN-seq    & 47.96 & 0.42 & 0.26 & 34.64 & 0.24 & 0.11 \\
CoordNet-seq & 50.54 & 0.52 & 1.00 & 36.10 &  0.28 & 0.12 \\
NeurComp-seq & 49.39 & 1.05 & 0.34 & 34.95 & 0.55 & 0.14 \\

\midrule
\rowcolor{red!5}
\multicolumn{7}{l}{\textit{MoE Networks}} \\
\midrule
Neural Experts-seq & 47.29 & 0.83 & 0.30 & 33.02 & 0.49 & 0.13 \\
Switch-NeRF-seq    & 50.06 & 0.69 & 0.37 & 32.78 & 0.48 & 0.15 \\
\ours              & 52.41 & 0.67 & 0.78 & 42.41 & 0.32 & 0.19 \\

\bottomrule
\end{tabular}
\end{table}

\subsection{Additional Analysis}
\label{subsec:analysis}

\textbf{Adapting learning-based INRs for Sequence.}
To further analyze the proposed formulation in our framework, we adapt several learning-based INRs to directly predict temporal sequences using spatial coordinates as input. 
This allows us to examine whether the sequence-wise formulation can benefit existing INR architectures beyond our framework. The quantitative comparison is summarized in~\cref{tab:learning_with_sequence}.
On the ionization (H2) dataset, conventional INR architectures such as CoordNet and NeurComp achieve competitive PSNR values under this setting. Among MoE-based models, Switch-NeRF attains comparable performance to Neural Experts. This is consistent with the relatively simple temporal dynamics in H2, where sequence modeling does not introduce significant additional difficulty.

On the vortex dataset, overall performance decreases for all methods due to more complex temporal dynamics. Notably, conventional INR models suffer a significant drop when adapted to sequence prediction. For example, CoordNet achieves a PSNR of 44.09 under the original coordinate-to-scalar formulation at the same compression ratio, but drops to 36.10 when predicting temporal sequences~\cref{tab:learning_with_sequence}. In contrast, MoE-based INRs exhibit a smaller performance gap under this setting. Neural Experts and Switch-NeRF achieve 32.20 and 35.68 PSNR, respectively, in their original formulations, which are closer to their performance in the sequence modeling setting shown in~\cref{tab:learning_with_sequence}. This suggests that the proposed formulation is more compatible with MoE-based architectures, as it aligns better with their ability to model heterogeneous temporal patterns.

\hot{
Overall, while directly modeling temporal sequences increases the difficulty of the task and may disadvantage methods originally designed for scalar coordinate-to-value prediction, it reduces the performance gap between conventional and MoE-based INR methods under the adapted ``-seq'' setting. Therefore, these results should be interpreted as a controlled analysis of backbone compatibility with our formulation, rather than as a definitive comparison with the original methods.
}

\begin{figure}[!t]
    \centering
    \includegraphics[width=1\linewidth]{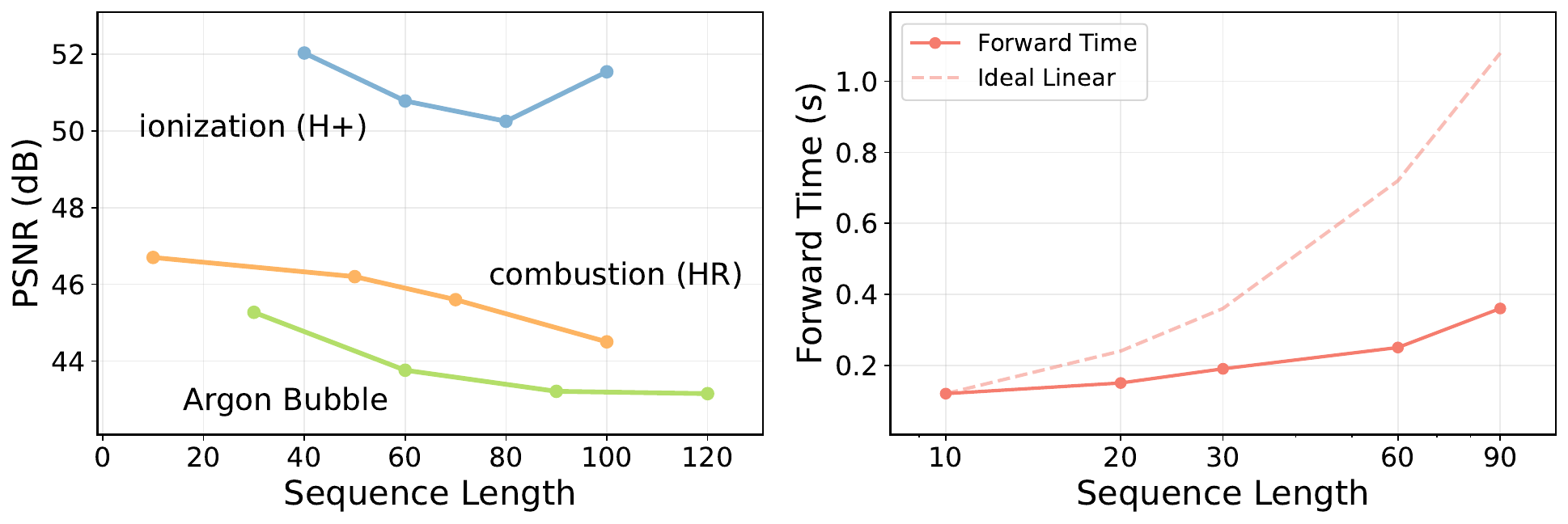}
    \caption{Impact of output sequence length on reconstruction quality and forward efficiency. 
    Left: PSNR under different sequence lengths on three datasets. 
    Right: forward time (seconds) as a function of sequence length, compared with an ideal linear growth.}
    \label{fig:sequence_psnr}
    \vspace{-0.2cm}
\end{figure}

\begin{figure}[!t]
    \centering
    \includegraphics[width=1\linewidth]{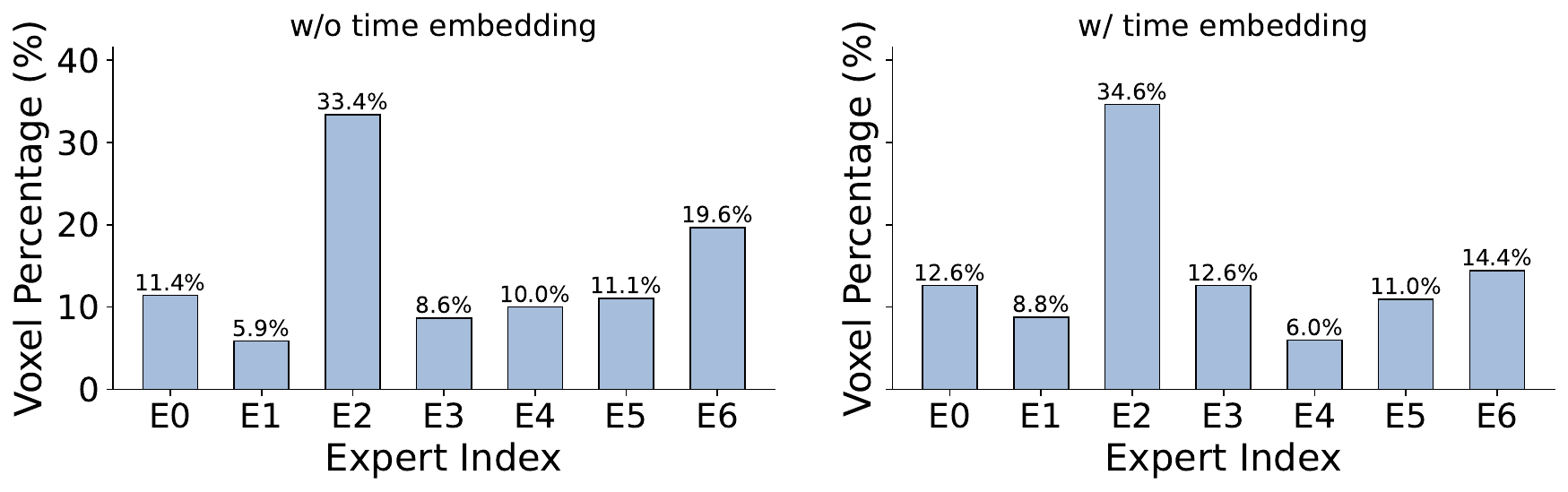}
    \caption{Distribution of voxel assignments across seven experts on the vortex data set, comparing models trained with and without time embedding.}
    \label{fig:expert_voxel_distribution}
    \vspace{-0.2cm}
\end{figure}

\begin{figure}[!t]
    \centering
    \includegraphics[width=1\linewidth]{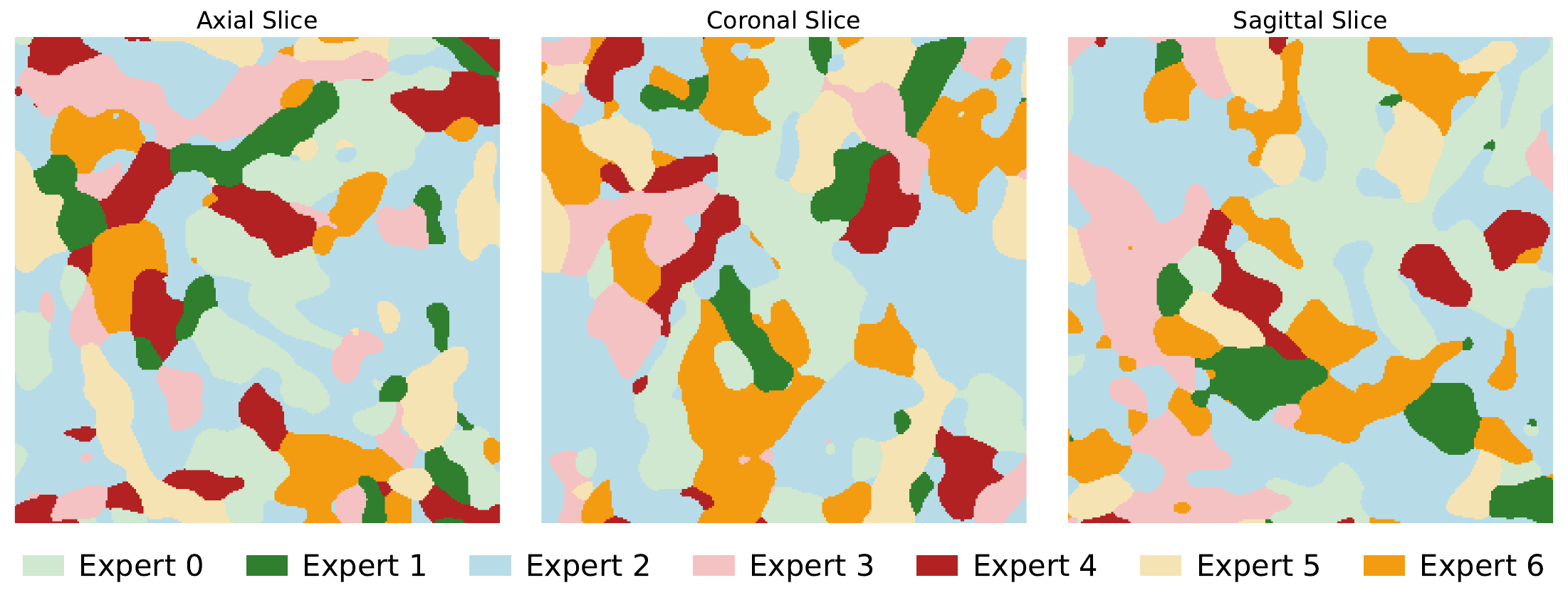}
    \caption{Visualization of voxel-level routing results on three orthogonal slices. Each color represents the expert assigned to the corresponding spatial location, illustrating the spatial distribution of expert routing across the volume.}
    \label{fig:expert_slices}
    \vspace{-0.2cm}
\end{figure}

\textbf{Effect of sequence length.}
To investigate the effect of output sequence length, we vary the output sequence length within our framework to compress time-varying volumetric data of different temporal sizes, while keeping the compression ratio controlled. Specifically, the length of the predicted temporal sequence is directly determined by adjusting the output dimension of the decoder, without modifying other components of the framework.
\cref{fig:sequence_psnr} illustrates the effect of output sequence length on reconstruction quality and forward efficiency. The ionization (H+) dataset exhibits consistently high PSNR, with only minor fluctuations as the sequence length increases. The stable performance on the ionization (H+) dataset is likely due to its relatively simple temporal patterns, especially in the base region of the volume. For the combustion (HR) and argon bubble datasets, PSNR slightly decreases with longer sequences, indicating that increasing the temporal output dimension introduces additional modeling difficulty.
The right plot further shows that the forward time grows slightly slower than the ideal linear trend. This is because increasing the sequence length only enlarges the computational cost in the decoder, while the encoder and routing components remain unchanged. As a result, the overall time scales efficiently with the output dimension.

\textbf{Expert routing.}
\cref{fig:expert_voxel_distribution}~presents the distribution of voxel assignments across seven experts on the vortex dataset, comparing models with and without time embedding. Without time embedding, the router heavily favors a single expert E2, which receives 33.4$\%$ of the total assignments, while others such as E1 and E3 are underutilized.
It is worth noting that imbalanced value distributions are common in volumetric data~\cite{han2025moeinr}.
When time embedding is incorporated, the expert assignments become more evenly distributed across experts other than E2, resulting in a more balanced routing pattern.
This shift indicates that in MoE architectures, balanced routing is often desirable to enhance expert specialization and overall representation capacity. 
The voxel-level routing visualized in the orthogonal slices further supports this observation. As shown in~\cref{fig:expert_slices}, neighboring voxels that form coherent structures, such as iso-surfaces, tend to be assigned to the same expert. This spatial consistency suggests that the routing behavior is correlated with underlying structural patterns, rather than being fragmented or random.

\begin{table}[t]
\centering
\caption{\hot{Decompression time (DT) in seconds and memory usage for reconstructing different numbers of consecutive frames.}}
\label{tab:dt_analysis}
\setlength{\tabcolsep}{3pt}
\footnotesize
\begin{tabular}{llcccc c}
\toprule
\multirow{2}{*}{\textbf{Dataset}} &
\multirow{2}{*}{\textbf{Method}} &
\multicolumn{4}{c}{\textbf{DT (s) for \# Frames}} &
\multirow{2}{*}{\textbf{Mem. (MB)$\downarrow$}} \\
\cmidrule(lr){3-6}
&
& \textbf{T=1}
& \textbf{T=2}
& \textbf{T=4}
& \textbf{T=16}
& \\
\midrule

\multirow{3}{*}{Vortex}
& CoordNet & 7.53 & 15.12 & 30.48 & 120.44 & 6,772 \\
& MoE-INR  & \textbf{6.64} & 13.05 & 26.05 & 103.80 & 5,988 \\
& Ours     & 6.99 & \textbf{7.10} & \textbf{7.18} & \textbf{7.42} & \textbf{5,887} \\

\midrule

\multirow{3}{*}{combustion (CHI)}
& CoordNet & 21.48 & 42.90 & 85.39 & 340.23 & 7,403 \\
& MoE-INR  & \textbf{21.46} & 42.85 & 84.51 & 335.81 & 7,028 \\
& Ours     & 25.38 & \textbf{25.78} & \textbf{25.82} & \textbf{25.91 }& \textbf{6,615} \\

\bottomrule
\end{tabular}
\end{table}

\hot{
\textbf{Decoding multiple time frames.}
The sequence-based formulation introduces a trade-off in time-specific reconstruction and rendering. Unlike coordinate-to-scalar INRs that can directly query arbitrary spatiotemporal coordinates, our decoder predicts a temporal sequence for each spatial coordinate. Therefore, reconstructing or rendering a single target time step may require evaluating the sequence output and then selecting the corresponding frame, which is less flexible for random time-step access. This may also introduce additional memory pressure in rendering scenarios if the full temporal outputs for many sampled spatial locations are cached simultaneously. In practice, this can be mitigated by chunk-based decoding and by retaining only the queried time step during rendering. On the other hand, this design is advantageous for continuous temporal reconstruction, where multiple consecutive frames can be decoded from the same spatial query. As shown in~\cref{tab:dt_analysis}, the decompression time of our method remains nearly constant when reconstructing larger temporal windows, while coordinate-to-scalar baselines increase almost linearly with the number of reconstructed frames. This indicates that our formulation is more suitable for full-sequence or consecutive-frame reconstruction, but less flexible for applications requiring frequent access to specific time steps.
}

\section{Discussion And Conclusion}

\textbf{Limitations and future work.}
We acknowledge two limitations of our framework and discuss future directions. First, although the proposed sequence-based formulation improves compression and decompression efficiency, it does not consistently achieve the highest reconstruction quality across all datasets, particularly for highly complex temporal dynamics. Since the formulation is architecture-agnostic, future work will explore more advanced model instantiations, such as adaptive expert selection and dynamic capacity allocation, to better capture heterogeneous temporal patterns while preserving its efficiency advantages.
Second, our formulation models each spatial location independently and does not explicitly capture temporal correlations among neighboring coordinates. Incorporating such spatial-temporal dependencies through structured priors, e.g., tensor decomposition, may further improve representation capacity while maintaining computational efficiency.

\textbf{Discussion.}
Our formulation models time-varying data as spatially indexed temporal sequences, motivated by the different sampling characteristics across dimensions. In most scientific datasets, time is sampled regularly and evolves smoothly, making temporal sequences naturally suitable for sequence-level modeling. In contrast, spatial variations are often more irregular, making flexible spatial sampling better suited to capture local value changes. While sequences can also be constructed along spatial dimensions (e.g., x or y), they generally exhibit weaker continuity and higher variability. Such a perspective may enable in-situ compression during simulation by progressively encoding data without waiting for all time steps, although it currently relies on structured grids and is not readily applicable to unstructured data.

\textbf{Conclusion.}
\rev{In this work, we present a sequence-wise formulation for time-varying volumetric data compression, where each spatial location is represented as a temporal sequence to reduce dense spatiotemporal sampling and enable efficient sequence-level modeling. We instantiate this formulation with an MoE-based INR framework for adaptive temporal modeling, and experiments on multiple datasets show competitive reconstruction quality with improved compression and decompression efficiency.}

\section*{Acknowledgements}
This research was supported in part by the National Natural Science Foundation of China through grants 62372484 and 62172456.
The authors would like to thank the anonymous reviewers for their insightful comments.

\FloatBarrier
\bibliographystyle{abbrv-doi-hyperref}
\bibliography{template}

\clearpage
\appendix

\setcounter{page}{1}
\setcounter{figure}{0}
\setcounter{section}{0}
\setcounter{table}{0}

\begin{figure*}[ht]
    \centering
    \includegraphics[width=\linewidth]{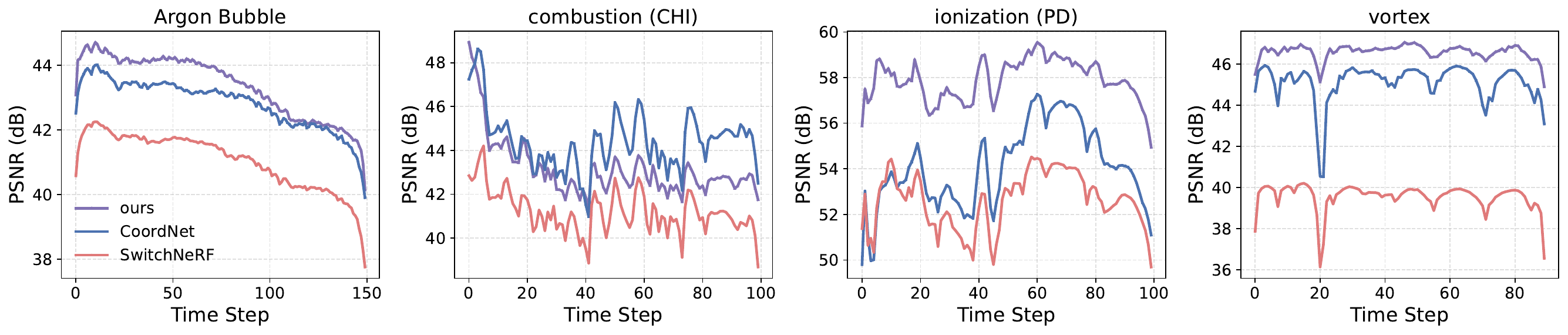}
    \caption{Per-time-step PSNR comparison among different methods on four datasets: Argon Bubble, combustion (CHI), ionization (PD), and vortex. The curves illustrate reconstruction quality variations across time steps, showing the performance of three methods under the same compression setting.}
    \label{fig:all_time_psnr}
\end{figure*}

\section{Network Configuration}
In this section, we provide the network architecture and training of our framework. 

\textbf{Network architecture.}
Our model follows a mixture-of-experts design that maps a 3D spatial coordinate directly to its corresponding temporal sequence. Given input coordinates, the network first encodes spatial information into a shared high-dimensional representation. In parallel, the router determines the most suitable expert for each coordinate by incorporating both spatial features and a learnable global time embedding. Each expert decoder then predicts the full temporal sequence in a single forward pass, and the final output is selected through hard routing. 


\begin{table}[h]
\centering
\footnotesize
\setlength{\tabcolsep}{3pt}
\caption{\rev{Architecture details of framework. $M$ denotes the embedding dimension, $K$ the number of experts, $T$ the sequence length, and $r$ the LoRA rank.}}
\label{tab:arch_details}
\begin{tabular}{l l c c}
\toprule
\textbf{Module} & \hot{\textbf{Layer/Pars}} & \textbf{In} & \textbf{Out} \\
\midrule

\multirow{4}{*}{Encoder}
& Linear (PE)      & $3$   & $M$ \\
& Linear (Sine)    & $M$   & $2M$ \\
& Linear (Sine)    & $2M$  & $4M$ \\
& Res. Sine Block  & $4M$  & $8M$ \\

\midrule

\multirow{5}{*}{Router}
& Linear (Sine)    & $3$   & $M$ \\
& Linear (Sine)    & $M$   & $M$ \\
& \hot{Time Emb. Table}  & \hot{-- }   & \hot{$T\times M$} \\
& Time Proj.       & $M$   & $M$ \\
& Classifier       & $9M$  & $K$ \\

\midrule

\multirow{3}{*}{Decoder}
& Shared Expert ($\mathbf{W}_0$) & $8M$ & $T$ \\
& LoRA-A ($\mathbf{A}_k$)        & $8M$ & $r$ \\
& LoRA-B ($\mathbf{B}_k$)        & $r$  & $T$ \\

\bottomrule
\end{tabular}
\end{table}

\textbf{The Encoder} maps a 3D spatial coordinate into a high-dimensional latent representation. It consists of an initial linear projection followed by multiple sine-activated layers and a residual sine block. This design enhances the capacity to model high-frequency variations while maintaining stable optimization. 

\textbf{The Router} determines expert assignment for each coordinate. It employs lightweight sine layers to extract spatial features and integrates a trainable time embedding through a fusion operation. The fused representation is then passed to a linear classifier to produce routing logits over K experts.
The trainable global time embedding is first projected into the routing feature space and summarized using mean and variance pooling across time steps. The resulting temporal context is fused through a linear layer and added to the spatial routing features to modulate expert assignment. The time embedding parameters are optimized jointly with the entire network in an end-to-end manner.
\hot{Since the embedding table contains only $T \times M$ parameters, its memory overhead is negligible compared with the full model. For example, for the vortex configuration with $T=90$ and $M=352$, the time embedding contains 31.7K parameters, accounting for about 1.2$\%$ of the model parameters.}

\textbf{The Decoder} is implemented as a parallel expert structure. All experts share a base weight matrix, while expert-specific variations are modeled through low-rank matrix updates. Each routed expert predicts the full temporal sequence of length T in a single forward pass. Since the decoder is designed as a lightweight linear mapping that directly predicts the temporal sequence, the LoRA rank r can be set to a small value without significantly degrading reconstruction quality. In practice, we set the rank to 8. 
As for the number of experts, we follow prior MoE-based INRs~\cite{zhao2023moec,han2025moeinr} and set the number of experts to seven (excluding the shared base component). In practice, we observe that using fewer experts leads to noticeable performance degradation, while increasing the number of experts beyond this setting yields only marginal improvements.

\textbf{Training and inference.}
We compress each variable of a multivariate time-varying volumetric dataset independently. For each variable in training, the network is trained to learn a mapping from 3D spatial coordinates to the corresponding temporal sequence. 
During inference, the trained network directly predicts the full temporal sequence for a queried spatial coordinate in a single forward pass.
Our implementation is based on PyTorch, and all training and inference are performed on an NVIDIA RTX A100 GPU equipped with 80GB of memory.
During both training and inference, the input coordinates and corresponding output values are normalized to the range of $[-1, 1]$, following the preprocessing strategy in~\cite{han2022coordnet}. Specifically, the output values are normalized independently at each time step to account for potential variations in value ranges across different temporal frames.
We initialize the network parameters following the strategy proposed by Sitzmann~et~al~\cite{sitzmann2020siren}. The model is trained for 300 epochs with a batch size of 1,600 sampled spatial coordinates per iteration. Following Sitzmann~et~al.~\cite{sitzmann2020siren}, we observe that using relatively small batch sizes is beneficial for stabilizing INR training. Since our formulation operates purely on spatial coordinates without explicitly sampling the temporal dimension, it naturally allows efficient training with moderate batch sizes. We warm up the routing network during the first 10$\%$ of the training epochs to guide the expert assignment process toward a more stable initialization.
We use the Adam optimizer with an initial learning rate of $10^{-5}$, momentum parameters $\beta_1 = 0.9$, $\beta_2 = 0.999$, and apply $L_2$ regularization with a weight decay of $10^{-6}$. The learning rate decayed by half with a multistep scheduler. All computations are performed in 32-bit floating point during training and inference, while 16-bit floating point precision is used for model storage to reduce memory consumption.

\section{Additional Ablation}

\begin{table}[t]
\centering
\caption{CR, PSNR, and DreamSim with and without PE on the combustion (CHI) dataset.}
\label{tab:pe_ablation}
\setlength{\tabcolsep}{6pt}
\begin{tabular}{l c c c}
\toprule
\textbf{PE} & \textbf{CR $\uparrow$} & \textbf{PSNR (dB) $\uparrow$} & \textbf{DreamSim $\downarrow$} \\
\midrule
w/o PE & 2,484 & 41.22 &  0.0898\\
\hot{w/ hash PE~\cite{muller2022instant}} & \hot{1,710} &\hot{38.82} & \hot{0.1294}\\
\hot{w/ harmonic PE~\cite{han2025dcinr}}  & 2,585 & 43.85 & 0.0374 \\
\bottomrule
\end{tabular}
\end{table}
\begin{table}[t]
\centering
\caption{Effect of different warm-up training ratios on reconstruction quality on the ionization (PD) dataset.}
\label{tab:warmup_ratio}
\begin{tabular}{ccc}
\toprule
\textbf{Warm-up Ratio} & \textbf{PSNR (dB)$\uparrow$} & \textbf{LPIPS$\downarrow$} \\
\midrule
w/o warm up    & 55.89 & 0.071 \\
5\%   & 56.49 & 0.047 \\
10\%   & 57.89 & 0.010 \\
20\%   & 57.32 &  0.012\\
30\%   & 56.98 & 0.064 \\
\bottomrule
\end{tabular}
\end{table}

\textbf{Positional encoding.}
\hot{To analyze the impact of positional encoding (PE), we compare no PE, a learnable hash-grid PE~\cite{muller2022instant}, and the harmonic PE used in our model\cite{han2025dcinr} on the combustion (CHI) dataset. For the hash-grid PE, the deterministic harmonic mapping is replaced by a learnable multi-resolution grid encoding, and all grid parameters are included in the compressed representation when computing CR.
As shown in~\cref{tab:pe_ablation}, harmonic PE achieves the best trade-off among the tested settings, improving PSNR from 41.22 dB to 43.85 dB and reducing DreamSim from 0.0898 to 0.0374 while maintaining a similar CR. In contrast, the hash-grid PE introduces additional learnable parameters, reducing CR to 1,710, and does not improve PSNR or DreamSim on this dataset. These results motivate our use of harmonic PE as a lightweight deterministic encoding for sequence-level volumetric compression.}

\textbf{Warm-up.}
To evaluate the influence of warm-up training in optimization, we vary the warm-up ratio on the ionization (PD) dataset and report the corresponding PSNR and LPIPS values in~\cref{tab:warmup_ratio}.
As shown in the table, introducing warm-up training consistently improves reconstruction quality compared to the setting without warm-up. Performance peaks at a 10$\%$ warm-up ratio, achieving the highest PSNR and the lowest LPIPS. Increasing the ratio beyond 10$\%$ does not further improve performance; instead, both PSNR and LPIPS slightly degrade at 20$\%$ and 30$\%$.
This trend suggests that a moderate warm-up phase helps the router learn more reliable expert assignments in the early stage of training. However, excessive warm-up may over-constrain the routing behavior, reducing the flexibility of joint optimization in later stages.

\begin{table}[t]
\centering
\caption{\hot{Effect of LoRA rank on reconstruction quality, expert diversity, and model size. Expert variance is computed as the mean variance of predictions across experts, following the uncertainty analysis in MoE-INR~\cite{han2025moeinr}.}}
\label{tab:lora_rank_analysis}
\footnotesize
\setlength{\tabcolsep}{4pt}
\begin{tabular}{l c r c c}
\toprule
\textbf{Dataset} & \textbf{LoRA Rank} & \textbf{Params} & \textbf{PSNR (dB)$\uparrow$} & \textbf{Expert Var.} \\
\midrule
\multirow{5}{*}{combustion (MF)}
& -- & 7.82M & 42.46 & $5.65{\times}10^{-2}$ \\
& 2  & 6.71M & 40.55 & $3.98{\times}10^{-5}$ \\
& 4  & 6.72M & 40.53 & $4.85{\times}10^{-5}$ \\
& 8  & 6.75M & 40.87 & $8.10{\times}10^{-5}$ \\
& 16  & 6.80M & 40.61 & $7.00{\times}10^{-5}$ \\
\midrule
\multirow{5}{*}{Vortex}
& -- & 3.45M & 46.17 & $1.21{\times}10^{-2}$ \\
& 2  & 2.70M & 46.84 & $8.07{\times}10^{-6}$ \\
& 4  & 2.71M & 46.85 & $9.79{\times}10^{-6}$ \\
& 8  & 2.73M & 48.18 & $2.17{\times}10^{-5}$ \\
& 16  & 2.77M & 47.80 & $1.44{\times}10^{-5}$ \\

\bottomrule
\end{tabular}
\end{table}

\begin{table}[t]
\centering
\caption{\hot{Effect of the number of experts on compression ratio, reconstruction quality, and memory usage. The dataset used is vortex.}}
\label{tab:num_experts_analysis}
\setlength{\tabcolsep}{6pt}
\footnotesize
\begin{tabular}{c c c c}
\toprule
\textbf{\# Experts} & \textbf{CR$\uparrow$} & \textbf{PSNR (dB)$\uparrow$} & \textbf{Mem. (MB)$\downarrow$} \\
\midrule
4  & 1,124 & 47.85 & 5,465 \\
7  & 1,120 & 48.18 & 5,887 \\
12  & 1,113 & 48.33 & 6,201 \\
32  & 1,088 & 48.44 & 12,195 \\
\bottomrule
\end{tabular}
\end{table}

\hot{
\textbf{LoRA.}
Table~\ref{tab:lora_rank_analysis} reports the impact of different LoRA ranks on reconstruction quality, model size, and expert diversity. The primary motivation for introducing LoRA is to reduce the number of trainable parameters and improve model compactness, which is particularly important in the context of neural compression. Across both datasets, LoRA reduces the parameter count by approximately 20$\%$ while preserving most of the reconstruction quality. Furthermore, increasing the LoRA rank from 1 to 16 results in only minor changes in PSNR, suggesting that reconstruction quality is relatively insensitive to the rank value. Therefore, a small rank is generally sufficient to achieve a favorable trade-off between model compactness and reconstruction quality.
To further analyze this behavior, we follow MoE-INR~\cite{han2025moeinr} and measure expert diversity using the variance of predictions across experts. As shown in Table~\ref{tab:lora_rank_analysis}, applying LoRA significantly reduces expert variance compared with the full decoder, indicating that experts become more similar due to the shared low-rank parameterization.
Interestingly, the impact on reconstruction quality is dataset dependent. For the combustion (MF) dataset, LoRA introduces a noticeable PSNR drop together with a substantial reduction in expert variance. This suggests that MF contains highly heterogeneous dynamics that benefit from stronger expert specialization. In contrast, the vortex dataset maintains comparable reconstruction quality despite the reduced expert variance. We attribute this behavior to the more homogeneous data distribution in vortex. A histogram analysis of voxel values will show that the value distribution in vortex is more concentrated and uniform than that of MF. Consequently, reducing expert diversity has a much smaller impact on reconstruction quality. In this case, the parameter sharing introduced by LoRA acts as a lightweight regularizer and achieves a favorable trade-off between model compactness and reconstruction quality.

\textbf{Expert numbers.}
As shown in~\cref{tab:num_experts_analysis}, increasing the number of experts improves reconstruction quality but also introduces additional memory cost and slightly reduces the compression ratio. On the vortex dataset, increasing the number of experts from 4 to 12 improves PSNR from 47.85 dB to 48.33 dB, while the memory usage increases moderately from 5,465 MB to 6,201 MB. However, further increasing the number of experts to 32 only brings a marginal PSNR improvement to 48.44 dB, but nearly doubles the memory usage to 12,195 MB and further reduces the CR to 1,088. This indicates that using more experts improves modeling capacity, but the benefit quickly saturates while the memory overhead continues to grow. In our main experiments, we use 7 experts following the settings of prior MoE-based INR methods~\cite{zhao2023moec,han2025moeinr}. The additional result with 32 experts serves as a more aggressive configuration to examine whether substantially increasing the expert number can further improve performance. The results show that such an increase brings only limited PSNR improvement but substantial memory overhead. Therefore, we select 7 experts as a practical trade-off between reconstruction quality, compression ratio, and memory efficiency.

}

\hot{\section{Quantitative Analysis of Temporal Behavior}}
\textbf{Performance across timesteps.}
\cref{fig:all_time_psnr} shows the per-time-step PSNR curves on four datasets: Argon Bubble, combustion (CHI), ionization (PD), and vortex. For Argon Bubble, the reconstruction quality gradually decreases as the time step increases, indicating a progressive increase in temporal complexity. In ionization (PD), the PSNR values fluctuate more irregularly across time, reflecting non-uniform temporal dynamics.
For the combustion (CHI) dataset, the decreasing PSNR values over time are due to the increasingly turbulent nature of the combustion process. As the data gets more complex, the volume rendering results show larger differences with respect to the GT.
The vortex dataset exhibits relatively stable performance with occasional sharp drops at specific time steps. These observations suggest that reconstruction quality is closely related to the temporal evolution characteristics of each dataset, and different datasets exhibit distinct time-dependent behaviors.

As shown in~\cref{fig:all_time_psnr}, the temporal variations of our method closely follow those of scalar-prediction INR models, indicating that the sequence-wise formulation preserves the underlying temporal evolution. Moreover, our method exhibits smaller fluctuations across time steps, suggesting more stable behavior when modeling temporal sequences. This reduced variance is more evident in datasets with relatively smoother dynamics, while in more complex cases, the overall trends remain consistent despite increased variability. For example, on the ionization (PD) and vortex datasets, our method exhibits relatively smaller fluctuations across time steps, resulting in more stable temporal behavior.

\begin{figure}[h]
    \centering
    \includegraphics[width=1\linewidth]{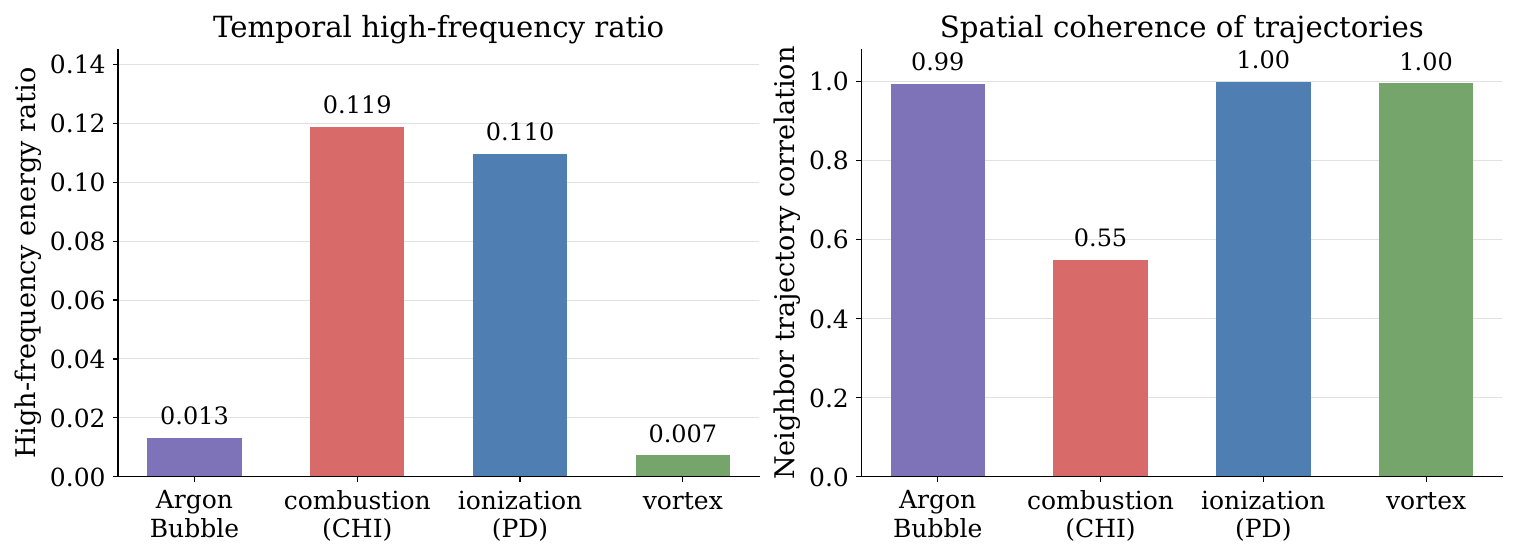}
    \caption{Analysis of temporal complexity and spatial coherence of temporal trajectories.}
    \label{fig:temporal_complexity_appendix}
\end{figure}

\hot{
\textbf{Impact of temporal complexity and spatial coherence of sequences.}
To further understand this behavior, we analyze the temporal high-frequency ratio and the spatial coherence of temporal trajectories in~\cref{fig:temporal_complexity_appendix}. 
These two metrics are computed as follows. We randomly sample a set of spatial locations and perform Fourier analysis on the temporal sequence at each location. For each sequence, the temporal high-frequency ratio is computed as the proportion of high-frequency energy in these sampled trajectories, reflecting how rapidly the scalar values change over time.
We also randomly sample neighboring voxel pairs and compute the Pearson correlation between their temporal sequences. The neighbor trajectory correlation is obtained by averaging these correlations over all sampled pairs, measuring the degree to which adjacent spatial locations exhibit similar temporal evolution.

The results show that combustion (CHI) has a high temporal high-frequency ratio and the lowest neighbor trajectory correlation among the evaluated datasets. This indicates that CHI contains rapid temporal variations that are also less spatially coherent, making full-sequence prediction more challenging. 
In contrast, ionization (PD) also exhibits relatively strong temporal variations, but its neighbor trajectory correlation remains high. This suggests that although PD changes rapidly, its temporal evolution is more spatially coordinated and structured, making it more suitable for sequence-level modeling. 
Similarly, vortex exhibits high spatial trajectory coherence, indicating that neighboring spatial locations tend to share similar temporal evolution patterns. Such characteristics are favorable for the proposed sequence-based formulation, as the model can exploit both temporal regularity and spatially coherent temporal trajectories.

Argon Bubble shows fewer high-frequency signals and high spatial coherence, but its reconstruction quality still gradually decreases over time, indicating that temporal complexity can also accumulate progressively as the simulation evolves. 
Overall, these results suggest that the effectiveness of sequence modeling depends not only on the magnitude of temporal complexity, but also on whether the temporal dynamics are spatially coherent. Datasets with high temporal frequency and weak spatial trajectory coherence, such as combustion (CHI), represent more challenging cases where the full-sequence prediction target may lead to reconstruction degradation.
}

\end{document}